\documentclass[letterpaper]{article} 
\usepackage[submission]{aaai2027}  
\usepackage[hyphens]{url}  
\usepackage{graphicx} 
\def\UrlFont{\rm}  
\usepackage{natbib}  
\usepackage{caption} 
\usepackage{amsmath,amssymb,amsfonts}
\usepackage{booktabs}
\usepackage{multirow}
\usepackage{xspace}
\usepackage{xcolor}
\usepackage{tikz}
\usetikzlibrary{arrows.meta,positioning,fit,backgrounds,calc,shapes.geometric,decorations.pathreplacing}

\definecolor{cpre}{HTML}{F4F6FA}   
\definecolor{cenc}{HTML}{FAE9D4}   
\definecolor{cagg}{HTML}{E3EBF7}   
\definecolor{cpred}{HTML}{DDEEDC}  
\definecolor{closs}{HTML}{FAE4E4}  
\definecolor{cobj}{HTML}{F0F0EE}   
\definecolor{crule}{HTML}{23262B}
\definecolor{cdash}{HTML}{9A9A9A}
\definecolor{clead}{HTML}{8B8B8B}

\newcommand{\method}{\textsc{FABLE-Therm}\xspace}
\newcommand{\R}{\mathbb{R}}
\newcommand{\pmstd}[1]{\ensuremath{{\,\scriptstyle\pm#1}}}

\title{Representation Is Not Enough: Body-Localized Thermal Evidence for\\
Contactless Stress and Craving Sensing in Opioid Use Disorder}

\makeatletter \def\showauthors@on{T} \makeatother

\author{
    Sachin Deb,
    Harshit Sharma,
    Asif Salekin
}
\affiliations{
    Arizona State University\\
    sachinde@asu.edu, hsharm62@asu.edu, asif.salekin@asu.edu
}
\begin{document}
\maketitle

\begin{abstract}
\noindent
Removing the wearable from physiological monitoring also removes its supervision: the signal indicating \emph{where} on the body and \emph{when} a stress response occurred. Contactless stress sensing therefore becomes a weakly supervised evidence-localization problem, where a single clip label must be traced back to the body regions and moments that produced it. We argue this problem is solved by one principle: \emph{preserve localized evidence under weak supervision}, keeping evidence tied to body regions, temporal evolution, and encoder-specific representations until the final decision. We realize this in \method, a weakly supervised architecture that fuses frozen foundation-model encoders at the embedding level, with theory explaining why localized fusion succeeds where feature concatenation and prediction averaging fail. We study this in opioid use disorder (OUD), where stress is a leading relapse trigger and wearables are difficult to sustain through early recovery, making contactless sensing enabling rather than merely convenient. From a fixed thermal camera, \method reaches $0.938$ AUROC on held-out participants, and its representation transfers to self-reported craving, providing, to our knowledge, the first evidence that craving is recoverable from contactless thermal video. Beyond improving accuracy, localized evidence enables a participant-level analysis of \emph{why} deployment fails, not merely \emph{that} it does. The central finding is that improving representation alone is insufficient for equitable deployment: collecting more data from the underserved group would recover only about half of the cohort gap, the remainder reflecting person-to-person heterogeneity that additional data cannot remove. This modality-agnostic decomposition applies to any model with an identifiable subpopulation. Together with the first cohort-structured contactless thermal OUD benchmark, our results show that preserving localized evidence enables both accurate contactless sensing and a principled understanding of \emph{who} a model fails, and \emph{why}.
\end{abstract}



\section{Introduction}

Imagine detecting stress with a thermal camera mounted on a wall instead of a wearable, the conventional approach to physiological stress sensing \citep{schmidt2018wesad}. A camera is far less burdensome, but it also removes the supervision provided by wearable biomarkers, which reveal \emph{where} and \emph{when} a stress response occurs \citep{shah2024unveiling}. Left with only one coarse label per clip, the model must identify itself the few regions and moments that support it.

This is a \emph{weakly supervised evidence-localization} problem, and we argue it is solved by one principle: \textbf{preserve localized evidence under weak supervision}---keeping each region's and moment's potential evidence intact and deciding which matter only at the final decision.
We instantiate this principle in \method, a weakly supervised ensemble of frozen foundation models (FMs) that treats multiple encoders as a \emph{single} multiple-instance learning (MIL) problem while preserving spatial and temporal provenance throughout inference. We study it in opioid use disorder (OUD), where stress is a leading relapse trigger \citep{volkow2016neurobiology}, yet those who would benefit most from continuous monitoring are least well served by wearables that must be worn, charged, and tolerated through recovery \citep{moon2025brief}. A thermal camera removes that burden, sensing autonomic arousal at a distance \citep{ioannou2014thermal,cardone2015thermal}, and on a cohort-structured thermal benchmark of Control and OUD participants that we collect and release, \method turns that raw signal into a reliable stress readout.


We evaluate \method along three axes. First, it generalizes to unseen participants, reaching $0.938$ AUROC, $0.203$ above the strongest of 27 baselines. Second, retrained on self-reported craving labels, a related but distinct relapse-associated construct, it reaches $0.752$ AUROC and outperforms the strongest craving baseline, providing a feasibility demonstration that contactless thermal sensing carries craving-related signals. Third, it transfers unchanged to the public StressNet dataset despite differences in sensor, stressor, and spatial resolution. Notably, trained with or without the OUD cohort while resolving performance participant by participant, we decompose the \method's deployment deficit into a \emph{representation} term (recovered by training on the underserved group) and a \emph{within-cohort heterogeneity} term (person-to-person variation that persists despite it). These are comparable ($\Delta_{\text{repr}}{=}0.158$ vs.\ $\Delta_{\text{heter}}{=}0.169$, $\approx$48/52), showing that improving representation alone is insufficient for equitable deployment: collecting more data from the underserved group closes only about half the gap, the rest reflecting inter-individual heterogeneity. This is the paper's scientific finding through evaluation.

\paragraph{Contributions.}
The novelty is not frozen encoders, MIL, or ensembling individually---each is established---but their integration into a weakly supervised framework that keeps evidence body-localized until the decision. Concretely:

\begin{enumerate}
\item \textbf{A problem formulation:} we recast contactless thermal stress or craving sensing as weakly supervised \emph{evidence localization}: which regions, over which moments, support a thermal video window label (Sec.~\ref{sec:formulation}).

\item \textbf{\method}, an ensemble of $E$ frozen encoders whose evidence stays tied to (view, region) instances until a single learned gate fuses it at the \emph{embedding} level. Three properties new to multi-view MIL each motivate a design choice. Attention pooling over regional \emph{trajectories} is order-blind (Prop.~1), so we add a parameter-free trajectory-statistics path whose signed slope restores time order. The $E$-view ensemble is \emph{exactly} one MIL problem over $E{\times}N$ (view, region) pairs (Prop.~2), so we pool once over the product bag; the allocation factorizes into a shared region scorer and a per-window view gate, letting the model reallocate across encoders clip by clip. And fusion is well-posed only under a \emph{shared} aggregation operator (Prop.~3), so we tie the encoder branch weights; untying them collapses AUROC by up to $0.19$ despite more parameters. Embedding-level fusion reaches $0.938$ AUROC vs.\ $0.736$ (feature-level) and $0.735$ (posterior-level) fusion of the same features (Sec.~\ref{sec:method}). Props.~2--3 are a design rule for \emph{any} fusion of multiple frozen FM encoders; not a thermal-specific construction.

\item \textbf{A public cohort-structured thermal benchmark}: $71$ sessions from $67$ participants ($42$ Control, $25$ OUD; $16{,}273$ windows) from a fixed thermal camera under a shared protocol, with \emph{two} label sets: task-induced stress on all windows and self-reported craving on the $4{,}863$ OUD windows; so one corpus supports
both a general stress task and a harder, clinically salient craving task---\emph{the first evidence, to our knowledge}, that OUD craving is recoverable from contactless thermal video at all. To our knowledge, it is the first openly available benchmark pairing contactless thermal video of a clinical OUD cohort with a matched control group; we release derived representations, cohort labels, and fixed person-disjoint splits publicly, with raw clinical video under an IRB data-use agreement (Sec.~\ref{subsec:datacollection}).

\item \textbf{A cohort-transfer decomposition} separating a subgroup's deficit into representation vs. heterogeneity terms with participant-level significance (Sec.~\ref{sec:gap}). It needs nothing thermal or OUD-specific; any model with an identifiable subpopulation and a held-out slice can compute it, so we argue it should be default reporting whenever an equity claim rests on data collection.

\end{enumerate}

\section{Related Work}
We summarize the literature most relevant to this work; Appendix~A provides a fuller treatment and structured comparison with the closest methods.

\paragraph{Stress sensing and thermal psychophysiology.} Physiological stress recognition was built on contact sensors \citep{schmidt2018wesad}; camera-based methods reduce contact but change the supervision signal, via rPPG \citep{aljebrni2020stress} or datasets pairing video with contact references \citep{sabour2023ubfcphys}, with subject dependence and inconsistent labeling recurring as barriers \citep{giannakakis2019review}. In addiction science the challenge is specific: stress reactivity and craving are related but distinct; stress often induces craving in OUD, yet craving also occurs without stress and varies across individuals and early abstinence \citep{sinha2009modeling,sinha2000psychological,fox2008enhanced}. So we supervise stress from task structure while treating craving as a separate self-reported outcome (Section~\ref{sec:eval}), making person-to-person variability an expected phenomenon rather than noise. HHISS studies heterogeneity across participants and OUD status \emph{using wearables} \citep{xiao2025hhiss}; we study it without one. Craving itself has been tracked through wearables \citep{luo2026personalized}, but to our knowledge not from contactless thermal video. Thermally, imaging reflects perfusion, perspiration, and vasomotor tone rather than a single channel \citep{ioannou2014thermal,cardone2015thermal}. Early work localized stress to fixed facial regions \citep{shastri2012perinasal}, but the responding site and even the direction of temperature change vary across people, development, and stressors \citep{nazzari2025review}, and a review of 315 studies reports inconsistent emissivity, distance, and ambient-control protocols \citep{stanic2025review}. Fixed ROIs thus encode one cohort's physiology as universal and can fail silently on new populations; we instead learn relationships \emph{among} regions. Existing thermal methods still rely on predefined ROIs, auxiliary sensing, or healthy participants under a single protocol \citep{kumar2021stressnet,xiao2024reading,liu2024blush}; \citet{brugge2026taskaware} is the closest weak-supervision approach, but its instances are purely temporal, whereas ours are spatiotemporal.

\paragraph{Foundation models, MIL, and Disparity in health.}
Frozen self-supervised encoders transfer across domains \citep{oquab2024dinov2,dosovitskiy2021vit}, and thermal-FMs have emerged \citep{liu2024infmae,maheshwari2026anythermal}. Multiple-instance learning is standard for weak supervision \citep{ilse2018abmil,campanella2019clinical,kurata2025pathfm_mil}, but existing methods assume static instances from a single encoder---assumptions Section~\ref{sec:method} relaxes. Separately, algorithmic bias in health care is well documented \citep{obermeyer2019bias,chen2021ethical}, yet most work attributes disparities primarily to under-representation; Section~\ref{sec:gap} tests an alternative by decomposing performance differences into representation and within-population heterogeneity on a corpus.

\section{Problem Formulation}
\label{sec:formulation}

Participant $p_i$ contributes a thermal video window $X_i=(x_{i1},\dots,x_{iT})$, where $x_{it}$ is a single frame, with one binary label $y_i$ from the elicitation condition, and a model predicts $\hat p_i=f_\theta(X_i)$. This label $y_i$ does not specify \emph{where} or \emph{when} the response occurs, so the model must recover that latent structure; evaluation with person-disjoint splits must then test whether the recovered evidence generalizes across cohorts.


\paragraph{Learning under weak supervision.} In thermal video, three important quantities are latent: the responding body region $n$, the time $t$ at which the response develops, and the surrounding body context needed to distinguish true responses from nuisance variation (e.g., the same warming everywhere is ambient; a warm site against an otherwise stable body is a response). Existing approaches remove this ambiguity by assumption, using global pooling or fixed ROIs, but such assumptions can fail silently when physiology differs across cohorts. We instead model the latent structure explicitly. Each window is treated as a bag $B_i=\{h_{i1},\ldots,h_{iN}\}$ of body-localized regional trajectories, and learning asks which instances support the window label $y_i$. This is a MIL problem yielding three design requirements addressed in Section~\ref{sec:method}: \textbf{(R1)} preserve \emph{where}, since global pooling dilutes localized responses; \textbf{(R2)} preserve \emph{how} the response evolves, since warming and cooling trajectories may share the same mean; and \textbf{(R3)} interpret each region relative to the rest of the body.

\paragraph{Evaluation under cohort shift.} Because performance may differ across cohorts, aggregate accuracy is not the
right objective. In a practical health setting, selecting models by average AUROC alone can hide failures on the underserved cohort. Because our corpus is majority-Control (42 versus 29), a model can improve average performance while degrading OUD performance. We therefore evaluate each cohort both when it is included in training and when it is held out entirely. These paired measurements make the representation-vs.-heterogeneity decomposition of Section~\ref{sec:gap} possible.


\section{FABLE-Therm Architecture}
\label{sec:method}

\subsection{Overview}
\begin{figure}[]
\centering
\includegraphics[width=\linewidth]{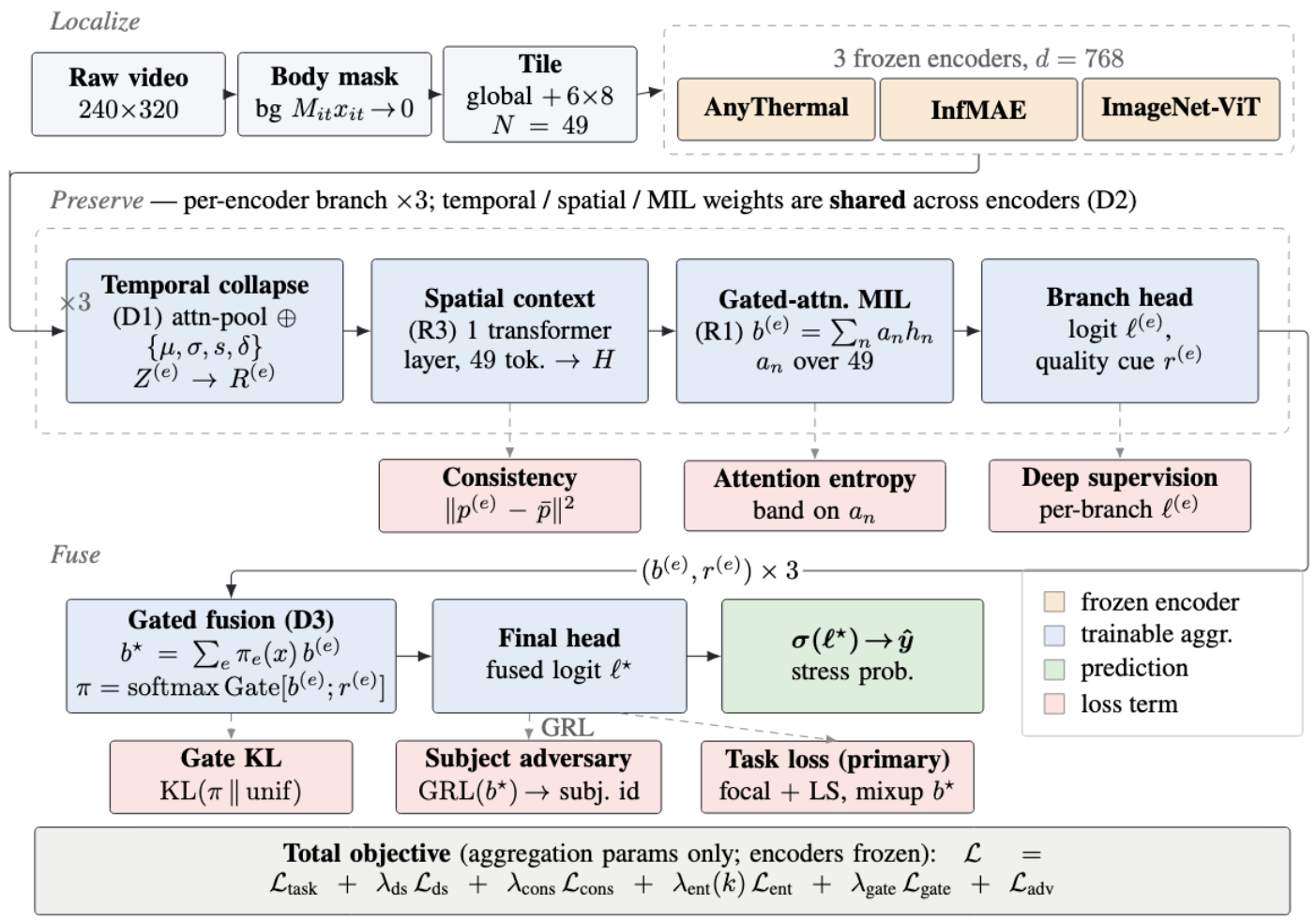}
\caption{\textbf{Model: Localize $\rightarrow$ Preserve $\rightarrow$ Fuse.}
Illustrated for Dataset I, $3$ frozen encoders each drive an identically structured branch (temporal collapse $\to$ spatial context $\to$ gated MIL $\to$ head) whose temporal, spatial, and MIL weights are a \emph{single shared operator} across encoders (D2); only the aggregation parameters are trained. A per-window gate fuses the branch embeddings once, at the embedding level (D3). Red boxes mark the six training losses at their tap points. Blue = trainable, orange = frozen.}
\label{fig:pipeline}
\end{figure}

Figure~\ref{fig:pipeline} shows the pipeline. A thermal-video window is body-masked, tiled into $N$ regions (a global token plus a grid), and passed through $E$ frozen encoders, each producing a short embedding \emph{trajectory} per region. The \emph{aggregator}---the only trained component---then (i) collapses each trajectory into a region vector, (ii) lets regions attend to one another and pools them into one bag embedding per encoder, and (iii) fuses the $E$ bag embeddings with a learned per-window gate, which a final head maps to the stress probability. The contribution is this aggregator, not the encoders. Standard attention MIL \citep{ilse2018abmil} assumes static instances from a single source; here each instance is a trajectory, $E$ encoders give complementary views of each region, and the weak label must supervise both \emph{where} and \emph{when}. \method therefore departs from standard MIL on three axes, each backed by a proposition: \textbf{(D1)} instances are trajectories, not points, and attention pooling is provably order-blind (Prop.~1); \textbf{(D2)} the $E$ branches share one aggregation operator, since untied operators leave the fused embedding ill-posed (Prop.~3); and \textbf{(D3)} a learned per-window gate fuses \emph{pooled embeddings}, dominating feature- and posterior-level fusion (Prop.~2, Obs.~4). The subsections develop each in turn.

\subsection{Representation}
\paragraph{Localized tiling.} To keep small regional responses from being averaged away, we retain both a global token and a grid of local regions. A segmenter \citep{qiao2021detectors} suppresses the background (since walls and equipment can dominate the image and correlate with room or cohort) by masking each frame, $\tilde{x}_{it}=M_{it}x_{it}$. Each window is tiled into $N=1+RC$ tokens: one global token plus an $R\times C$ grid ($R$ rows, $C$ columns). Because the architecture depends only on $N$, adapting to a new thermal sensor requires only changing the grid; dataset-specific frame rates, window sizes, and grid dimensions are in Appendix~C.

\paragraph{Frozen encoders.} With only $67$ participants, fine-tuning a backbone would mostly memorize identity, so we freeze the encoders and spend the limited supervision on the aggregator (compared against LoRA variations in Section~\ref{sec:arch}). Because no single FM is clearly best for thermal affect, we use three with complementary pretraining: \textbf{AnyThermal}, a thermal-distilled DINOv2 ViT-B/14 \citep{oquab2024dinov2,maheshwari2026anythermal}; \textbf{InfMAE}, an infrared masked autoencoder \citep{liu2024infmae}; and \textbf{ImageNet-ViT} \citep{dosovitskiy2021vit}. All stay frozen and emit $d{=}768$-dimensional embeddings, so each encoder $e$ maps region $n$ to a trajectory $Z^{(e)}_{\cdot,n}$ over the $T$ frames. Note: These FMs were selected empirically (Appendix~H).

\subsection{Weakly Supervised Aggregation}
Each branch turns one encoder's regional trajectories into a bag embedding $b^{(e)}$ in three steps.

\paragraph{Temporal collapse (D1; R2).} Attention over time, $\alpha_{t,n}\propto\exp(w_2^\top\tanh(W_1Z_{t,n}))$, yields a weighted average $\bar Z_n=\sum_t\alpha_{t,n}Z_{t,n}$ that captures \emph{when} evidence appears. But it is blind to \emph{how} evidence evolves: attention pooling is order-invariant, so a steadily warming or cooling region can collapse to the same vector (Prop.~1, Appendix~B). We therefore add a parameter-free path summarizing each trajectory by four descriptors---mean $\mu_n$, standard deviation $\sigma_n$, signed slope $s_n$, and mean successive change $\delta_n$---added residually, $R_n=\mathrm{LN}(\bar Z_n+\mathrm{MLP}[\mu_n;\sigma_n;s_n;\delta_n])$. The signed slope $s_n$ is antisymmetric under time reversal, supplying exactly the order-sensitive signal attention discards.

\paragraph{Spatial context (R3).} A region is informative only relative to the body: a warm cheek beside a warm nose reads differently than beside a cool one. A single transformer layer therefore lets regions attend to one another before pooling, $H=\mathrm{LN}(\tilde R+\mathrm{FFN}(\tilde R))$ with $\tilde R=\mathrm{LN}(R+\mathrm{MHSA}(R))$; residual connections preserve local evidence when context is uninformative.

\paragraph{Gated MIL pooling (R1).} The weak label $y_i$ names the window, not the region, so the model must learn to assign credit across regions, the move that makes this MIL. For encoder $e$, gated attention \citep{ilse2018abmil} computes region weights $a^{(e)}_n=\mathrm{softmax}_n(w^\top(\tanh(Vh^{(e)}_n)\odot\sigma(Uh^{(e)}_n)))$ over the region embeddings $h^{(e)}_n$ (the rows of $H$) and pools them into the branch embedding $b^{(e)}=\sum_n a^{(e)}_n h^{(e)}_n$. The weights $a^{(e)}_n$ are non-negative and sum to one over the $N$ regions. A small per-encoder head then maps $b^{(e)}$ to a branch logit $\ell^{(e)}$ (supervised directly) and a scalar routing cue $r^{(e)}$ the fusion gate reads. We treat $a^{(e)}_n$ as pooling coefficients, not causal explanations (SHAP analysis in Appendix~I).

\subsection{Multi-Encoder Fusion}
\paragraph{One product bag (D3).} A naive ensemble would train $E$ independent MIL models, one per encoder, and combine their $E$ predictions after the fact. \method instead performs \emph{one} pooling operation over a single bag containing every (encoder, region) pair, so the encoders are fused as views of one problem rather than as separate models. A per-window gate fuses the $E$ branch embeddings $b^{(e)}$, $b^\star=\sum_e \pi_e(x)\,b^{(e)}$, where the gate weights $\pi(x)=\mathrm{softmax}_e\,\mathrm{Gate}([b^{(e)};r^{(e)}])$ score each encoder from its own branch embedding $b^{(e)}$ and routing cue $r^{(e)}$, and sum to one over the $E$ encoders. Substituting $b^{(e)}=\sum_n a^{(e)}_n h^{(e)}_n$ gives
\begin{equation}
b^\star=\sum_{e=1}^{E}\sum_{n=1}^{N} c_{e,n}(x)\,h^{(e)}_n,
\qquad c_{e,n}(x):=\pi_e(x)\, a^{(e)}_n,
\label{eq:productbag}
\end{equation}
where the combined weights are non-negative and sum to one over all $EN$ instances, $c_{e,n}(x)\ge0$ and $\sum_{e,n}c_{e,n}(x)=1$. Equation~\eqref{eq:productbag} is therefore a single attention-MIL pooling over the product bag of $EN$ (view, region) instances, not $E$ separate poolings combined afterward (Prop.~2, Appendix~B). The weights factorize as $c_{e,n}=\pi_e(x)\,a^{(e)}_n$: the region scorer $a^{(e)}_n$ decides which regions carry evidence, while the per-window gate $\pi_e(x)$ decides which encoder to trust. Because the gate is recomputed for each window, the model reallocates weight across encoders window by window, which is the source of the fusion gain (Section~\ref{sec:arch}). The cue $r^{(e)}$ enters only through $\mathrm{Gate}(\cdot)$, so its effect is already carried by $\pi_e(x)$; it is an uncalibrated softplus scalar used for routing, not a confidence or uncertainty estimate.

\paragraph{Why sharing is required (D2).} Adding branch embeddings in Eq.~\eqref{eq:productbag} is meaningful only if a channel means the same thing across encoders. But embedding coordinates can be arbitrary: permuting a branch's channels leaves its own prediction unchanged, since its head can absorb the permutation. Two branches may therefore settle on different channel orderings, and adding embeddings across different orderings is like summing vectors in mismatched coordinate systems: the result has no stable meaning. Sharing the aggregator pins all branches to one ordering: \emph{untied operators leave the fused embedding identifiable only up to an independent per-branch reordering ($\mathcal{S}_d^{\,E}$); sharing collapses this to a single reordering ($\mathcal{S}_d$) the final head absorbs} (Prop.~3, Appendix~B). This is a statement about well-posedness, and it is testable: untying the operator causes a large accuracy drop (Section~\ref{sec:arch}).

\paragraph{Why embedding-level (Obs.~4).} Of the three fusion depths, \emph{feature-level} concatenation forces one aggregator to reconcile incompatible geometries, and \emph{posterior-level} fusion depends only on the $E$ branch logits, where two windows with identical logits are indistinguishable to it regardless of their spatial evidence (Obs.~4, Appendix~B). Embedding-level fusion pools each encoder's regions first, then fuses: late enough to keep each representation, early enough to fuse structured evidence rather than a single score. Section~\ref{sec:arch} measures both limits.

\subsection{Training}
Only the aggregator is trained, minimizing a sum of six terms (equations and details in Appendix~B), each targeting a specific failure mode, briefly outlined below.
{\small
\begin{equation*}
\mathcal{L}=\mathcal{L}_{\text{task}}
 +\lambda_{\text{ds}}\mathcal{L}_{\text{ds}}
 +\lambda_{\text{cons}}\mathcal{L}_{\text{cons}}
 +\lambda_{\text{ent}}(k)\mathcal{L}_{\text{ent}}
 +\lambda_{\text{gate}}\mathcal{L}_{\text{gate}}
 +\mathcal{L}_{\text{adv}}.
\end{equation*}
}

 \emph{\textbf{Task loss}} ($\mathcal{L}_{\text{task}}$): focal loss with label smoothing on a mixup of $b^\star$; mixup is applied \emph{after} pooling, since mixing raw thermograms would blend bodies and masks into implausible frames.
\emph{\textbf{Deep supervision}} ($\mathcal{L}_{\text{ds}}$): supervises every branch logit $\ell^{(e)}$, so each encoder stays predictive rather than letting the gate ride one dominant branch.
\emph{\textbf{Consistency}} ($\mathcal{L}_{\text{cons}}$): ties branch probabilities to their mean, so encoders agree on the shared event instead of fitting view-specific noise.
\emph{\textbf{Entropy band}} ($\mathcal{L}_{\text{ent}}$, novel): a \emph{two-sided} penalty pinning attention entropy near an intermediate target $\rho\log N$ rather than merely maximizing or minimizing it. This preserves \emph{how much} spatial evidence is kept: a one-cell collapse is fragile to pose and mask error, while diffuse attention reverts to the global averaging we set out to avoid, and the band holds the model between the two.
\emph{\textbf{Gate KL}} ($\mathcal{L}_{\text{gate}}$): pulls the fusion gate toward uniform, so an encoder is trusted over others only when the evidence earns it.
\emph{\textbf{Subject adversary}} ($\mathcal{L}_{\text{adv}}$): a gradient-reversal identity classifier on $b^\star$ that blocks identity shortcuts under person-disjoint evaluation; ramped in gradually and active only in Control$\rightarrow$OUD.

Loss weights are selected on validation rather than factorially ablated, so the experiments support the complete system.

\paragraph{Participant centering.} Because baseline body temperature varies across people, features are centered by a label-free per-participant mean, so the classifier responds to each person's \emph{deviations} rather than absolute temperature. The mean is estimated from up to $24$ unlabeled windows, including at test time, making the method transductive: \method requires a short ($\approx$1.5-min) calibration period per participant and is \textbf{not} a cold-start detector. Classification thresholds are selected on the validation participants only.

\section{Evaluation Design}
\label{sec:eval}
The evaluation used different datasets to test different claims, rather than relying on one result to support population, dataset, and construct generalization simultaneously.

\textbf{RQ1 (person/cohort transfer):} Does regional aggregation generalize to unseen participants, and what is lost when a model trained only on Control participants is applied to OUD? \textbf{RQ2 (external transfer):} Does the architecture generalize to the public StressNet dataset despite differences in stressor, sensor, spatial coverage, and grid size? \textbf{RQ3 (endpoint transfer):} Does the learned representation retain useful information for craving when supervision changes from task labels to self-report?---Answering these needs three datasets: our own cohort-structured corpus (RQ1 and RQ3) and the public StressNet corpus (RQ2). Datasets, protocols, baselines, and metrics are outlined below:

\subsection{Datasets: cohort-structured thermal corpus}
\label{subsec:datacollection}
We collected the primary corpus (Dataset~I) because no public dataset pairs contactless thermal video of a clinical OUD cohort with a matched control group under a shared protocol. Under IRB approval $\#$ we recorded \textbf{67 participants} across \textbf{71 sessions}: \textbf{42 control} adults and \textbf{25 individuals with opioid use disorder (OUD)}, four of whom were recorded twice (giving 29 OUD sessions). Because splits, windows, and results are defined at the session level, we report the $42$/$29$ session counts throughout; the four repeated participants' sessions are used only for training, so no individual appears in both train and test. The OUD cohort spans different medication and treatment-phase states, reflecting naturally occurring variations. Data collection followed an established protocol from \citep{xiao2024reading}. Sessions ran in a private room; a \emph{fixed thermal camera} recorded the participant. Stress is induced by task structure (non-stress baselines vs.\ validated Stroop, sing-a-song, and mental-arithmetic blocks) \citep{kirschbaum1993trier,xiao2024reading}; separately, OUD participants self-report craving after each task block, giving the independent RQ3 target---so stress is never a proxy for craving, and craving occurs under both block types (Appendix~\ref{data-charact}). Full recruitment, apparatus, task sequence, and annotation protocol are in Appendix~C.

\textbf{Dataset I: the resulting corpus.}
Windowing yields $16{,}273$ windows (Control $11{,}410$; OUD $4{,}863$) under the single protocol above. Labels are $77.5\%$ stress, so the accuracy floor is $\approx0.775$. For RQ1, the stress prior is a property of the protocol, not of the cohort (Control $76.7\%$, OUD $79.2\%$), so the cohort gap cannot be explained by label composition.

\textbf{Protocols.} Four person-disjoint settings. Three share one group-stratified split (train $47$ / val $10$ / test $14$) and are evaluated on \textbf{Test-Both} ($3{,}306$ windows), \textbf{Test-Control} ($8$ participants, $2{,}306$), and \textbf{Test-OUD} ($6$, $1{,}000$). The fourth, \textbf{Control$\rightarrow$OUD}, trains and validates on Control only ($36$/$6$) and tests on all $29$ OUD sessions ($25$ individuals; $4{,}863$ windows). Splits are fixed and shared by every method.

\textbf{Datasets II and III.} StressNet \citep{kumar2021stressnet} supplies $5{,}054$ protocol-filtered windows over a subject-disjoint $26$/$5$/$3$ fold; Dataset III reuses the $4{,}863$ OUD windows with a task-block craving rating replacing the stress target ($55.5\%$ craving) over a $21$/$4$/$4$ split. Both are bounded tests (Sec.~\ref{sec:threats}).

\textbf{Baselines and metrics.}
Twenty-seven methods on identical splits and three seeds span six families---linear probes, LoRA, weight-space ensembles, learned multimodal fusion, fixed posterior combination rules, and non-deep tabular ensembles (full list and citations in Appendix~D). Three published MIL aggregators \citep{ilse2018abmil,shao2021transmil,li2021dsmil} are reserved for Sec.~\ref{sec:arch}. AUROC is primary because it is threshold-free; F1 is reported, and accuracy is read against the $\approx0.775$ floor \citep{brodersen2010balanced}.

\paragraph{Uncertainty and significance protocol.}\label{par:uncertainty}
Because our claims concern transfer to unseen \emph{people}, completed sampling analyses resample the \emph{participant}, not the window, using a participant-level cluster bootstrap ($B{=}2000$). Comparisons reuse the same participant draws across paired terms and report percentile $95\%$ CIs and one-sided empirical bootstrap tail probabilities; variation across training seeds is reported separately as seed SD. A family-wide FDR-corrected comparison against every baseline has not been run and is not claimed. Full procedure in Appendix~D.


\section{Results}\label{sec:results}
\subsection{RQ1: person and cohort transfer}

On held-out participants from both OUD and control cohorts in Dataset I, \method reaches $0.938$ AUROC--- $0.203$ above the strongest of 27 general baselines (TMC, $0.735$), and above the three published MIL aggregators too, which peak at $0.736$ on Test-Both and $0.679$ on Control$\rightarrow$OUD (analyzed separately in Section~\ref{sec:arch}). On the harder Control$\rightarrow$OUD split, \method still reaches $0.771$. Appendix~D (Table~\ref{tab:main}) reports these margins in full; the paired-participant significance analysis is below.

Two patterns common in health applications appear here. First, ranking and operating point come apart: Random Forest hits $0.802$ accuracy but ranks at only $0.612$ AUROC, riding the $77.5\%$ stress prior rather than truly separating the classes. Second, adapting the encoder does not help when labels are scarce: LoRA never exceeds $0.686$ (Appendix~D).

\subsection{RQ1, continued: decomposing the cohort gap}
\label{sec:gap}

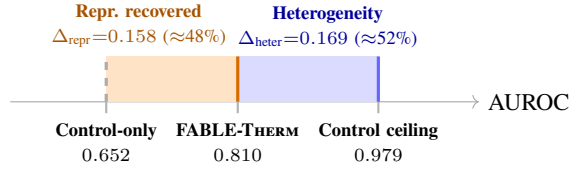
\begin{figure}[]
\centering
\begin{tikzpicture}[font=\small]
\def\ya{0.7}\def\yb{1.3}                        
\def\xctrl{1.57}\def\xoud{3.31}\def\xceil{5.17} 

\draw[->,gray!70] (0.3,\ya) -- (6.5,\ya) node[right,black] {AUROC};

\fill[blue!14] (\xoud,\ya) rectangle (\xceil,\yb);
\fill[orange!22] (\xctrl,\ya) rectangle (\xoud,\yb);

\draw[very thick,blue!60] (\xceil,\ya)--(\xceil,\yb);
\draw[very thick,orange!85!black] (\xoud,\ya)--(\xoud,\yb);
\draw[very thick,dashed,gray!65] (\xctrl,\ya)--(\xctrl,\yb);

\draw[gray!60,thick] (\xctrl,\ya) -- ++(0,-0.15)
  node[below,align=center,text width=2.0cm,text=black,font=\scriptsize] {\textbf{Control-only}\\[-0.2ex]$0.652$};
  
\draw[gray!60,thick] (\xoud,\ya) -- ++(0,-0.15)
  node[below,align=center,text width=1.8cm,text=black,font=\scriptsize] {\textbf{\method}\\[-0.2ex]$0.810$};
  
\draw[gray!60,thick] (\xceil,\ya) -- ++(0,-0.15)
  node[below,align=center,text width=2.1cm,text=black,font=\scriptsize] {\textbf{Control ceiling}\\[-0.2ex]$0.979$};

\node[above,align=center,orange!65!black,text width=2.8cm, font=\scriptsize]
  at (2.0,\yb)
  {\textbf{Repr. recovered}\\[-0.2ex]$\Delta_{\text{repr}}{=}0.158$ ($\approx$48\%)};

\node[above,align=center,blue!60!black,text width=2.8cm, font=\scriptsize]
  at (4.5,\yb)
  {\textbf{Heterogeneity}\\[-0.2ex]$\Delta_{\text{heter}}{=}0.169$ ($\approx$52\%)};

\end{tikzpicture}
\caption{\textbf{The cohort gap, decomposed} (higher AUROC is better). For the \emph{same} six held-out OUD participants, adding OUD data raises AUROC from $0.652$ to $0.810$ (\emph{orange}, $\Delta_{\text{repr}}{=}0.158$), leaving a comparable gap to the $0.979$ Control ceiling (\emph{blue}, $\Delta_{\text{heter}}{=}0.169$). The two terms account for roughly $48\%$ and $52\%$ of the total deficit, showing that heterogeneity is a co-equal barrier. }
\label{fig:decomp}
\end{figure}

Aggregate scores hide the finding that matters, and \method makes it reachable: because the same architecture trains with or without the OUD cohort while still scoring each held-out participant, we can decompose \emph{why} it underperforms on OUD, not merely report \emph{that} it does---an attribution a global representation could not support. On held-out OUD participants \method scores $0.810$ AUROC, well below $0.979$ on Control. Figure~\ref{fig:decomp} splits that $0.169$ gap into two causes: \emph{representation} (does the model fail simply because it never trained on OUD physiology?---if so, more data closes the gap) and \emph{heterogeneity} (does a model that \emph{did} train on OUD still fail to generalize person-to-person?---if so, more data of the same kind will not help).

\paragraph{A matched decomposition.}
To isolate representation without confounding it with a change of test set, we
evaluate \emph{one fixed} set---the six Test-OUD participants---under two regimes:
the model trained \emph{with} OUD ($0.810$) and the Control-only model applied to
those same six without retraining
($\text{AUROC}_{\text{ctrl-only}@6}=0.652\pmstd{.011}$). The drop
between them is what representation buys, on one population; the heterogeneity term
is the distance from the Control ceiling that persists even with OUD in training.
\begin{align}
\Delta_{\text{repr}} &= 0.810-0.652 = 0.158
   \;(\text{matched, }n{=}6) \nonumber\\
\Delta_{\text{heter}} &= 0.979-0.810 = 0.169 \;(\text{Control ceiling}-\text{OUD}) \nonumber
\end{align}
$\Delta_{\text{repr}}$ is the part including the cohort \emph{removes};
$\Delta_{\text{heter}}$ is the part it \emph{does not touch}. We find the two
terms comparable---$0.158$ versus $0.169$, roughly a $48$/$52$ split of the total
deficit---so the deficit is not simply an absence of OUD data: within-cohort
variability is a comparably sized contributor. Testing on the shared
six-participant estimate by participant bootstrap ($B{=}2000$), heterogeneity
exceeds representation in $61.45\%$ of resamples. On the aligned seed-$43$
prediction triplet, the paired difference is
$\Delta_{\text{heter}}-\Delta_{\text{repr}}=0.021$, with $95\%$ CI
$[-0.076,0.200]$ and one-sided $p=0.386$. Thus the terms are comparable, but
their ordering is not statistically resolved. The component intervals are
$[0.099,0.229]$ for $\Delta_{\text{heter}}$ and
$[-0.015,0.200]$ for $\Delta_{\text{repr}}$; these sampling intervals are
seed-$43$ estimates and are kept distinct from the three-seed means above. The full per-protocol breakdown is in Appendix~D (Table~\ref{tab:gap}).

The implication is uncomfortable for a common equity remedy: ``collecting more data from the underserved group'' is necessary but recovers only about half the deficit, the remainder reflecting between-person variation \emph{within} the OUD cohort (plausibly medication, comorbidity, thermoregulation, etc.; none measured here). Deployments relying only on additional OUD data will therefore under-deliver and should also support per-person calibration. \emph{More broadly, the decomposition is modality-agnostic}: any model with an identifiable subpopulation and a held-out cohort can compute the same representation-versus-heterogeneity split. It turns ``does sampling the underserved group help?'' from an assumption into a measurable quantity, which we argue should be reported whenever equity claims rest on data collection.



\subsection{Ablating: what each design choice buys}
\label{sec:arch}
This is the controlled comparison that carries the architectural claims: holding the frozen front-end, splits, and seeds fixed and varying \emph{only} the aggregator, so any difference is attributable to the design choice rather than to features or data. Every configuration below sees identical inputs from Dataset~I. Table~\ref{tab:mil} summarizes the head-to-head, and Table~\ref{tab:abl_full} gives the per-encoder breakdown; we read the key rows off in turn (from Appendix~G).

\paragraph{The branch helps before fusion.}
A single-encoder \method branch reaches $0.747$ AUROC (ImageNet-ViT) and $0.727$ (AnyThermal), exceeding ABMIL, DSMIL, and TransMIL on the same features; so even before ensembling, treating instances as trajectories beats treating them as points (D1). Grouped Kernel-SHAP (Appendix~I) confirms localized trajectories carry the evidence, moving the logit $4.76\times$ more than random trajectories.

\paragraph{Provenance should survive to the decision stage.}
We compare the three fusion depths on identical features, splits, and seeds.
\emph{Feature-level} fusion (Concat-3, one aggregator over concatenated encoders)
tops out at $0.736$ on Test-Both (Table~\ref{tab:abl_full}, best Concat-3);
\emph{posterior-level} fusion (learned combination of per-encoder logits) tops out
at $0.735$, realized by the TMC baseline in the main sweep
(Table~\ref{tab:main}); and \method's \emph{embedding-level} fusion
with the learned gate reaches $0.938$ (Table~\ref{tab:mil}), also lifting the best
non-embedding Control$\rightarrow$OUD result from $0.679$ to $0.771$
(Table~\ref{tab:abl_full}, final row). Every feature- and posterior-level variant
falls between $0.66$ and $0.74$ on Test-Both while embedding-level fusion exceeds
$0.93$, showing that fusion \emph{depth}, not encoder choice, is the operative
variable (Observation~4). The mechanism is the learned per-window gate reallocating
credit across encoders (Prop.~2(ii)), a flexibility unavailable to fixed
combination rules or single-view aggregators.

\paragraph{The shared operator is necessary, not convenient (Prop.~3, measured).}
The untied competitor gives each encoder its own temporal, spatial, and gated-MIL
operators, then applies learned attention to the three branch logits, using
$1.80\times$ the trainable parameters of the shared model
(compared in Table~\ref{tab:mil}). Despite the extra capacity, it reaches
only $0.744\pmstd{.012}$ AUROC on Test-Both and $0.674\pmstd{.053}$ on Test-OUD,
against $0.938\pmstd{.004}$ and $0.810\pmstd{.002}$ for \method---a collapse of
$0.194$ and $0.136$ AUROC. Its seed SD is also markedly larger---$3\times$ on
Test-Both ($.012$ vs.\ $.004$) and over $20\times$ on Test-OUD; the instability Proposition~3 predicts when the fused
embedding is pinned down only up to an independent per-branch relabeling. \emph{The gauge collapse of Proposition~3 is therefore not cosmetic: sharing the operator is what makes embedding-level fusion well-posed}, and untying it, even with more parameters, does not recover the performance. This is not thermal-specific: any system that fuses embeddings from multiple frozen FMs (multi-encoder pathology, multimodal medical fusion, remote sensing) faces the same per-branch relabeling ambiguity, and the shared-operator construction is a general remedy.



\subsection{RQ2: external transfer (StressNet)}
Does the \emph{architecture} generalize to a second dataset, retrained on its own subject-disjoint split? On the public StressNet corpus \method averages $0.808\pm0.112$ AUROC across three subject-disjoint folds (we report the average, not a hand-picked best fold). The wide spread ($\pm0.112$) is the lesson of Sec.~\ref{sec:gap} reappearing: which people land in the test fold moves the score more than the method does. We claim only that the pipeline \emph{operates} under a second sensor, stressor, and grid resolution ($N{=}97$; Appendix~E), not a new benchmark.

\subsection{RQ3: endpoint transfer to craving}
Substituting a self-reported craving rating for the stress target on the 29 OUD sessions ($25$ individuals), \method attains $0.752$ AUROC, beating the strongest craving baseline (MIMMO, a multi-input multi-output ensemble \citep{ferianc2023mimmo}). To our knowledge, this is the \emph{first evidence} that OUD craving is recoverable from contactless thermal video at all. We frame it as a feasibility probe, not a detector: those $4{,}863$ windows carry just $172$ independent labels and $16$ of $29$ sessions are single-class, so the supportable claim is that a body-localized representation \emph{carries} craving-associated signal (Appendix~F).

\section{Threats to Validity}
\label{sec:threats}
We bound each claim rather than collect caveats at the end. \emph{Ablation scope:} the headline comparison is system-level (\method uses an Optuna-tuned recipe, most baselines a vanilla loop); Table~\ref{tab:mil} isolates the aggregator and measures the shared-operator choice directly, but the statistics path, spatial attention, and auxiliary losses are argued, not each ablated (Appendix~G.1). \emph{Small samples:} with $6$--$29$ sessions per protocol, bootstrap intervals are wide, so we report uncertainty explicitly and do not lean on unresolved effect ordering. \emph{Bounded external tests:} StressNet is retrained on its own split (RQ2) and craving reuses the OUD windows with few independent labels (RQ3), so both are feasibility checks, not benchmarks. \emph{Normalized appearance:} per-region scaling removes absolute temperature, so the model learns appearance, not calibrated units. \emph{Label semantics:} stress labels are elicitation conditions, not verified states. \emph{Calibration:} participant centering is transductive, so \method is not a cold-start detector. \emph{Unaudited subgroups:} cohort tests do not replace audits across age, sex, skin properties, disability, and medication. \emph{Discrimination, not benefit:} we measure accuracy, not whether acting on a prediction helps anyone.

\section{Responsible Deployment and Release}
\label{sec:ethics}

\paragraph{Intended deployment.}
The intended role is opt-in clinical decision support: a revocable check-in, not diagnosis, substance-use inference, or a replacement for clinical judgment. Thermal video is body-derived and person-linkable, not anonymous \citep{stanic2025review}, and the validity of affective inference remains contested even for facial images \citep{mohammad2022ethics}; accordingly, we predict elicitation conditions rather than ``true'' internal states, and such signals increasingly fall under cognitive-biometric protections \citep{ienca2024cognitive}. Deployment assumes a short participant-specific calibration period because the method is transductive and therefore not a cold-start detector. It also requires cohort-specific operating thresholds rather than a universal $0.5$, since decisions are made at a threshold, not by AUROC. At our validation-selected threshold, OUD bears the greater operational burden: FPR is $0.495$ versus $0.090$ for Control ($5.5\times$ higher), and FNR is $0.144$ versus $0.073$ (Table~\ref{tab:oppoint}), the deployment-facing manifestation of $\Delta_{\text{heter}}$. Every elevated prediction should be reviewed by a clinician, never acted on automatically. Because roughly half of the OUD deficit reflects within-cohort heterogeneity (Sec.~\ref{sec:gap}), collecting more data alone will not remove the disparity; deployments without participant-specific calibration are unlikely to perform reliably for this population. The reported cohort-specific operating points provide the quantities needed to estimate burden and risk before a clinical study.

\textbf{Prohibited uses.}
The method must not be used for surveillance, coercive decision-making, or non-consensual inference, including law enforcement, probation or parole, employment, insurance, and benefits or treatment eligibility. These restrictions are enforced through the artifact license, not left as recommendations. Continuous stress or craving inference can become a surveillance tool for a population already subject to supervisory drug testing \citep{cheetham2022stigma}, and because reducing sensing burden also lowers the barrier to misuse, deployment safeguards must accompany the released artifact rather than be left to downstream users.

\textbf{Release and governance.}
We publicly release all code, derived frozen representations, training and evaluation scripts, fixed splits, per-seed predictions, the attribution checkpoint, and a machine-readable non-use license---enough to reproduce every result without the raw video and, because StressNet is public, the external-transfer experiment end to end. Raw clinical recordings remain protected under an IRB-governed data-use agreement. Reproduction and sensor-porting instructions are in Appendix~\ref{app:repro}.

\section{Conclusion}
Contactless sensing removes not only the wearable but also the supervision indicating \emph{where} and \emph{when} a physiological response occurs. \method shows that preserving localized evidence turns this loss into an opportunity: the representation becomes not only more accurate but also \emph{interrogable}. Because performance remains participant-resolved, deployment failures become questions rather than statistics. On our cohort, improving representation alone recovers only about half of the deployment gap, with the remainder reflecting within-cohort heterogeneity. More broadly, whenever an identifiable subpopulation exists, ``collect more data'' is a hypothesis rather than a remedy, one any model can test by training with and without that group, but only localized evidence explains \emph{who} a model fails and \emph{why}, and that is what equitable deployment requires.

\newpage


{\small
\bibliography{references-v1}

\begin{thebibliography}{80}
\providecommand{\natexlab}[1]{#1}

\bibitem[{Abdelrahman et~al.(2017)}]{abdelrahman2017cognitive}
Abdelrahman, Y.; et~al. 2017.
\newblock Cognitive heat: Exploring the usage of thermal imaging to unobtrusively estimate cognitive load.
\newblock \emph{Proceedings of the ACM on Interactive, Mobile, Wearable and Ubiquitous Technologies}, 1(3): 1--20.

\bibitem[{Akiba et~al.(2019)}]{akiba2019optuna}
Akiba, T.; et~al. 2019.
\newblock Optuna: A next-generation hyperparameter optimization framework.
\newblock In \emph{Proceedings of the 25th ACM SIGKDD International Conference on Knowledge Discovery \& Data Mining}, 2623--2631.

\bibitem[{Al-Jebrni et~al.(2020)}]{aljebrni2020stress}
Al-Jebrni, A.~H.; et~al. 2020.
\newblock {AI}-enabled remote and objective quantification of stress at scale.
\newblock \emph{Biomedical Signal Processing and Control}, 59: 101929.

\bibitem[{Anderson et~al.(2018)}]{anderson2018ethical}
Anderson, B.~J.; et~al. 2018.
\newblock Ethical considerations in opioid craving induction research.
\newblock \emph{Journal of Addiction Medicine}, 12(4): 253--259.

\bibitem[{Baillet et~al.(2024)}]{baillet2024craving}
Baillet, E.; et~al. 2024.
\newblock Craving changes in first 14 days of addiction treatment: an outcome predictor of 5 years substance use status?
\newblock \emph{Translational Psychiatry}, 14(1): 497.

\bibitem[{Breiman(2001)}]{breiman2001randomforest}
Breiman, L. 2001.
\newblock Random forests.
\newblock \emph{Machine Learning}, 45(1): 5--32.

\bibitem[{Brodersen et~al.(2010)}]{brodersen2010balanced}
Brodersen, K.~H.; et~al. 2010.
\newblock The balanced accuracy and its posterior distribution.
\newblock \emph{Proceedings of the 20th International Conference on Pattern Recognition (ICPR)}, 3121--3124.

\bibitem[{Br{\"u}gge et~al.(2026)}]{brugge2026taskaware}
Br{\"u}gge, N.~S.; et~al. 2026.
\newblock Task-aware multiple instance learning for stress detection from facial video data.
\newblock \emph{Journal of Affective Disorders}, 405: 121472.

\bibitem[{Calderon-Uribe et~al.(2026)}]{calderon2026peripheral}
Calderon-Uribe, S.; et~al. 2026.
\newblock Peripheral thermoregulation patterns of stress and relaxation assessed via infrared thermography and linear discriminant analysis.
\newblock \emph{Journal of Thermal Biology}, 139: 104507.

\bibitem[{Campanella et~al.(2019)}]{campanella2019clinical}
Campanella, G.; et~al. 2019.
\newblock Clinical-grade computational pathology using weakly supervised deep learning on whole slide images.
\newblock \emph{Nature Medicine}, 25(8): 1301--1309.

\bibitem[{Cardone and Merla(2017)}]{cardone2015thermal}
Cardone, D.; and Merla, A. 2017.
\newblock New frontiers for applications of thermal infrared imaging devices: Computational psychophysiology in the neurosciences.
\newblock \emph{Sensors}, 17(5): 1042.

\bibitem[{Cheetham et~al.(2022)}]{cheetham2022stigma}
Cheetham, A.; et~al. 2022.
\newblock The impact of stigma on people with opioid use disorder, opioid treatment, and policy.
\newblock \emph{Substance Abuse and Rehabilitation}, 13: 1--12.

\bibitem[{Chen et~al.(2021)}]{chen2021ethical}
Chen, I.~Y.; et~al. 2021.
\newblock Ethical machine learning in healthcare.
\newblock \emph{Annual Review of Biomedical Data Science}, 4: 123--144.

\bibitem[{Chen and Guestrin(2016)}]{chen2016xgboost}
Chen, T.; and Guestrin, C. 2016.
\newblock {XGBoost}: A scalable tree boosting system.
\newblock In \emph{Proceedings of the 22nd ACM SIGKDD International Conference on Knowledge Discovery and Data Mining (KDD)}, 785--794.

\bibitem[{Di~Giacinto et~al.(2014)}]{digiacinto2014fear}
Di~Giacinto, A.; et~al. 2014.
\newblock Thermal signature of fear conditioning in mild post-traumatic stress disorder.
\newblock \emph{Neuroscience}, 266: 216--223.

\bibitem[{Dosovitskiy et~al.(2021)}]{dosovitskiy2021vit}
Dosovitskiy, A.; et~al. 2021.
\newblock An image is worth 16x16 words: Transformers for image recognition at scale.
\newblock In \emph{International Conference on Learning Representations (ICLR)}.

\bibitem[{Engert et~al.(2014)}]{engert2014stress}
Engert, V.; et~al. 2014.
\newblock Exploring the use of thermal infrared imaging in human stress research.
\newblock \emph{PLOS ONE}, 9(3): e90782.

\bibitem[{Ferianc and Rodrigues(2023)}]{ferianc2023mimmo}
Ferianc, M.; and Rodrigues, M. 2023.
\newblock {MIMMO}: Multi-input massive multi-output neural network.
\newblock In \emph{Proceedings of the IEEE/CVF Conference on Computer Vision and Pattern Recognition Workshops (CVPRW)}.

\bibitem[{Fox et~al.(2008)}]{fox2008enhanced}
Fox, H.~C.; et~al. 2008.
\newblock Enhanced sensitivity to stress and drug/alcohol craving in abstinent cocaine-dependent individuals compared to social drinkers.
\newblock \emph{Neuropsychopharmacology}, 33(4): 796--805.

\bibitem[{Ganin et~al.(2016)}]{ganin2016dann}
Ganin, Y.; et~al. 2016.
\newblock Domain-adversarial training of neural networks.
\newblock \emph{Journal of Machine Learning Research}, 17(59): 1--35.

\bibitem[{Garbey et~al.(2007)}]{garbey2007contact}
Garbey, M.; et~al. 2007.
\newblock Contact-free measurement of cardiac pulse based on the analysis of thermal imagery.
\newblock \emph{IEEE Transactions on Biomedical Engineering}, 54(8): 1418--1426.

\bibitem[{Garipov et~al.(2018)}]{garipov2018fge}
Garipov, T.; et~al. 2018.
\newblock Loss surfaces, mode connectivity, and fast ensembling of {DNNs}.
\newblock In \emph{Advances in Neural Information Processing Systems (NeurIPS)}.

\bibitem[{Geurts, Ernst, and Wehenkel(2006)}]{geurts2006extratrees}
Geurts, P.; Ernst, D.; and Wehenkel, L. 2006.
\newblock Extremely randomized trees.
\newblock \emph{Machine Learning}, 63(1): 3--42.

\bibitem[{Ghassemi et~al.(2020)}]{ghassemi2020review}
Ghassemi, M.; et~al. 2020.
\newblock A review of challenges and opportunities in machine learning for health.
\newblock \emph{AMIA Summits on Translational Science Proceedings}, 2020: 191.

\bibitem[{Giannakakis et~al.(2022)}]{giannakakis2019review}
Giannakakis, G.; et~al. 2022.
\newblock Review on psychological stress detection using biosignals.
\newblock \emph{IEEE Transactions on Affective Computing}, 13(1): 440--460.

\bibitem[{Gioia et~al.(2021)}]{gioia2021discriminating}
Gioia, F.; et~al. 2021.
\newblock Discriminating stress from cognitive load using contactless thermal imaging devices.
\newblock In \emph{43rd Annual International Conference of the IEEE Engineering in Medicine and Biology Society (EMBC)}, 608--611.

\bibitem[{Gioia et~al.(2022)}]{gioia2022contactless}
Gioia, F.; et~al. 2022.
\newblock Towards a contactless stress classification using thermal imaging.
\newblock \emph{Sensors}, 22(3): 976.

\bibitem[{Golden, Freshwater, and Golden(1978)}]{golden1978stroop}
Golden, C.; Freshwater, S.~M.; and Golden, Z. 1978.
\newblock Stroop color and word test.

\bibitem[{Grunevski, Kong, and Konova(2024)}]{grunevski2024predictive}
Grunevski, S.; Kong, J.; and Konova, A. 2024.
\newblock Predictive Relationship Between Different Timescales of Opioid Craving and Opioid Use.
\newblock \emph{Drug and Alcohol Dependence}, 260: 109974.

\bibitem[{Han et~al.(2024)}]{han2024fusemoe}
Han, X.; et~al. 2024.
\newblock {FuseMoE}: Mixture-of-experts transformers for fleximodal fusion.
\newblock In \emph{Advances in Neural Information Processing Systems (NeurIPS)}.

\bibitem[{Han et~al.(2021)}]{han2021tmc}
Han, Z.; et~al. 2021.
\newblock Trusted multi-view classification.
\newblock In \emph{International Conference on Learning Representations (ICLR)}.

\bibitem[{Huang et~al.(2017)}]{huang2017snapshot}
Huang, G.; et~al. 2017.
\newblock Snapshot ensembles: Train 1, get {M} for free.
\newblock In \emph{International Conference on Learning Representations (ICLR)}.

\bibitem[{Ienca and Malgieri(2024)}]{ienca2024cognitive}
Ienca, M.; and Malgieri, G. 2024.
\newblock Beyond neural data: Cognitive biometrics and mental privacy.
\newblock \emph{Neuron}, 112(18): 3017--3028.

\bibitem[{Ilse, Tomczak, and Welling(2018)}]{ilse2018abmil}
Ilse, M.; Tomczak, J.; and Welling, M. 2018.
\newblock Attention-based deep multiple instance learning.
\newblock In \emph{International Conference on Machine Learning (ICML)}, 2127--2136.

\bibitem[{Ioannou, Gallese, and Merla(2014)}]{ioannou2014thermal}
Ioannou, S.; Gallese, V.; and Merla, A. 2014.
\newblock Thermal infrared imaging in psychophysiology: Potentialities and limits.
\newblock \emph{Psychophysiology}, 51(10): 951--963.

\bibitem[{Izmailov et~al.(2018)}]{izmailov2018swa}
Izmailov, P.; et~al. 2018.
\newblock Averaging weights leads to wider optima and better generalization.
\newblock In \emph{Conference on Uncertainty in Artificial Intelligence (UAI)}.

\bibitem[{Jacobs et~al.(1991)}]{jacobs1991moe}
Jacobs, R.~A.; et~al. 1991.
\newblock Adaptive mixtures of local experts.
\newblock \emph{Neural Computation}, 3(1): 79--87.

\bibitem[{Ke et~al.(2017)}]{ke2017lightgbm}
Ke, G.; et~al. 2017.
\newblock {LightGBM}: A highly efficient gradient boosting decision tree.
\newblock In \emph{Advances in Neural Information Processing Systems (NeurIPS)}.

\bibitem[{Kirschbaum, Pirke, and Hellhammer(1993)}]{kirschbaum1993trier}
Kirschbaum, C.; Pirke, K.-M.; and Hellhammer, D.~H. 1993.
\newblock The `Trier Social Stress Test'--a tool for investigating psychobiological stress responses in a laboratory setting.
\newblock \emph{Neuropsychobiology}, 28(1-2): 76--81.

\bibitem[{Kittler et~al.(1998)}]{kittler1998combining}
Kittler, J.; et~al. 1998.
\newblock On combining classifiers.
\newblock \emph{IEEE Transactions on Pattern Analysis and Machine Intelligence}, 20(3): 226--239.

\bibitem[{Kosonogov et~al.(2017)Kosonogov, De~Zorzi, Honor{\'e}, Mart{\'i}nez-Vel{\'a}zquez, Nandrino, Martinez-Selva, and Sequeira}]{kosonogov2017arousal}
Kosonogov, V.; De~Zorzi, L.; Honor{\'e}, J.; Mart{\'i}nez-Vel{\'a}zquez, E.~S.; Nandrino, J.-L.; Martinez-Selva, J.~M.; and Sequeira, H. 2017.
\newblock Facial thermal variations: A new marker of emotional arousal.
\newblock \emph{PLOS ONE}, 12(9): e0183592.

\bibitem[{Kumar et~al.(2021)}]{kumar2021stressnet}
Kumar, S.; et~al. 2021.
\newblock {StressNet}: Detecting stress in thermal videos.
\newblock In \emph{Proceedings of the IEEE/CVF Winter Conference on Applications of Computer Vision (WACV)}, 999--1009.

\bibitem[{Kurata et~al.(2025)Kurata, Homma, Oka et~al.}]{kurata2025pathfm_mil}
Kurata, Y.; Homma, T.; Oka, K.; et~al. 2025.
\newblock Multiple instance learning using pathology foundation models effectively predicts kidney disease diagnosis and clinical classification.
\newblock \emph{Scientific Reports}, 15(1): 19567.

\bibitem[{Lakshminarayanan, Pritzel, and Blundell(2017)}]{lakshminarayanan2017deepensembles}
Lakshminarayanan, B.; Pritzel, A.; and Blundell, C. 2017.
\newblock Simple and scalable predictive uncertainty estimation using deep ensembles.
\newblock In \emph{Advances in Neural Information Processing Systems (NeurIPS)}.

\bibitem[{Li, Li, and Eliceiri(2021)}]{li2021dsmil}
Li, B.; Li, Y.; and Eliceiri, K.~W. 2021.
\newblock Dual-stream multiple instance learning network for whole slide image classification with self-supervised contrastive learning.
\newblock In \emph{Proceedings of the IEEE/CVF Conference on Computer Vision and Pattern Recognition (CVPR)}, 14318--14328.

\bibitem[{Lin et~al.(2017)}]{lin2017focal}
Lin, T.-Y.; et~al. 2017.
\newblock Focal loss for dense object detection.
\newblock In \emph{Proceedings of the IEEE International Conference on Computer Vision (ICCV)}, 2980--2988.

\bibitem[{Liu et~al.(2024{\natexlab{a}})}]{liu2024infmae}
Liu, F.; et~al. 2024{\natexlab{a}}.
\newblock {InfMAE}: A foundation model in the infrared modality.
\newblock In \emph{European Conference on Computer Vision (ECCV)}.

\bibitem[{Liu et~al.(2024{\natexlab{b}})}]{liu2024blush}
Liu, I.; et~al. 2024{\natexlab{b}}.
\newblock Your blush gives you away: Detecting hidden mental states with remote photoplethysmography and thermal imaging.
\newblock \emph{PeerJ Computer Science}, 10: e1912.

\bibitem[{Loshchilov and Hutter(2019)}]{loshchilov2019adamw}
Loshchilov, I.; and Hutter, F. 2019.
\newblock Decoupled weight decay regularization.
\newblock In \emph{International Conference on Learning Representations (ICLR)}.

\bibitem[{Lundberg and Lee(2017)}]{lundberg2017unified}
Lundberg, S.~M.; and Lee, S.-I. 2017.
\newblock A Unified Approach to Interpreting Model Predictions.
\newblock In \emph{Advances in Neural Information Processing Systems (NeurIPS)}, volume~30.

\bibitem[{Luo et~al.(2026)}]{luo2026personalized}
Luo, Y.; et~al. 2026.
\newblock Personalized entropy-informed deep learning for identifying opioid misuse.
\newblock \emph{Nature Mental Health}, 1--13.

\bibitem[{MacLean, Sofuoglu, and Rosenheck(2019)}]{maclean2019stress}
MacLean, R.~R.; Sofuoglu, M.; and Rosenheck, R. 2019.
\newblock Stress and opioid use disorder: A systematic review.
\newblock \emph{Addictive Behaviors}, 98: 106010.

\bibitem[{Maddox et~al.(2019)}]{maddox2019swag}
Maddox, W.~J.; et~al. 2019.
\newblock A simple baseline for {Bayesian} uncertainty in deep learning.
\newblock In \emph{Advances in Neural Information Processing Systems (NeurIPS)}.

\bibitem[{Maheshwari et~al.(2026)}]{maheshwari2026anythermal}
Maheshwari, R.; et~al. 2026.
\newblock {AnyThermal}: A task-agnostic vision foundation model for thermal images.
\newblock \emph{arXiv preprint arXiv:2601.18598}.

\bibitem[{Mohammad(2022)}]{mohammad2022ethics}
Mohammad, S.~M. 2022.
\newblock Ethics sheet for automatic emotion recognition and sentiment analysis.
\newblock \emph{Computational Linguistics}, 48(2): 239--278.

\bibitem[{Moon et~al.(2025)}]{moon2025brief}
Moon, S. J.~E.; et~al. 2025.
\newblock A brief report: Lessons learned using wearable technology to collect data from adults receiving medications for opioid use disorder.
\newblock \emph{Journal of Substance Use}, 30(6): 929--935.

\bibitem[{Nazzari et~al.(2025)}]{nazzari2025review}
Nazzari, S.; et~al. 2025.
\newblock Infrared thermal imaging ({ITI}), a non-invasive window into early emotion regulation: A systematic review.
\newblock \emph{Developmental Psychobiology}, 67(5): e70071.

\bibitem[{Obermeyer et~al.(2019)}]{obermeyer2019bias}
Obermeyer, Z.; et~al. 2019.
\newblock Dissecting racial bias in an algorithm used to manage the health of populations.
\newblock \emph{Science}, 366(6464): 447--453.

\bibitem[{Ollander et~al.(2016)}]{ollander2016comparison}
Ollander, S.; et~al. 2016.
\newblock A comparison of wearable and stationary sensors for stress detection.
\newblock In \emph{2016 IEEE International Conference on Systems, Man, and Cybernetics (SMC)}, 004362--004366. IEEE.

\bibitem[{Oquab et~al.(2024)}]{oquab2024dinov2}
Oquab, M.; et~al. 2024.
\newblock {DINOv2}: Learning robust visual features without supervision.
\newblock \emph{Transactions on Machine Learning Research (TMLR)}.

\bibitem[{Qiao, Chen, and Yuille(2021)}]{qiao2021detectors}
Qiao, S.; Chen, L.-C.; and Yuille, A. 2021.
\newblock {DetectoRS}: Detecting objects with recursive feature pyramid and switchable atrous convolution.
\newblock \emph{Proceedings of the IEEE/CVF Conference on Computer Vision and Pattern Recognition (CVPR)}, 10213--10224.

\bibitem[{Sabour et~al.(2021)Sabour, Benezeth, De~Oliveira, Chappe, and Yang}]{sabour2023ubfcphys}
Sabour, R.~M.; Benezeth, Y.; De~Oliveira, P.; Chappe, J.; and Yang, F. 2021.
\newblock Ubfc-phys: A multimodal database for psychophysiological studies of social stress.
\newblock \emph{IEEE Transactions on Affective Computing}, 14(1): 622--636.

\bibitem[{Schmidt et~al.(2018)}]{schmidt2018wesad}
Schmidt, P.; et~al. 2018.
\newblock Introducing {WESAD}, a multimodal dataset for wearable stress and affect detection.
\newblock In \emph{Proceedings of the 20th ACM International Conference on Multimodal Interaction (ICMI)}, 400--408.

\bibitem[{Shah, Kumari, and Jain(2024)}]{shah2024unveiling}
Shah, K.; Kumari, R.; and Jain, M. 2024.
\newblock Unveiling stress markers: A systematic review investigating psychological stress biomarkers.
\newblock \emph{Developmental Psychobiology}, 66(5): e22490.

\bibitem[{Shao et~al.(2021)}]{shao2021transmil}
Shao, Z.; et~al. 2021.
\newblock {TransMIL}: Transformer based correlated multiple instance learning for whole slide image classification.
\newblock In \emph{Advances in Neural Information Processing Systems (NeurIPS)}.

\bibitem[{Shastri et~al.(2012)}]{shastri2012perinasal}
Shastri, D.; et~al. 2012.
\newblock Perinasal imaging of physiological stress and its affective potential.
\newblock \emph{IEEE Transactions on Affective Computing}, 3(3): 366--378.

\bibitem[{Shatte, Hutchinson, and Teague(2019)}]{shatte2019digital}
Shatte, A. B.~R.; Hutchinson, D.~M.; and Teague, S.~J. 2019.
\newblock Machine learning in mental health: A scoping review of methods and applications.
\newblock \emph{Psychological Medicine}, 49(9): 1426--1448.

\bibitem[{Sinha(2009)}]{sinha2009modeling}
Sinha, R. 2009.
\newblock Modeling stress and drug craving in the laboratory: implications for addiction treatment development.
\newblock \emph{Addiction biology}, 14(1): 84--98.

\bibitem[{Sinha et~al.(2000)}]{sinha2000psychological}
Sinha, R.; et~al. 2000.
\newblock Psychological stress, drug-related cues and cocaine craving.
\newblock \emph{Psychopharmacology}, 152(2): 140--148.

\bibitem[{Stani{\'c} and Ger{\v{s}}ak(2025)}]{stanic2025review}
Stani{\'c}, V.; and Ger{\v{s}}ak, G. 2025.
\newblock Facial thermal imaging: A systematic review with guidelines and measurement uncertainty estimation.
\newblock \emph{Measurement}, 242: 115879.

\bibitem[{Theroude et~al.(2026)}]{theroude2026nose}
Theroude, P.; et~al. 2026.
\newblock The nose knows: Thermal responses to active psychological stressors.
\newblock \emph{PLOS ONE}, 21(1): e0338108.

\bibitem[{Vaezi~Joze et~al.(2020)}]{joze2020mmtm}
Vaezi~Joze, H.~R.; et~al. 2020.
\newblock {MMTM}: Multimodal transfer module for {CNN} fusion.
\newblock In \emph{Proceedings of the IEEE/CVF Conference on Computer Vision and Pattern Recognition (CVPR)}.

\bibitem[{Volkow, Koob, and McLellan(2016)}]{volkow2016neurobiology}
Volkow, N.~D.; Koob, G.~F.; and McLellan, A.~T. 2016.
\newblock Neurobiologic advances from the brain disease model of addiction.
\newblock \emph{New England Journal of Medicine}, 374(4): 363--371.

\bibitem[{Wen, Tran, and Ba(2020)}]{wen2020batchensemble}
Wen, Y.; Tran, D.; and Ba, J. 2020.
\newblock {BatchEnsemble}: An alternative approach to efficient ensemble and lifelong learning.
\newblock In \emph{International Conference on Learning Representations (ICLR)}.

\bibitem[{Xiao et~al.(2023)}]{xiao2024reading}
Xiao, Y.; et~al. 2023.
\newblock Reading between the heat: Co-teaching body thermal signatures for non-intrusive stress detection.
\newblock \emph{Proceedings of the ACM on Interactive, Mobile, Wearable and Ubiquitous Technologies}, 7(4): 1--30.

\bibitem[{Xiao et~al.(2025)}]{xiao2025hhiss}
Xiao, Y.; et~al. 2025.
\newblock Human Heterogeneity Invariant Stress Sensing.
\newblock \emph{Proceedings of the ACM on Interactive, Mobile, Wearable and Ubiquitous Technologies}, 9(3): 1--42.

\bibitem[{Xue and Marculescu(2023)}]{xue2023dynmm}
Xue, Z.; and Marculescu, R. 2023.
\newblock Dynamic multimodal fusion.
\newblock In \emph{Proceedings of the IEEE/CVF Conference on Computer Vision and Pattern Recognition (CVPR)}.

\bibitem[{Zhang et~al.(2018)}]{zhang2018mixup}
Zhang, H.; et~al. 2018.
\newblock mixup: Beyond empirical risk minimization.
\newblock In \emph{International Conference on Learning Representations (ICLR)}.

\bibitem[{Zhang et~al.(2023)}]{zhang2023qmf}
Zhang, Q.; et~al. 2023.
\newblock Provable dynamic fusion for low-quality multimodal data.
\newblock In \emph{International Conference on Machine Learning (ICML)}.

\bibitem[{Zhao et~al.(2023)}]{zhao2023affective}
Zhao, S.; et~al. 2023.
\newblock Affective and physiological responses to laboratory stress induction.
\newblock In \emph{Proceedings of the ACM on Interactive, Mobile, Wearable and Ubiquitous Technologies}, volume~7, 1--26.

\end{thebibliography}
}

\appendix
%
\renewcommand{\thesection}{\Alph{section}}
\renewcommand{\thetable}{\Alph{section}.\arabic{table}}
\renewcommand{\thefigure}{\Alph{section}.\arabic{figure}}
\setcounter{table}{0}
\setcounter{figure}{0}

\begin{center}
{\Large\bfseries Appendix}\\[2pt]
\small Supporting material for ``Representation Is Not Enough: Body-Localized Thermal Evidence for Contactless Stress and Craving Sensing in Opioid Use Disorder''
\end{center}

\def\pdflink#1#2{\leavevmode\pdfstartlink attr{/Border[0 0 0]}user{%
  /Subtype/Link/A<</Type/Action/S/URI/URI(#1)>>%
}#2\pdfendlink}

\noindent\textbf{Code and data availability.} Code and test data to reproduce the
results in this paper are available at
\pdflink{https://drive.google.com/drive/folders/1ZshKBEA5jcKXgfahKGvXK_sbPMdvWwXF?usp=sharing}{\url{https://drive.google.com/drive/folders/1ZshKBEA5jcKXgfahKGvXK_sbPMdvWwXF?usp=sharing}}.
The full dataset will be released upon acceptance of the paper.

\noindent\textbf{How to read this appendix.}
The section order follows the first use of each appendix reference in the
submitted paper: related work (A), the complete model and the scope of its
formal statements (B), data and labels (C), primary-corpus results (D), the
external StressNet study (E), craving (F), aggregator comparisons (G),
implementation (H), attribution (I), reproducibility (J), and governance (K).
Throughout, a ``measured
result'' means a configuration represented by a row in
a table.  Mechanistic explanations that have not been isolated by a matched
ablation are identified as hypotheses or design rationale.

\section{Extended Related Work}
\setcounter{table}{0}
This section expands Section~2 of the body. Table~\ref{tab:related_full}
compares the closest systems. We deliberately omit headline scores: the studies
use different stressors, targets, populations, splits, and label definitions, so
a score column would invite an invalid ranking.

\begin{table*}[t]
\centering\small
\caption{Representative work closest to thermal and weakly supervised stress
recognition. Scores are intentionally omitted (incompatible populations,
stressors, splits, and label definitions). Supports Section~2 of the body.}
\label{tab:related_full}
\begin{tabular}{@{}p{2.6cm}p{3.4cm}p{4.0cm}p{5.6cm}@{}}
\toprule
Study & Signal and spatial support & Modeling and supervision & Relevance to the present work \\
\midrule
\citet{shastri2012perinasal} & Perinasal thermal video & Morphological/wavelet perspiration extraction; lab and surgeon-expertise comparisons & Demonstrates a contactless sympathetic channel, but relies on a tracked, prespecified facial ROI and hand-crafted dynamics. \\
\citet{engert2014stress} & Six facial thermal imprints with conventional stress markers & Longitudinal statistics; multivariate phase classification & Shows regional sensitivity while documenting difficulty separating anticipation, stress, and recovery in a small healthy sample. \\
\citet{gioia2022contactless} & Multiple predefined facial ROIs ($\pm$ ECG/EDA/resp.) & Thermal statistics, recursive feature elimination, SVM & Supports thermal-only subject-independent recognition, but fixes ROI and feature vocabulary before learning. \\
\citet{kumar2021stressnet} & Facial thermal video with ECG/ICG reference & Heat-emission representation, spatiotemporal ISTI reconstruction & Learns video dynamics via a physiologically grounded cardiac intermediary; spatial support remains facial. \\
\citet{xiao2024reading} & Body thermal video plus EDA during training & Privileged-modality co-teaching, thermal-only deployment & Establishes whole-body thermal stress sensing, but uses a concurrent wearable teacher to shape thermal learning. \\
\citet{liu2024blush} & Facial thermal video and RGB rPPG & Feature-level fusion, conventional classifiers, SHAP & Complementary camera-derived physiology and explainability; fixed-feature facial multimodal design. \\
\citet{brugge2026taskaware} & RGB facial-video snippets & Task-conditioned temporal attention, top-/bottom-$k$ MIL & Direct precedent for weak task-level supervision and \emph{temporal} localization; models visible facial behaviour rather than regional body thermodynamics. \\
\midrule
\textbf{This work} & Whole-body thermal video, global token $+$ configurable region grid & Frozen multi-encoder features; dual-path temporal collapse; spatial set attention; gated MIL; branch-wise late fusion & Localizes evidence in \emph{space and time} under window labels; no contact-sensor teacher; cohort-resolved evaluation is part of the claim. \\
\bottomrule
\end{tabular}
\end{table*}

\paragraph{A.1 Literature outside computer science.}
The design rests on four non-CS literatures, summarized here because they
constrain what a thermal model may claim.
\emph{(a) Thermal psychophysiology.} Infrared thermography reflects interacting
changes in superficial perfusion and vasomotor tone, perspiration, respiration,
muscle activity, and metabolism, not a single stress channel
\citep{ioannou2014thermal,cardone2015thermal}. Contactless pulse and respiration
estimation \citep{garbey2007contact} shows thermal video
preserves fast periodic physiology alongside slow drift, which is why a
transient regional fluctuation may be physiologically real yet irrelevant to a
stress label.
\emph{(b) Regional non-universality.} Nose-tip temperature tracks arousal rather
than valence \citep{kosonogov2017arousal}; continuous nasal measurement shows
larger cooling under social than cognitive stress
\citep{theroude2026nose}; facial regions discriminate stress from cognitive load
\citep{gioia2021discriminating}; fingertip thermography has been explored for
stress--relaxation separation \citep{calderon2026peripheral}; and a systematic
review reports mixed directions of change across face and hand ROIs in children
\citep{nazzari2025review}. Thermal imaging has also been applied to cognitive
workload \citep{abdelrahman2017cognitive} and fear conditioning in PTSD
\citep{digiacinto2014fear}.
\emph{(c) Measurement validity.} \citet{stanic2025review} review 315
facial-thermography studies and find inconsistent reporting of camera
calibration, emissivity, geometry, acclimatization, ambient temperature,
airflow, occlusion, motion, and ROI definition; they propose explicit
acquisition and uncertainty guidelines. Our masking, normalization, whole-body
context, and person-disjoint evaluation are designed to \emph{reduce}, not
eliminate, these sources.
\emph{(d) Addiction medicine and health equity.} Stress is an established
relapse trigger under the brain-disease model of addiction
\citep{volkow2016neurobiology}; algorithmic bias in health systems is documented
at scale \citep{obermeyer2019bias,chen2021ethical,ghassemi2020review}.
Beyond the sources cited in the body, this literature draws on: OUD stigma and
its link to punitive policy \citep{cheetham2022stigma}; the
construct-validity critique of affect inference
\citep{mohammad2022ethics}; cognitive-biometric and
mental-privacy law \citep{ienca2024cognitive}; and machine learning in mental
health more broadly \citep{shatte2019digital}.

\section{Full Method Specification}
\setcounter{table}{0}
Supports Section~4 of the body. Notation follows the body. Numbers are provided with respect to Dataset I.

\paragraph{B.1 Window construction.}
For a window beginning at clock time $\tau_i$, the $T{=}25$ target times are
$q_{it}=\tau_i+(t-1)\tfrac{5}{25}$, $t=1,\dots,25$. For each target time the
builder selects the nearest recorded timestamp in the high-resolution reference
stream and matches the same physical frame in the other encoder streams. A
window is rejected if a match exceeds tolerance, repeats a frame, or leaves the
task interval. Timestamp alignment is a logical prerequisite for branch-wise
fusion: without it, the three branches would describe different instants.
The first and last target times are $4.8$ seconds apart; ``5-s window'' denotes
the builder's nominal interval, not five seconds between the first and last
sample.  This short span supports local latent changes, but it is not long
enough by itself to identify slow thermal equilibration or recovery dynamics.

\paragraph{B.2 Thermal rendering and its physical scope.}
For every region in every frame, valid pixels are min--max mapped independently
to $8$-bit,
$I_r(q)=255\,\frac{r(q)-r_{\min}}{r_{\max}-r_{\min}}$, replicated to three
channels, resized, and normalized; a zero image is used when no valid range
exists. This makes heterogeneous thermal matrices compatible with pretrained
image encoders and emphasizes within-region spatial structure.  It also removes
calibrated absolute temperature, a uniform warming or cooling of a region, and
calibrated comparisons between regions.  Accordingly, the learned trajectory
describes changes in an encoder's representation of normalized regional
appearance; it must not be interpreted as a temperature slope in degrees or as
direct evidence of vasoconstriction (body Section~\ref{sec:threats}(iii)). A
frozen encoder $\phi_e$ then produces
$F^{(e)}_i[t,n]=\phi_e(I_{i,t,n})\in\R^{768}$, cached once, retaining time and
region axes.

\paragraph{B.3 Projection.}
The frozen backbones do not share a coordinate system despite equal
dimensionality, so each is aligned before the shared operator:
$Z^{(e)}=\mathrm{Dropout}(\mathrm{LN}(F^{(e)}W_e+b_e))\in\R^{T\times N\times d}$.
Temporal collapse, spatial attention, and MIL pooling share parameters across
encoders; only projections and branch heads are encoder-specific.

\paragraph{B.4 Temporal descriptors.}
With $\tilde t$ the mean-centered frame index,
\begin{equation*}
\mu_n=\tfrac1T\textstyle\sum_t Z_{t,n},\quad
\sigma_n=\Big[\tfrac1T\sum_t (Z_{t,n}-\mu_n)^2\Big]^{1/2},
\end{equation*}
\begin{equation*}
s_n=\frac{\sum_t \tilde t\, Z_{t,n}}{\sum_t \tilde t^{\,2}},\quad
\delta_n=\tfrac{1}{T-1}\textstyle\sum_{t=1}^{T-1}\lvert Z_{t+1,n}-Z_{t,n}\rvert .
\end{equation*}
These represent latent level, dispersion, signed trend, and short-term
variation. They are parameter-free and computed before any learned mixing, so
the distinction between a stable, a directional, and an irregular trajectory is
available to the head rather than reconstructed by it.

\paragraph{B.5 Spatial contextualization and MIL pooling.}
For each branch, the temporal descriptor of every token is first contextualized
with the other spatial tokens by the shared spatial-attention operator.  A
gated MIL scorer then assigns normalized region weights
$a^{(e)}\in\Delta^{N-1}$ and pools the contextualized instances $h_n^{(e)}$ as
$b^{(e)}=\sum_n a_n^{(e)}h_n^{(e)}$.  Thus time is summarized within a region
before regions exchange context, and evidence is localized over regions only
after that exchange.  The global token is simply one member of this bag; it is
not privileged by a fixed pooling rule.

\paragraph{B.6 Fusion policy.}
$b^\star=\sum_e\pi_e(x)b^{(e)}$ with
$\pi(x)=\mathrm{softmax}_e(\mathrm{Gate}([b^{(e)};r^{(e)}]))$ and
$r^{(e)}=\mathrm{softplus}(q^{(e)})+\epsilon$. The gate is a per-window function
of the branch embeddings and their quality cues, so the view weights are
recomputed for every window. $r^{(e)}$ has no separate probabilistic calibration
objective and is interpreted only as a learned branch-quality feature the gate
may read---not a calibrated predictive variance and not a stand-alone confidence
score. All primary-corpus and craving results use this learned gate; a hard
uniform variant ($\pi_e\equiv1/E$) is available and used only where noted.

\paragraph{B.7 Full objective.}
For batch size $B$ and $E{=}3$ branches,
{\small
\begin{equation}
\mathcal{L}=\mathcal{L}_{\text{task}}+\lambda_{ds}\mathcal{L}_{ds}
+\lambda_{\text{cons}}\mathcal{L}_{\text{cons}}
+\lambda_{\text{ent}}(k)\mathcal{L}_{\text{ent}}
+\lambda_{\text{gate}}\mathcal{L}_{\text{gate}}
+\mathcal{L}_{\text{adv}},
\end{equation}
}
with $k$ the epoch and $\mathcal{L}_{\text{adv}}$ applied through gradient
reversal $R_{\lambda_{adv}(k)}(b^\star)$. Table~\ref{tab:losses} maps each term
to the failure mode it addresses and states where it attaches. Inactive terms
carry zero weight; gate regularization is irrelevant under hard uniform fusion,
and the subject adversary is disabled in the selected within-group
configuration.

\begin{table}[t]
\centering\small
\setlength{\tabcolsep}{3pt}
\caption{Loss terms as a map from failure mode to constraint. Supports body
Section~4, ``Training objective''. The paper does not factorially ablate these
terms; they are design rationale, not measured effects.}
\label{tab:losses}
\begin{tabular}{@{}llp{3.9cm}@{}}
\toprule
Term & Attaches to & Failure mode addressed \\
\midrule
$\mathcal{L}_{\text{task}}$ & fused logit (mixup of $b^\star$) & class imbalance; hard labels treated as certain \\
$\mathcal{L}_{ds}$ & each branch logit $\ell^{(e)}$ & fusion rescuing uninformative branches \\
$\mathcal{L}_{\text{cons}}$ & branch probabilities & view-specific noise masquerading as complementarity \\
$\mathcal{L}_{\text{ent}}$ & MIL weights $a_n$ & attention collapse to one cell / drift to uniform \\
$\mathcal{L}_{\text{gate}}$ & fusion policy $\pi$ & premature encoder monopoly \\
$\mathcal{L}_{\text{adv}}$ & $b^\star$ via GRL & identity shortcut surviving a person-disjoint split \\
\bottomrule
\end{tabular}
\end{table}

\emph{Task loss.} With $p_i=\sigma(\ell_i)$ and label smoothing
$\tilde y_i=(1-\varepsilon)y_i+\varepsilon/2$, the focal form is
$\mathcal{L}_{\text{task}}=\tfrac1B\sum_i \alpha_{t,i}(1-p_{t,i})^\gamma c_i$
with $p_{t,i}=\tilde y_ip_i+(1-\tilde y_i)(1-p_i)$ and $c_i$ the smoothed BCE
\citep{lin2017focal}. Balanced sampling already
changes how often each class is seen, so an additional positive-class weight
would double-correct the prior; focal modulation instead down-weights easy
examples. When focal loss is disabled the implementation uses smoothed BCE.
\emph{Mixup} is applied to the fused embedding, not to raw thermograms---mixing
thermograms would combine bodies, masks, and temperatures in physically
implausible ways \citep{zhang2018mixup}.
\emph{Deep supervision:} $\mathcal{L}_{ds}=\tfrac1E\sum_e
\mathcal{L}_{\text{task}}(\ell^{(e)},y)$, which makes equal
fusion weighting meaningful by ensuring each branch is independently predictive.
\emph{Consistency:} $\mathcal{L}_{\text{cons}}=\tfrac{1}{BE}\sum_{i,e}
(p^{(e)}_i-\bar p_i)^2$.
\emph{Entropy band:} $\mathcal{L}_{\text{ent}}=\tfrac{1}{BE}\sum_{i,e}
[H(a^{(e)}_i)-\rho\log N]^2$, targeting intermediate selectivity---a one-cell
solution is fragile to pose and mask error, while uniform attention recreates
global averaging. Its weight is ramped during warmup.
\emph{Gate KL:} $\mathcal{L}_{\text{gate}}=\tfrac1B\sum_i
\mathrm{KL}(\pi_i\|u)$ against uniform $u$.
\emph{Subject adversary:} $\mathcal{L}_{\text{adv}}=\tfrac1B\sum_i
\mathrm{CE}(\phi(R_{\lambda_{adv}(k)}(b^\star_i)),s_i)$ over training-participant
identity $s_i$ \citep{ganin2016dann}, ramped because removing identity before
the task is learned can also erase useful physiology.

\paragraph{B.8 Participant centering.}
For the proposed model, the denominator is a global, per-feature standard
deviation estimated from at most 256 randomly sampled \emph{training} windows
(sampling seed 0).  The numerator subtracts a participant-specific per-feature
mean estimated from at most 24 of that participant's available windows, sampled
without replacement and averaged over time.  Consequently, held-out
participants require an unlabeled calibration batch of up to 24 windows at
inference.  These are not necessarily the first 24 windows or a calm baseline;
the archived implementation can sample them across that participant's task
blocks. Across all 71 session records, this procedure sampled 397 non-stress
windows among 1,685 centering windows (23.56\%): 231/989 for Control and 166/696
for OUD. On the two held-out within-group cohorts specifically, the counts are
43/192 for Test-Control and 35/144 for Test-OUD. Because no labels are read,
this is transductive calibration rather than label leakage, but it is an
operational assumption and can encode a participant's task mixture. A streaming
or pre-task deployment must instead specify a causal calibration period and
evaluate it prospectively.

The published MIL heads in Appendix~G are trained and evaluated under the same
participant-centering transform as \method, so Tables~\ref{tab:mil} and~\ref{tab:abl_full}
vary the aggregator alone.

\paragraph{B.9 Formal statements and their scope.}
Supports Propositions~1--3 and Observation~4 of the body.

\emph{Proposition 1 (attention pooling is order-blind).} Let
$\alpha_{t,n}=\mathrm{softmax}_t\,g(Z_{t,n})$ where $g$ depends only on frame
content. Fix a permutation $\pi$ of $\{1,\dots,T\}$ and set
$Z'_{t,n}=Z_{\pi(t),n}$. Then $g(Z'_{t,n})=g(Z_{\pi(t),n})$ and
$\sum_s \exp g(Z'_{s,n})=\sum_s \exp g(Z_{s,n})$ because the sum is over all
frames. Hence $\alpha'_{t,n}=\alpha_{\pi(t),n}$ and
$\bar Z'_n=\sum_t \alpha_{\pi(t),n}Z_{\pi(t),n}=\sum_u \alpha_{u,n}Z_{u,n}
=\bar Z_n$. $\square$

\emph{Scope of Proposition 1.} This result applies to content-only attention
pooling with no positional code or order-sensitive preprocessing.  It is not a
claim about attention in general: positional embeddings, temporal convolutions,
or recurrent layers can encode order, and these alternatives were not tested in
the reported tables.  The four descriptors split accordingly. $\mu_n$ and $\sigma_n$ are
symmetric functions of the frame multiset and are therefore permutation-invariant,
adding no information the attention path lacks; they are retained because they
are cheap and stabilize the residual. $s_n$ is order-dependent and
\emph{antisymmetric} under time reversal ($\tilde t\mapsto-\tilde t$ gives
$s_n\mapsto -s_n$), and $\delta_n$ is order-dependent though reversal-invariant,
separating smooth drift from oscillation. The statistics path supplies these two
order-sensitive summaries to this particular content-only path.  Its empirical
increment has not been isolated by a removal ablation, and the short nominal
5-s span plus per-region/per-frame rendering in B.1--B.2 bounds any physiological
interpretation.

\emph{Proposition 2 (product-bag equivalence).} Fix a window $x$. With
$a^{(e)}\in\Delta^{N-1}$ and the learned gate $\pi(x)\in\Delta^{E-1}$, set
$c_{e,n}(x)=\pi_e(x)\,a^{(e)}_n$. Non-negativity is immediate, and
$\sum_{e,n}c_{e,n}(x)=\sum_e\pi_e(x)\sum_n a^{(e)}_n=\sum_e\pi_e(x)=1$, so
$c(x)$ is a distribution over the $EN$ product instances and
$b^\star=\sum_{e,n}c_{e,n}(x)\,h^{(e)}_n$ is an attention-MIL pooling of
$\mathcal{B}=\{(e,n)\}$. $\square$

\emph{Scope of Proposition 2.}
$c_{e,n}(x)=\pi_e(x)\,a^{(e)}_n$ is the chain rule $p(e,n)=p(e)p(n\mid e)$ and
hence parameterizes all of $\Delta^{EN-1}$ for each $x$; the factorization is not
a restriction on \emph{which} allocations are reachable. Its content is
procedural and \emph{window-adaptive}: $a^{(e)}$ is produced by the shared
scorer from view $e$ alone, and the gate $\pi(x)$ re-weights views per window, so
the within-branch distribution $a^{(e)}$ is view-local and view credit is
recomputed for every window rather than fixed.  The final product credit
$c_{e,n}$ is \emph{not} view-local, because $\pi_e(x)$ depends jointly on all
branch embeddings.  The proposition is an algebraic description of the model,
not evidence that the factorization is more expressive, causal, or more
accurate than an unconstrained joint scorer over the $EN$ instances. That
scorer is the matched ablation this comparison would need; we have not run it.

\emph{Proposition 3 (gauge reduction).} Let $P\in\mathcal{S}_d$ act by permuting
channels. LayerNorm computes its mean and variance as symmetric functions of the
channels, so $\mathrm{LN}_{P\gamma,P\beta}(Px)=P\,\mathrm{LN}_{\gamma,\beta}(x)$.
The gated scorer satisfies
$w^\top(\tanh(VP^{-1}Px)\odot\sigma(UP^{-1}Px))=w^\top(\tanh(Vx)\odot\sigma(Ux))$
once $U,V$ are post-multiplied by $P^{-1}$, so the pooling weights $a_n$ are
unchanged. Post-multiplying the branch's final projection by $P$ therefore sends
$h_n\mapsto Ph_n$ and $b=\sum_n a_nh_n\mapsto Pb$, while
$\mathrm{Head}\circ P^{-1}$ restores the logit exactly.

If the $\theta_e$ are untied, this construction may be applied independently in
each branch with distinct $P_e$: every $\ell^{(e)}$ is preserved, hence
$\mathcal{L}_{ds}$ and $\mathcal{L}_{\mathrm{cons}}$ are preserved, while
$b^\star\mapsto\sum_e\pi_e(x) P_eb^{(e)}$. Because the gate reads $b^{(e)}$, its
weights $\pi(x)$ also shift under the $P_e$, so the fused embedding is unanchored
in two ways at once; no single reparameterization of $\mathrm{FinalHead}$ can
restore it for general $P_e$. If $\theta_e\equiv\theta$,
a single parameter set is permuted, forcing $P_e\equiv P$; then
$b^\star\mapsto Pb^\star$ and $\mathrm{FinalHead}\circ P^{-1}$ leaves the fused
prediction invariant. The gauge group therefore collapses from
$\mathcal{S}_d^{\,E}$ to the diagonal $\mathcal{S}_d$. $\square$

\emph{Scope of Proposition 3.} Three qualifications are essential.  First, the
encoder-specific learned projections in B.3 already offer a mechanism for
aligning branch coordinates, even when downstream operators are untied.  Second,
the fused task loss does see
$b^\star$, so training an untied model would impose \emph{some} alignment through
that term alone; the claim is that nothing else in the objective constrains it and
that the loss surface carries an $E$-fold independent permutation symmetry, not
that untied fusion cannot be trained. Third, the argument establishes
identifiability of the fused representation, not superior accuracy; the empirical
comparison of fusion depths is Table~\ref{tab:abl_full}, but its ``separate MIL
+ logit attention'' comparator changes both parameter sharing and fusion depth.
It is therefore not the matched untied-operator test implied by the proposition.
A true test copies the complete operator once per branch while holding fusion,
normalization, losses, width, and tuning budget fixed; we have not run it. The
closest available comparator, which instead changes fusion depth as well as
parameter sharing, is discussed in Appendix~G.1.

\emph{Observation 4 (posterior fusion is a rank-$E$ bottleneck).} A posterior
rule is by definition a function $\psi(\ell^{(1)},\dots,\ell^{(E)})$ of the $E$
per-view logits. If $\ell^{(e)}(x)=\ell^{(e)}(x')$ for all $e$ then
$\psi$ returns the same value on $x$ and $x'$, irrespective of the allocations
$c(x),c(x')$ or the pooled evidence $b^{(e)}(x),b^{(e)}(x')$. Embedding fusion
distinguishes such a pair whenever $b^\star(x)\neq b^\star(x')$. The fixed
combination rules of \citet{kittler1998combining}, deep ensembles, and
quality-weighted late fusion are all of this form. $\square$

\emph{Parameter accounting.} Under sharing, $E$ branches cost
$|\theta|+E(768d+|{\rm Head}|)$ parameters rather than $E(|\theta|+768d+|{\rm
Head}|)$. With $d{=}96$ (Table~\ref{tab:hparams}) the projection $W_e$ dominates
the per-branch cost while $\theta$---the transformer layer, statistics MLP, and
gated scorer---is amortized across all three views, which is why the aggregator
remains small enough to train on 16{,}273 windows without overfitting within one
epoch (Appendix~H.3).

\section{Datasets and Splits}
\setcounter{table}{0}
Supports body Section~5.

\paragraph{C.1 Dataset I (our collection).} 16{,}273 windows, 71 sessions from
$67$ participants (Control $42$ / $11{,}410$ windows; OUD $29$ sessions from $25$
individuals / $4{,}863$). Overall $77.5\%$
stress. Cohort-wise: Control $8{,}754$ stress / $2{,}656$ non-stress ($76.7\%$);
OUD $3{,}851$ / $1{,}012$ ($79.2\%$). Blocks labelled \texttt{stress},
\texttt{stroop}, \texttt{song}, \texttt{arithmetic}, \texttt{bad} receive
$y{=}1$; \texttt{jelly} and \texttt{count} receive $y{=}0$. Labels therefore
encode the elicitation condition. Each window is a $25\times49\times768$ tensor
per encoder. This is a corpus we collected; the remainder of this subsection
gives the collection protocol and separates the full session protocol from the
seven blocks retained in the analytical corpus.

\subparagraph{Recruitment and consent.}
The collecting institution's IRB approved the study; all participants gave
informed consent. Control participants were recruited from the local community
and university population; OUD participants through clinical and community
partners. Participation was voluntary and OUD participants received
IRB-approved gift-card compensation for their time. The OUD cohort is \textbf{25 unique
individuals}; four participants were each recorded in two separate sessions,
approximately six months apart and under different treatment and medication conditions,
giving \textbf{29 OUD sessions}. Because splits, windows, and results are
defined at the session level, we report $29$ OUD sessions throughout and note the
$25$ individuals here. These recordings were collected across both medication
states---some before and some after the participant's daily medication for OUD
(MOUD). We include both so the corpus reflects naturally occurring variation in
medication timing; we do \emph{not} treat medication state as a studied variable. To keep evaluation strictly
person-disjoint, the four participants with two sessions are used only for training
and excluded from the held-out evaluation sets, so no individual appears in both
training and test. Demographics are summarized in
Table~\ref{tab:demographics}.

\begin{table*}[t]
\centering\small
\setlength{\tabcolsep}{5pt}
\begin{tabular}{@{}llll@{}}
\toprule
\textbf{Group} & \textbf{Age} & \textbf{Gender} & \textbf{Race/Ethnicity} \\
\midrule
Control ($n{=}42$) & $30.8\pm9.6$ & Female $70.0\%$ & White $60.0\%$, Asian $32.0\%$, \\
 & $[22\text{--}65]$ & Male $30.0\%$ & Black $6.0\%$, Other $2.0\%$ \\
\midrule
OUD ($n{=}25$) & $35.7\pm4.2$ & Female $75.0\%$, & White $58.3\%$, Black $29.2\%$, \\
 & $[29\text{--}47]$ & Male $20.8\%$, & Other $8.3\%$, \\
 & & Non-binary $4.2\%$ & Am.\ Indian/AK Native $4.2\%$ \\
\bottomrule
\end{tabular}
\caption{Participant demographics (mean\,$\pm$\,SD [range] or \%). OUD row is over
the $25$ unique individuals ($29$ sessions).}
\label{tab:demographics}
\end{table*}

\subparagraph{Apparatus and setting.}
Sessions ran in a private room---either a university research facility or a
hospital---with only the participant and a trained experimenter present, to
protect comfort and confidentiality while participants disclosed stress- and
craving-related experience. Participants stood $8$--$12$\,ft from a display
presenting the visual stimuli. A fixed long-wave infrared thermal camera recorded
the participant at $5$\,fps as $240\times320$ temperature matrices; this is the only sensor \method uses at inference. In this dataset, stress labels are assigned from task
condition and craving from post-block self-report. 

\subparagraph{Protocol rationale.}
The design follows SUD laboratory studies in which controlled stress exposure
elicits craving-related subjective and physiological responses
\citep{sinha2000psychological,fox2008enhanced,sinha2009modeling}. In-the-wild
craving data carry heavy label noise---episodes are brief, self-reports delayed or
missed, triggers uncontrolled---so we instead collect under stronger temporal
control. Because direct opioid-craving induction (e.g.\ enforced abstinence)
raises ethical and standardization concerns \citep{anderson2018ethical}, we use
validated stress-induction tasks as reproducible physiological probes. Crucially,
we do not treat stress as a proxy label for craving: stress is defined by task
structure, whereas craving is measured independently by self-report after each
block.

\begin{figure}[t]
\centering
\includegraphics[width=\columnwidth]{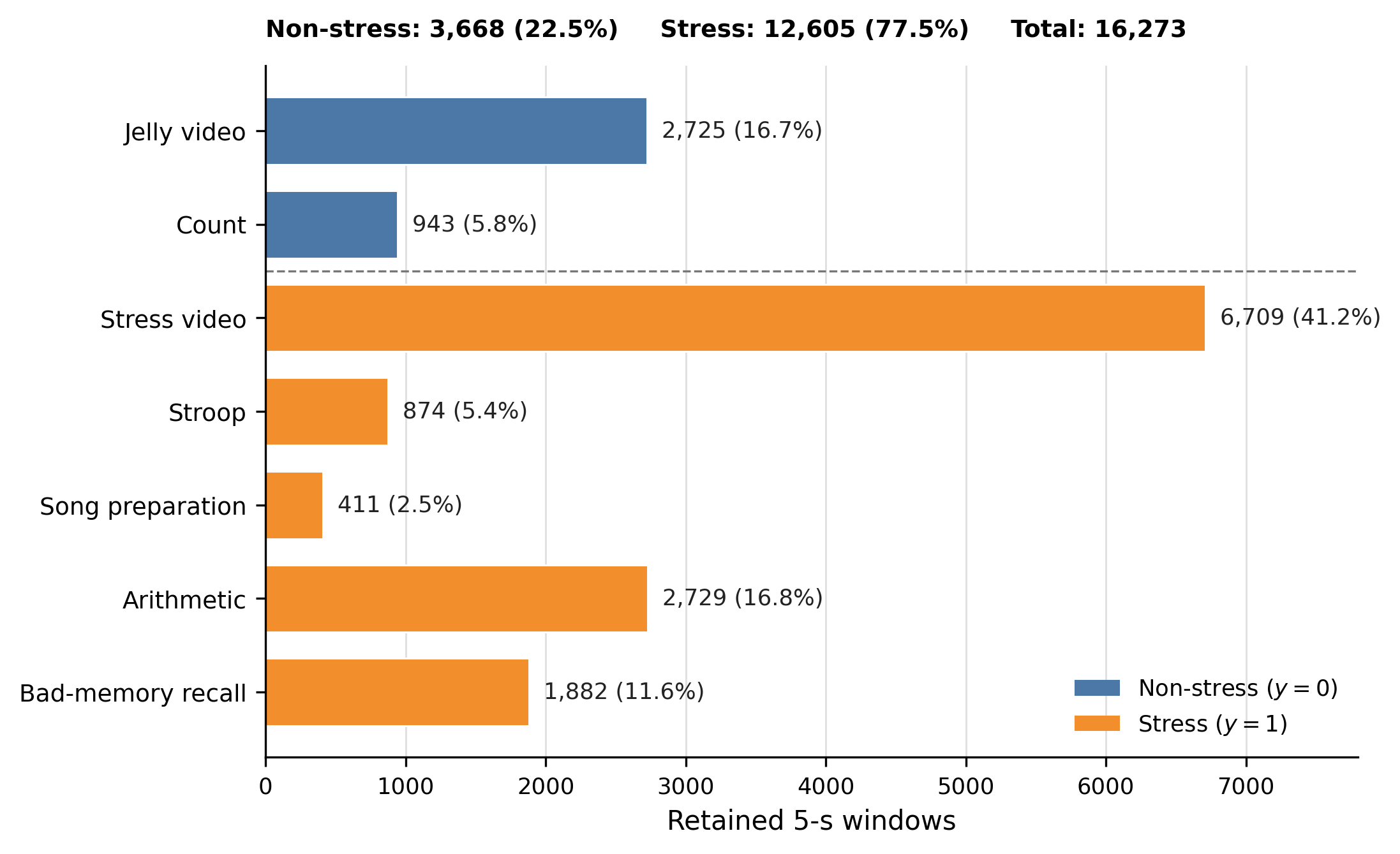}
\caption{\textbf{Distribution of the task windows used for stress/non-stress
prediction.} Exact counts and full-corpus percentages for the 16{,}273 retained
5-s windows in Dataset I. Jelly video and Count are assigned non-stress
($y{=}0$); Stress video, Stroop, Song preparation, Arithmetic, and Bad-memory
recall are assigned stress ($y{=}1$). The totals are 3{,}668 non-stress windows
($22.5\%$) and 12{,}605 stress windows ($77.5\%$). These are overlapping model
windows rather than independent trials, and task identity defines the
elicitation-condition label but is not supplied to the model as an input
feature.}
\label{fig:stress_task_dist}
\end{figure}

\subparagraph{Task sequence.}
Following laboratory affect/stress-elicitation protocols
\citep{xiao2024reading,zhao2023affective}, the source session interleaves calm
and stress-elicitation activities. The \emph{analytical stress corpus} uses
exactly seven task codes. The negative (non-stress/neutral/baseline) class contains \texttt{jelly} (a 3-min
jellyfish video intended to restore a calm state) and \texttt{count}
(self-paced counting from 0 to 59, approximately 1 min). The positive class
contains \texttt{stress} (passive stress-inducing video), \texttt{stroop}
(colour--word interference), \texttt{song} (the 30-s socially evaluative
sing-a-song preparation phase; singing itself is excluded),
\texttt{arithmetic} (approximately 3 min of serial subtraction), and
\texttt{bad} (approximately 2 min recalling a distressing personal memory).
Retained window counts are \texttt{jelly} 2{,}725, \texttt{count} 943, \texttt{stress} 6{,}709, \texttt{stroop} 874, \texttt{song} 411, \texttt{arithmetic} 2{,}729, and \texttt{bad} 1{,}882, summing to the reported 16{,}273 windows.
Figure~\ref{fig:stress_task_dist} visualizes this exact analytical-corpus
distribution and the task-to-class mapping used for model training and
evaluation.
The task-defined stress target is an elicitation-condition label, backed by literature, not a clinical diagnosis of stress \citep{golden1978stroop,kirschbaum1993trier}.

\subparagraph{Stress and craving annotation.}
Stress labels were defined by task structure (baseline vs.\ stress blocks),
leveraging established stress-induction protocols that elicit reliable and
reproducible physiological responses
\citep{ollander2016comparison,zhao2023affective}. Craving,
in contrast, was assessed independently through self-report because it is a
subjective motivational state that cannot be determined solely from task identity
\citep{maclean2019stress,sinha2000psychological,fox2008enhanced,sinha2009modeling}.
Although stress-inducing tasks may create conditions in which craving can arise,
craving is not assumed to occur during every stress block and may also occur
during non-stress periods, as shown in Appendix~C.2. Following each
task block, participants with OUD were asked to rate their perceived opioid
\emph{craving} as craving or not; control participants were not asked about opioid
craving.

Because craving can persist over short time intervals rather than occurring only
at a single instant
\citep{baillet2024craving,grunevski2024predictive}, each
post-task craving report was treated as reflecting the participant's craving state
over the corresponding task interval. This let us assign craving labels to the
physiological windows within that interval while keeping craving defined as a
participant-reported subjective state. For the binary craving target (RQ3),
non-craving and craving windows follow directly from these reports (label
distributions in Appendix~C.2); task identity is never used as the
craving label.

\paragraph{C.2 Dataset Characteristics.}\label{data-charact}
This section characterizes the distribution of physiological windows and craving
labels in the collected dataset. This analysis matters because craving is a
subjective state: even under an identical task protocol, sessions differ in
whether craving occurs, how often, and during which task blocks it is reported.

Table~\ref{tab:subject_craving_dist} reports the session-level breakdown. Window
counts are comparable across sessions (totals range $45$--$59$, median $50$), so
the heterogeneity is not an artifact of uneven data volume. Craving prevalence,
however, is sharply uneven: of the $29$ sessions, $10$ contain no craving windows
at all and $6$ are entirely craving, leaving only $13$ with both labels present.
In other words, $16$ of $29$ sessions ($55\%$) are single-class. Pooled across the
corpus the split looks near-balanced ($55.5\%$ craving over $4{,}863$ windows), but
that aggregate masks the fact that most individual sessions sit at one extreme or
the other. Because only $172$ independent participant--task labels underlie these
$4{,}863$ windows, the effective supervision is far sparser than the window count
suggests, and it is distributed very unevenly across people.

At task level, over the same seven retained task codes, craving rates vary
substantially by task:
approximately $0.40$ for Count, $0.41$ for Jelly, $0.53$ for Stress, $0.64$ for
Song, $0.64$ for Arithmetic, $0.66$ for Stroop, and $0.68$ for Bad-memory
recall. No task is purely one class for craving, so task identity is not the label.

Together these two views---single-class-dominated \emph{sessions} and
non-deterministic \emph{tasks}---show that craving labels are heterogeneous along
both axes. The practical consequences are the ones the body of the paper acts on:
supervision is sparse, and a subject-independent split is
essential (a random-window split would leak a session's single-class label into
its own test windows).

\begin{table}[t]
\centering
\small
\setlength{\tabcolsep}{6pt}
\caption{\textbf{Session-level distribution of craving labels across the $29$ OUD
sessions ($25$ individuals; 4 participants were recorded twice across different medication and treatment phase states, approximately six months apart).} Windows are
comparable in count across sessions, but craving prevalence is highly
heterogeneous---$16$ of $29$ sessions are single-class (no craving or all
craving), and only $13$ contain both labels. }
\label{tab:subject_craving_dist}
\begin{tabular}{lcc}
\toprule
Session profile & \# sessions & craving share \\
\midrule
No craving (all non-craving) & $10$ & $0\%$ \\
Mixed (both classes present) & $13$ & $1$--$99\%$ \\
All craving                  & $6$  & $100\%$ \\
\midrule
\textbf{Total}               & $\mathbf{29}$ & $\mathbf{55.5\%}$ \\
\bottomrule
\end{tabular}

\vspace{2pt}
{\small Corpus: $55.5\%$ craving, $172$ independent
task-block labels; $16/29$ sessions single-class. Per-session window totals range
$45$--$59$ (median $50$).}
\end{table}

\paragraph{C.3 Dataset II (StressNet).} $305$ trials from $34$ participants
under a cold-pressor protocol with session-level labels (session 2 = no stress,
session 3 = stress) \citep{kumar2021stressnet}. Video at $\approx7.5$\,fps after
source decimation; each example spans $100$ source frames ($13.3$\,s) with a
$50$-frame stride, retaining every fourth frame to yield $25$ sampled frames
without shortening the temporal span. $97$ regions (one global token plus a
$4\times24$ grid), so the same aggregation model operates on
$25\times97\times768$. Full construction yields $7{,}777$ windows; following the
cold-pressor timing protocol, all methods use only windows beginning after
$65$\,s, leaving $5{,}054$ ($2{,}566$ no-stress, $2{,}488$ stress). The fixed
subject-disjoint fold uses $26$/$5$/$3$ participants and $3{,}897$/$704$/$453$
windows; test participants are 108, 132, 161. This single fold was drawn at
random from the possible subject-disjoint partitions, not selected for
performance, and is then fixed across the proposed method, every baseline, and
seeds $\{42,100,2023\}$, so the within-fold comparison is controlled and
person-disjoint.

\paragraph{C.4 Dataset III (craving).} Reuses the $4{,}863$ OUD windows with the
per-block binary craving self-report replacing the stress target (craving vs.\
non-craving; see the annotation protocol above). This gives $2{,}698$ craving
($55.5\%$) and $2{,}165$ non-craving windows from $29$ OUD sessions ($25$
individuals).
All evaluations leveraged a fixed person-disjoint $21$/$4$/$4$ split, three seeds.
Craving uses a longer $20$\,s window than the stress task ($5$\,s), selected
empirically over shorter alternatives. A longer window helps because craving is a
slower, sustained motivational state rather than a discrete stimulus-locked
response: its thermal signature is a gradual autonomic drift that unfolds over
tens of seconds, so a short window captures only a fraction of the trajectory, and a longer one integrates more of the underlying dynamics, raising signal-to-noise. It also aligns the input extent with the supervision, since the craving label is assigned once per task block rather than at a single instant.

\section{Full Baseline Results (Primary Corpus)}
\setcounter{table}{0}
Table~\ref{tab:main} summarizes the best member of each baseline family (referenced
from body Section~\ref{sec:results}); the per-method breakdown follows.
AUROC is the controlled comparison; \method persists a validation-selected
threshold while most baselines retain $0.5$, so accuracy and F1 describe saved
operating points rather than a calibration comparison. ``Controlled'' here means
the same person-disjoint split and target, not identical preprocessing or tuning:
\method uses participant centering, the general baseline harness uses global
train-only standardization, and LoRA uses a separate raw-image path
(Appendix~B.8 and D.2). The AUROC gaps are therefore system-level comparisons.
Values are verbatim from the result CSVs.

\begin{table*}[t]
\centering
\small
\setlength{\tabcolsep}{4pt}
\caption{RQ1. Best AUROC per baseline family under the shared person-disjoint
splits ($3$ seeds); $k$ = methods in family. Underline marks the best of the 29
\emph{general} baselines per column; the published-MIL block below is analyzed
in Section~\ref{sec:arch} and is not counted among the 29. Full per-method tables
in Appendix~D. AUROC is the controlled comparison: \method uses a
validation-selected threshold while most baselines retain $0.5$, so accuracy and
F1 (Appendix~D) describe operating points rather than calibration.}
\label{tab:main}
\begin{tabular}{@{}lccc@{}}
\toprule
 & & \multicolumn{2}{c}{AUROC} \\
\cmidrule(l){3-4}
Baseline family (best member) & $k$ & Test-Both & Ctrl$\to$OUD \\
\midrule
Frozen linear probe (Concat-3)      & 4 & $0.675$ & $0.621$ \\
LoRA fine-tuning (ImageNet-ViT)     & 3 & $0.686$ & $0.608$ \\
Weight-space ens.\ (Snapshot)       & 6 & $0.720$ & $0.613$ \\
Learned fusion (TMC)                & 6 & \underline{$0.735$} & $0.657$ \\
Fixed comb.\ rules (Median/Min)     & 6 & $0.657$ & \underline{$0.662$} \\
Tabular ens.\ (XGBoost)             & 4 & $0.642$ & $0.582$ \\
\midrule
Published MIL, best (TransMIL/ABMIL)& 3 & $0.736$ & $0.679$ \\
\method, single encoder             & 1 & $0.747$ & $0.649$ \\
\midrule
\textbf{\method} (learned fusion)   & --& $\mathbf{0.938}\pmstd{.004}$ & $\mathbf{0.771}\pmstd{.013}$ \\
\bottomrule
\end{tabular}
\end{table*}

\begin{table*}[t]
\centering\small
\setlength{\tabcolsep}{5pt}
\caption{Full primary-corpus results: within-distribution Test-Both and
cross-cohort Control-to-OUD transfer, mean$\pm$std over $3$ seeds. The exact
positive-class accuracy floors are $0.788$ and $0.792$, respectively; $0.775$ is
the full-corpus rate, not either test-set floor. Underline: best baseline per
column; bold: best overall.}
\label{tab:full_main}
\begin{tabular}{@{}lcccccc@{}}
\toprule
& \multicolumn{3}{c}{Test-Both (within-distribution)} & \multicolumn{3}{c}{Control-to-OUD (cross-cohort)}\\
\cmidrule(lr){2-4}\cmidrule(l){5-7}
Method & Acc & F1 & AUROC & Acc & F1 & AUROC\\
\midrule
\multicolumn{7}{@{}l}{\emph{Frozen linear probe (logistic regression on global token)}}\\
AnyThermal    & 0.640 & 0.745 & 0.667\pmstd{.00} & 0.448 & 0.512 & 0.613\pmstd{.00}\\
ImageNet-ViT  & 0.612 & 0.713 & 0.644\pmstd{.00} & 0.518 & 0.608 & 0.615\pmstd{.00}\\
InfMAE        & 0.668 & 0.790 & 0.575\pmstd{.00} & 0.437 & 0.510 & 0.569\pmstd{.00}\\
Concat-3      & 0.638 & 0.741 & 0.675\pmstd{.00} & 0.440 & 0.496 & 0.621\pmstd{.00}\\
\multicolumn{7}{@{}l}{\emph{LoRA fine-tuning (adapted backbone + linear head)}}\\
AnyThermal    & 0.533 & 0.613 & 0.683\pmstd{.031} & 0.548 & 0.642 & 0.608\pmstd{.022}\\
ImageNet-ViT  & 0.408 & 0.417 & 0.686\pmstd{.014} & 0.686 & 0.784 & 0.590\pmstd{.039}\\
InfMAE        & 0.519 & 0.596 & 0.640\pmstd{.023} & 0.763 & 0.861 & 0.595\pmstd{.017}\\
\multicolumn{7}{@{}l}{\emph{Weight-space / randomization ensembles}}\\
Deep Ensemble \citep{lakshminarayanan2017deepensembles} & 0.691 & 0.781 & 0.716\pmstd{.025} & 0.462 & 0.536 & 0.618\pmstd{.005}\\
Snapshot Ensemble \citep{huang2017snapshot}             & 0.699 & 0.786 & 0.720\pmstd{.010} & 0.405 & 0.451 & 0.613\pmstd{.004}\\
Fast Geometric Ens.\ \citep{garipov2018fge}             & 0.683 & 0.774 & 0.711\pmstd{.002} & 0.398 & 0.440 & 0.613\pmstd{.007}\\
SWA-Gaussian \citep{maddox2019swag}                     & 0.694 & 0.785 & 0.711\pmstd{.003} & 0.425 & 0.487 & 0.595\pmstd{.015}\\
BatchEnsemble \citep{wen2020batchensemble}              & 0.670 & 0.770 & 0.686\pmstd{.031} & 0.463 & 0.533 & 0.626\pmstd{.005}\\
MIMMO \citep{ferianc2023mimmo}                          & 0.688 & 0.782 & 0.706\pmstd{.018} & 0.401 & 0.434 & 0.631\pmstd{.009}\\
\multicolumn{7}{@{}l}{\emph{Learned multimodal fusion}}\\
Mixture-of-Experts \citep{jacobs1991moe} & 0.563 & 0.661 & 0.641\pmstd{.009} & 0.541 & 0.636 & 0.580\pmstd{.023}\\
FuseMoE \citep{han2024fusemoe}           & 0.650 & 0.746 & 0.688\pmstd{.015} & 0.470 & 0.531 & 0.615\pmstd{.018}\\
DynMM \citep{xue2023dynmm}               & 0.655 & 0.752 & 0.688\pmstd{.012} & 0.424 & 0.470 & 0.614\pmstd{.005}\\
QMF \citep{zhang2023qmf}                 & 0.642 & 0.739 & 0.665\pmstd{.006} & 0.405 & 0.438 & 0.649\pmstd{.009}\\
MMTM \citep{joze2020mmtm}                & 0.679 & 0.770 & 0.706\pmstd{.025} & 0.451 & 0.527 & 0.594\pmstd{.008}\\
Trusted Multi-View \citep{han2021tmc}    & 0.710 & 0.802 & \underline{0.735}\pmstd{.022} & 0.610 & 0.708 & 0.657\pmstd{.019}\\
\multicolumn{7}{@{}l}{\emph{Fixed posterior combination rules \citep{kittler1998combining}}}\\
Sum rule      & 0.598 & 0.702 & 0.624\pmstd{.013} & 0.393 & 0.416 & 0.639\pmstd{.012}\\
Product rule  & 0.638 & 0.740 & 0.649\pmstd{.013} & 0.397 & 0.419 & 0.661\pmstd{.007}\\
Max rule      & 0.656 & 0.754 & 0.632\pmstd{.012} & 0.410 & 0.443 & 0.627\pmstd{.012}\\
Min rule      & 0.654 & 0.753 & 0.657\pmstd{.013} & 0.410 & 0.443 & 0.659\pmstd{.009}\\
Median rule   & 0.593 & 0.697 & 0.630\pmstd{.020} & 0.397 & 0.424 & \underline{0.662}\pmstd{.014}\\
Majority vote & 0.593 & 0.697 & 0.600\pmstd{.005} & 0.397 & 0.424 & 0.599\pmstd{.023}\\
\multicolumn{7}{@{}l}{\emph{Non-deep tabular ensembles (pooled frozen features)}}\\
Random Forest \citep{breiman2001randomforest} & 0.802 & 0.887 & 0.612\pmstd{.009} & 0.792 & 0.884 & 0.589\pmstd{.004}\\
ExtraTrees \citep{geurts2006extratrees}       & 0.793 & 0.882 & 0.621\pmstd{.005} & 0.790 & 0.883 & 0.611\pmstd{.004}\\
XGBoost \citep{chen2016xgboost}               & 0.665 & 0.775 & 0.642\pmstd{.010} & 0.716 & 0.823 & 0.582\pmstd{.006}\\
LightGBM \citep{ke2017lightgbm}               & 0.682 & 0.790 & 0.642\pmstd{.004} & 0.739 & 0.844 & 0.565\pmstd{.011}\\
\midrule
\textbf{\method (learned fusion)} & \textbf{0.884} & \textbf{0.926} & \textbf{0.938}\pmstd{.004} & \textbf{0.817} & \textbf{0.894} & \textbf{0.771}\pmstd{.013}\\
\bottomrule
\end{tabular}
\end{table*}

\paragraph{D.1 Reading Table~\ref{tab:full_main}.}
Three observations support claims made in body Section~6.
\emph{(a)} Random Forest and ExtraTrees post the highest baseline accuracy
($0.802$, $0.793$) with near-chance ranking ($0.612$, $0.621$), the clearest
demonstration that accuracy is prior-driven here.
\emph{(b)} Several methods invert across protocols: TMC is the best Test-Both
baseline but the median rule is the best Control-to-OUD baseline, so
within-distribution ranking does not predict cross-cohort ranking. Any
model-selection procedure that uses only within-distribution validation is
therefore selecting on the wrong criterion for the target population.
\emph{(c)} The LoRA rows show high Control-to-OUD accuracy ($0.686$--$0.763$)
with low AUROC ($0.590$--$0.608$), i.e.\ collapse toward the majority class.
The corresponding positive-class floors are Test-Control $0.792$, Test-OUD
$0.778$, Test-Both $0.788$, and Control-to-OUD $0.792$. Thus \method's Test-OUD
accuracy of $0.784$ is $0.006$ above its own split's majority predictor, not
below a $0.792$ floor; its AUROC $0.810$ is the more informative ranking metric.

The $0.203$ Test-Both AUROC margin over the best general baseline is arithmetically
$0.938-0.735$. Table~\ref{tab:abl_full} narrows the architectural comparison.

\paragraph{D.2 LoRA harness disclosure.}
LoRA baselines use rank-$8$, $\alpha{=}16$, dropout-$0.05$ adapters (PEFT)
injected into the query/value projections for AnyThermal (DINOv2 naming) and
the fused QKV projection for ImageNet-ViT and InfMAE (timm naming), with a
trainable $\mathrm{Linear}(19{,}200,2)$ head over flattened $25\times768$
features. Trainable adapter parameters are $\approx295$k (AnyThermal, 48
tensors), $\approx295$k (ImageNet-ViT, 24 tensors), and $\approx270$k (InfMAE,
22 tensors) against frozen backbones of $86.6$M/$85.8$M/$96.7$M, under $0.4\%$
of each. Training uses AdamW (lr $10^{-3}$, weight decay $10^{-4}$),
class-weighted cross-entropy, bf16 mixed precision
, and early stopping on validation balanced
accuracy (patience 30, up to 100 epochs), over seeds
$\{1234,1235,1236\}$---a different convention from the $\{42,100,2023\}$ used
elsewhere. Split files are identical, so results remain person-disjoint and
comparable, but ``identical harness'' does not literally apply to this family.
\emph{\method uses frozen features and trains no LoRA adapters.}

\begin{table*}[t]
\centering\small
\setlength{\tabcolsep}{5pt}
\caption{Within-group per-cohort breakdown: train on both cohorts, evaluate
separately on held-out Test-Control ($8$ subjects, $2{,}306$ windows) and
Test-OUD ($6$ subjects, $1{,}000$ windows). Same baselines, seeds, and
conventions as Table~\ref{tab:full_main}. These are the two protocols that make
the decomposition in body Table~2 computable.}
\label{tab:full_cohort}
\begin{tabular}{@{}lcccccc@{}}
\toprule
& \multicolumn{3}{c}{Test-Control} & \multicolumn{3}{c}{Test-OUD}\\
\cmidrule(lr){2-4}\cmidrule(l){5-7}
Method & Acc & F1 & AUROC & Acc & F1 & AUROC\\
\midrule
\multicolumn{7}{@{}l}{\emph{Frozen linear probe}}\\
AnyThermal   & 0.637 & 0.747 & 0.649\pmstd{.00} & 0.649 & 0.740 & 0.719\pmstd{.00}\\
ImageNet-ViT & 0.592 & 0.683 & 0.705\pmstd{.00} & 0.657 & 0.772 & 0.500\pmstd{.00}\\
InfMAE       & 0.663 & 0.782 & 0.589\pmstd{.00} & 0.681 & 0.806 & 0.523\pmstd{.00}\\
Concat-3     & 0.640 & 0.749 & 0.661\pmstd{.00} & 0.632 & 0.720 & 0.728\pmstd{.00}\\
\multicolumn{7}{@{}l}{\emph{LoRA fine-tuning}}\\
AnyThermal   & 0.523 & 0.594 & 0.726\pmstd{.054} & 0.555 & 0.644 & 0.588\pmstd{.042}\\
ImageNet-ViT & 0.457 & 0.485 & \underline{0.787}\pmstd{.018} & 0.298 & 0.230 & 0.440\pmstd{.021}\\
InfMAE       & 0.494 & 0.562 & 0.681\pmstd{.029} & 0.577 & 0.646 & 0.578\pmstd{.009}\\
\multicolumn{7}{@{}l}{\emph{Weight-space / randomization ensembles}}\\
Deep Ensemble     & 0.683 & 0.768 & 0.735\pmstd{.011} & 0.707 & 0.807 & 0.684\pmstd{.063}\\
Snapshot Ensemble & 0.692 & 0.777 & 0.709\pmstd{.008} & 0.718 & 0.804 & \underline{0.770}\pmstd{.006}\\
Fast Geometric Ens.& 0.687 & 0.777 & 0.707\pmstd{.006} & 0.674 & 0.765 & 0.723\pmstd{.008}\\
SWA-Gaussian      & 0.706 & 0.794 & 0.718\pmstd{.019} & 0.666 & 0.762 & 0.694\pmstd{.057}\\
BatchEnsemble     & 0.658 & 0.749 & 0.712\pmstd{.022} & 0.696 & 0.811 & 0.617\pmstd{.029}\\
MIMMO             & 0.678 & 0.765 & 0.724\pmstd{.011} & 0.711 & 0.812 & 0.681\pmstd{.051}\\
\multicolumn{7}{@{}l}{\emph{Learned multimodal fusion}}\\
Mixture-of-Experts & 0.518 & 0.597 & 0.650\pmstd{.021} & 0.668 & 0.776 & 0.651\pmstd{.025}\\
FuseMoE            & 0.637 & 0.730 & 0.701\pmstd{.030} & 0.680 & 0.778 & 0.678\pmstd{.044}\\
DynMM              & 0.648 & 0.746 & 0.688\pmstd{.008} & 0.673 & 0.765 & 0.700\pmstd{.027}\\
QMF                & 0.632 & 0.726 & 0.686\pmstd{.010} & 0.665 & 0.765 & 0.631\pmstd{.035}\\
MMTM               & 0.652 & 0.737 & 0.723\pmstd{.013} & 0.741 & 0.831 & 0.702\pmstd{.043}\\
Trusted Multi-View & 0.700 & 0.788 & 0.775\pmstd{.010} & 0.734 & 0.832 & 0.678\pmstd{.037}\\
\multicolumn{7}{@{}l}{\emph{Fixed posterior combination rules}}\\
Sum rule      & 0.563 & 0.666 & 0.616\pmstd{.018} & 0.680 & 0.776 & 0.665\pmstd{.010}\\
Product rule  & 0.613 & 0.717 & 0.644\pmstd{.022} & 0.696 & 0.789 & 0.671\pmstd{.008}\\
Max rule      & 0.644 & 0.742 & 0.627\pmstd{.015} & 0.683 & 0.779 & 0.669\pmstd{.030}\\
Min rule      & 0.642 & 0.741 & 0.675\pmstd{.024} & 0.681 & 0.778 & 0.646\pmstd{.022}\\
Median rule   & 0.562 & 0.664 & 0.632\pmstd{.027} & 0.665 & 0.764 & 0.655\pmstd{.031}\\
Majority vote & 0.562 & 0.664 & 0.588\pmstd{.011} & 0.665 & 0.764 & 0.642\pmstd{.013}\\
\multicolumn{7}{@{}l}{\emph{Non-deep tabular ensembles}}\\
Random Forest & 0.812 & 0.892 & 0.658\pmstd{.012} & 0.778 & 0.875 & 0.638\pmstd{.004}\\
ExtraTrees    & 0.799 & 0.885 & 0.668\pmstd{.008} & 0.778 & 0.875 & 0.625\pmstd{.007}\\
XGBoost       & 0.613 & 0.717 & 0.668\pmstd{.011} & 0.786 & 0.879 & 0.653\pmstd{.019}\\
LightGBM      & 0.638 & 0.744 & 0.675\pmstd{.008} & 0.782 & 0.877 & 0.653\pmstd{.008}\\
\midrule
\textbf{\method (learned fusion)} & \textbf{0.927} & \textbf{0.953} & \textbf{0.979}\pmstd{.002} & \textbf{0.784} & \textbf{0.862} & \textbf{0.810}\pmstd{.002}\\
\bottomrule
\end{tabular}
\end{table*}

\begin{table*}[t]
\centering
\small
\setlength{\tabcolsep}{4.5pt}
\caption{\method resolved by cohort and training composition. The matched
decomposition (Sec.~\ref{sec:gap}) evaluates the Control-only model on the same
six Test-OUD participants; the significance test comparing $\Delta_{\text{repr}}$
against $\Delta_{\text{heter}}$ favors heterogeneity in $61.45\%$ of paired
seed-$43$ participant resamples but is not significant ($p=0.386$; difference
$95\%$ CI $[-0.076,0.200]$).}
\label{tab:gap}
\begin{tabular}{@{}lccc@{}}
\toprule
Protocol & OUD in train? & AUROC & Acc. \\
\midrule
Test-Control      & yes & $0.979$ & $0.927$ \\
Test-Both         & yes & $0.938$ & $0.884$ \\
Test-OUD ($6$)    & yes & $0.810$ & $0.784$ \\
Control$\to$OUD ($29$) & \textbf{no}  & $0.771$ & $0.817$ \\
Control-only @ the same $6$ & \textbf{no} & $0.652$ & $0.785$ \\
\midrule
$\Delta_{\text{repr}}=0.810-{@6}$ & & $0.158$ & \\
$\Delta_{\text{heter}}=0.979-0.810$ & & $0.169$ & \\
\bottomrule
\end{tabular}
\end{table*}

\paragraph{D.3 The cohort gap recurs across model families.}
Table~\ref{tab:full_cohort} shows that the Control$\to$OUD asymmetry is not
observed only for \method.
Eighteen of the twenty-nine baselines also score lower on Test-OUD than Test-Control
in AUROC, and the ImageNet-ViT linear probe collapses to exactly $0.500$ on
Test-OUD while reaching $0.705$ on Test-Control. This repeated pattern is
consistent with cohort shift, but it does not identify its cause. Clinical
heterogeneity, sample size, site, ambient conditions, protocol execution, label
noise, and representation shortfall can all contribute.

The symbol $\Delta_{\text{heter}}=0.979-0.810$ in Table~\ref{tab:gap} is therefore
best read as an \emph{observed residual cohort gap}, not a pure estimate of
within-OUD physiological heterogeneity. Likewise,
$\Delta_{\text{repr}}=0.810-0.652$ is a matched-set performance increment from
including OUD data during training, not a representation-only causal effect.
The arithmetic decomposition is descriptive and the reported paired bootstrap
does not separate these mechanisms.

\paragraph{D.4 Uncertainty in the decomposition.}
The reported bootstrap uses 2,000 participant-level resamples at seed 43, the
$61.45\%$ ordering frequency, $p=0.386$, and the displayed difference interval.
With only six Test-OUD participants, the sampling distribution is necessarily
coarse. 

\begin{table}[t]
\centering\small
\caption{\method across the four person-disjoint protocols (mean$\pm$std, $3$
seeds). Expanded form of body Table~2.}
\label{tab:ours_full}
\setlength{\tabcolsep}{4pt}
\begin{tabular}{@{}lccc@{}}
\toprule
Setting & Accuracy & AUROC & Positive-class F1\\
\midrule
Test-Both       & 0.884\pmstd{0.008} & 0.938\pmstd{0.004} & 0.926\pmstd{0.006}\\
Test-Control    & 0.927\pmstd{0.008} & 0.979\pmstd{0.002} & 0.953\pmstd{0.006}\\
Test-OUD        & 0.784\pmstd{0.010} & 0.810\pmstd{0.002} & 0.862\pmstd{0.008}\\
Control-to-OUD  & 0.817\pmstd{0.007} & 0.771\pmstd{0.013} & 0.894\pmstd{0.005}\\
\bottomrule
\end{tabular}
\end{table}

\paragraph{D.5 StressNet on the primary corpus: a diagnostic comparison.}
\label{sec:stressnet_ours_cohort}
The comparison above uses \method's own cohort split throughout. As a further
check, we verified against an established foundation-model approach for
thermal-video stress recognition, StressNet \citep{kumar2021stressnet}, applied
directly to \emph{our own} cohort rather than to its native cold-pressor
dataset (the reverse direction, \method on the StressNet dataset, is
Appendix~E). StressNet's original supervision reconstructs an ECG-derived
heat-emission/ISTI signal; our corpus provides no ECG channel, so we did not
attempt to reproduce that pathway. Instead we trained and evaluated StressNet
directly against the same task-defined ground-truth labels used throughout this
work (Appendix~C.1), with $100$-frame windows and a person-disjoint random
train/validation/test split. Tables~\ref{tab:stressnet_within} and
\ref{tab:stressnet_controloud} report the result beside \method, whose numbers
are those already reported in Tables~\ref{tab:full_cohort}, \ref{tab:full_main},
\ref{tab:gap}, and \ref{tab:ours_full}. Pooled over all windows in this
configuration, StressNet reaches accuracy $0.595\pmstd{0.062}$, balanced
accuracy $0.593\pmstd{0.067}$, AUROC $0.506\pmstd{0.182}$, and F1
$0.440\pmstd{0.311}$ (mean$\pm$std, seeds $42/100/2023$)---near chance on the
controlled AUROC metric and with substantially higher variance than \method,
consistent with the person-disjoint, task-labelled setting being genuinely hard
for a method built around a cardiac intermediary it cannot access here. This is
a diagnostic transfer, not a faithful reproduction of StressNet's original
ECG-supervised objective and not a matched head-to-head architectural ablation:
window duration, representation, normalization, and tuning differ. It should not
be used to claim that the original StressNet method is inferior on its intended
problem. Domain-specific thermal baselines beyond this adaptation remain useful
future comparisons.

\begin{table*}[t]
\centering\small
\setlength{\tabcolsep}{4pt}
\caption{Within-group protocols: \method (already reported in
Tables~\ref{tab:full_cohort}/\ref{tab:full_main}/\ref{tab:ours_full}) vs.\
StressNet \citep{kumar2021stressnet} retrained and evaluated on our own cohort
under task-defined labels, $100$-frame windows, and a person-disjoint random
split (mean$\pm$std, seeds $42/100/2023$).}
\label{tab:stressnet_within}
\begin{tabular}{@{}lcc ccc ccc@{}}
\toprule
& & & \multicolumn{3}{c}{\method} & \multicolumn{3}{c}{StressNet (ours-relabelled)}\\
\cmidrule(lr){4-6}\cmidrule(l){7-9}
Protocol & $n$ subj. & $n$ win. & Acc & F1 & AUROC & Acc & F1 & AUROC \\
\midrule
Test-Control & $8$ & $2{,}306$ & $0.927$ & $0.953$ & $0.979$ & $0.582\pmstd{.265}$ & $0.578\pmstd{.409}$ & $0.484\pmstd{.035}$\\
Test-OUD     & $6$ & $1{,}000$ & $0.784$ & $0.862$ & $0.810$ & $0.573\pmstd{.249}$ & $0.567\pmstd{.401}$ & $0.551\pmstd{.024}$\\
Test-Both    & $14$ & $3{,}306$ & $0.884$ & $0.926$ & $0.938$ & $0.580\pmstd{.260}$ & $0.575\pmstd{.406}$ & $0.512\pmstd{.020}$\\
\bottomrule
\end{tabular}
\end{table*}

\begin{table*}[t]
\centering\small
\setlength{\tabcolsep}{4pt}
\caption{Cross-cohort protocol: \method (already reported in
Tables~\ref{tab:gap}/\ref{tab:ours_full}) vs.\ StressNet trained on Control and
evaluated on OUD, same relabeling and split convention as
Table~\ref{tab:stressnet_within}.}
\label{tab:stressnet_controloud}
\begin{tabular}{@{}lcc ccc ccc@{}}
\toprule
& & & \multicolumn{3}{c}{\method} & \multicolumn{3}{c}{StressNet (ours-relabelled)}\\
\cmidrule(lr){4-6}\cmidrule(l){7-9}
Protocol & $n$ subj. & $n$ win. & Acc & F1 & AUROC & Acc & F1 & AUROC \\
\midrule
Control$\to$OUD & $29$ & $4{,}863$ & $0.817$ & $0.894$ & $0.771$ & $0.466\pmstd{.243}$ & $0.459\pmstd{.362}$ & $0.446\pmstd{.034}$\\
\bottomrule
\end{tabular}
\end{table*}

\section{StressNet (RQ2) Full Results}
\setcounter{table}{0}
Supports body Section~6.4. Reminder of scope: one randomly selected, fixed
subject-disjoint fold with $26$/$5$/$3$ participants and three test participants
(Appendix~C.3), evaluated over three training seeds
($\{42,100,2023\}$); within-fold controlled, not dataset-wide. The body reports
\method's mean$\pm$std AUROC on this fold, $0.962\pm0.017$, as the honest
external estimate. Boldface in Tables~\ref{tab:sn_main}--\ref{tab:sn_mil}
therefore identifies the maximum on this selected fold only, not a
dataset-level state-of-the-art result.

\begin{table*}[t]
\centering\small
\setlength{\tabcolsep}{5pt}
\caption{StressNet baseline and ensemble comparison on one selected, fixed
subject-disjoint fold (mean$\pm$population std over seeds $\{42,100,2023\}$).
MIL methods are in Table~\ref{tab:sn_mil}. Underline: best non-MIL baseline;
bold: best overall.}
\label{tab:sn_main}
\begin{tabular}{@{}lcccc@{}}
\toprule
Method & Accuracy & Balanced accuracy & F1 & AUROC\\
\midrule
\multicolumn{5}{@{}l}{\emph{Frozen linear probes}}\\
AnyThermal   & 0.459\pmstd{.000} & 0.452\pmstd{.000} & 0.152\pmstd{.000} & 0.597\pmstd{.000}\\
ImageNet-ViT & 0.676\pmstd{.000} & 0.670\pmstd{.000} & 0.545\pmstd{.000} & 0.863\pmstd{.000}\\
InfMAE       & 0.737\pmstd{.000} & 0.733\pmstd{.000} & 0.649\pmstd{.000} & 0.902\pmstd{.000}\\
Concat-3     & 0.477\pmstd{.000} & 0.469\pmstd{.000} & 0.144\pmstd{.000} & 0.601\pmstd{.000}\\
\multicolumn{5}{@{}l}{\emph{Non-deep tabular ensembles}}\\
Random Forest & 0.521\pmstd{.008} & 0.512\pmstd{.008} & 0.117\pmstd{.020} & 0.816\pmstd{.016}\\
ExtraTrees    & 0.522\pmstd{.012} & 0.514\pmstd{.012} & 0.123\pmstd{.038} & 0.783\pmstd{.008}\\
XGBoost       & 0.560\pmstd{.007} & 0.552\pmstd{.008} & 0.225\pmstd{.025} & 0.868\pmstd{.018}\\
LightGBM      & 0.548\pmstd{.007} & 0.540\pmstd{.007} & 0.204\pmstd{.021} & 0.847\pmstd{.022}\\
\multicolumn{5}{@{}l}{\emph{Weight-space / randomization ensembles}}\\
Deep Ensemble      & 0.687\pmstd{.023} & 0.681\pmstd{.024} & 0.558\pmstd{.050} & 0.920\pmstd{.004}\\
Snapshot Ensemble  & 0.745\pmstd{.071} & 0.741\pmstd{.072} & 0.664\pmstd{.111} & 0.880\pmstd{.035}\\
Fast Geometric Ens.& 0.640\pmstd{.084} & 0.634\pmstd{.086} & 0.419\pmstd{.195} & 0.914\pmstd{.015}\\
SWA-Gaussian       & 0.667\pmstd{.050} & 0.662\pmstd{.051} & 0.511\pmstd{.112} & 0.874\pmstd{.029}\\
BatchEnsemble      & 0.745\pmstd{.016} & 0.741\pmstd{.016} & 0.668\pmstd{.035} & \underline{0.924}\pmstd{.036}\\
MIMMO              & 0.687\pmstd{.130} & 0.682\pmstd{.132} & 0.564\pmstd{.220} & 0.862\pmstd{.092}\\
\multicolumn{5}{@{}l}{\emph{Naive early/late fusion}}\\
Early concatenation & 0.666\pmstd{.074} & 0.660\pmstd{.075} & 0.482\pmstd{.176} & 0.830\pmstd{.088}\\
Late mean           & 0.751\pmstd{.038} & 0.747\pmstd{.039} & 0.662\pmstd{.067} & 0.900\pmstd{.076}\\
Late max            & 0.620\pmstd{.087} & 0.613\pmstd{.088} & 0.351\pmstd{.243} & 0.825\pmstd{.016}\\
Late vote           & \underline{0.776}\pmstd{.098} & \underline{0.772}\pmstd{.100} & \underline{0.686}\pmstd{.186} & 0.807\pmstd{.097}\\
\multicolumn{5}{@{}l}{\emph{Learned multi-encoder fusion}}\\
Mixture-of-Experts & 0.617\pmstd{.014} & 0.614\pmstd{.013} & 0.548\pmstd{.037} & 0.679\pmstd{.084}\\
FuseMoE            & 0.702\pmstd{.085} & 0.697\pmstd{.086} & 0.558\pmstd{.190} & 0.908\pmstd{.048}\\
DynMM              & 0.608\pmstd{.094} & 0.603\pmstd{.095} & 0.468\pmstd{.157} & 0.744\pmstd{.119}\\
QMF                & 0.695\pmstd{.074} & 0.690\pmstd{.075} & 0.557\pmstd{.147} & 0.912\pmstd{.034}\\
MMTM               & 0.644\pmstd{.039} & 0.638\pmstd{.040} & 0.465\pmstd{.095} & 0.879\pmstd{.011}\\
Trusted Multi-View & 0.676\pmstd{.069} & 0.670\pmstd{.071} & 0.491\pmstd{.170} & 0.843\pmstd{.020}\\
\multicolumn{5}{@{}l}{\emph{Fixed posterior combination rules}}\\
Sum      & 0.674\pmstd{.021} & 0.668\pmstd{.022} & 0.516\pmstd{.052} & 0.846\pmstd{.058}\\
Product  & 0.676\pmstd{.028} & 0.669\pmstd{.029} & 0.518\pmstd{.066} & 0.879\pmstd{.046}\\
Max      & 0.685\pmstd{.046} & 0.679\pmstd{.047} & 0.542\pmstd{.096} & 0.821\pmstd{.070}\\
Min      & 0.685\pmstd{.046} & 0.679\pmstd{.047} & 0.542\pmstd{.096} & 0.895\pmstd{.037}\\
Median   & 0.678\pmstd{.017} & 0.672\pmstd{.018} & 0.527\pmstd{.041} & 0.886\pmstd{.027}\\
Majority vote & 0.678\pmstd{.017} & 0.672\pmstd{.018} & 0.527\pmstd{.041} & 0.769\pmstd{.065}\\
\midrule
\textbf{\method} & \textbf{0.823}\pmstd{.042} & \textbf{0.820}\pmstd{.044} & \textbf{0.782}\pmstd{.070} & \textbf{0.962}\pmstd{.017}\\
\bottomrule
\end{tabular}
\end{table*}

\begin{table*}[t]
\centering\small
\setlength{\tabcolsep}{5pt}
\caption{StressNet MIL comparison on the same selected fixed fold
(mean$\pm$population std, three seeds). Concat-3 = early concatenation of the
three frozen encoders; ``single encoder'' = the proposed branch without ensemble
fusion. Underline: best baseline; bold: best overall.}
\label{tab:sn_mil}
\begin{tabular}{@{}llcccc@{}}
\toprule
Aggregator & Encoder/input & Accuracy & Balanced acc. & F1 & AUROC\\
\midrule
\multirow{5}{*}{ABMIL \citep{ilse2018abmil}}
 & AnyThermal (gated)   & 0.585\pmstd{.022} & 0.577\pmstd{.022} & 0.307\pmstd{.058} & 0.720\pmstd{.044}\\
 & ImageNet-ViT (gated) & \underline{0.780}\pmstd{.084} & \underline{0.776}\pmstd{.086} & \underline{0.702}\pmstd{.135} & \underline{0.949}\pmstd{.025}\\
 & InfMAE (gated)       & 0.715\pmstd{.061} & 0.711\pmstd{.063} & 0.603\pmstd{.147} & 0.891\pmstd{.066}\\
 & Concat-3 (gated)     & 0.666\pmstd{.074} & 0.660\pmstd{.075} & 0.482\pmstd{.176} & 0.830\pmstd{.088}\\
 & Concat-3 (plain)     & 0.659\pmstd{.030} & 0.655\pmstd{.029} & 0.580\pmstd{.010} & 0.779\pmstd{.077}\\
\multirow{4}{*}{TransMIL \citep{shao2021transmil}}
 & AnyThermal   & 0.589\pmstd{.043} & 0.582\pmstd{.044} & 0.334\pmstd{.172} & 0.775\pmstd{.080}\\
 & ImageNet-ViT & 0.734\pmstd{.042} & 0.729\pmstd{.043} & 0.628\pmstd{.084} & 0.884\pmstd{.067}\\
 & InfMAE       & 0.588\pmstd{.051} & 0.586\pmstd{.050} & 0.514\pmstd{.129} & 0.657\pmstd{.076}\\
 & Concat-3     & 0.717\pmstd{.044} & 0.711\pmstd{.045} & 0.588\pmstd{.085} & 0.885\pmstd{.063}\\
\multirow{4}{*}{DSMIL \citep{li2021dsmil}}
 & AnyThermal   & 0.557\pmstd{.034} & 0.550\pmstd{.035} & 0.309\pmstd{.056} & 0.659\pmstd{.101}\\
 & ImageNet-ViT & 0.690\pmstd{.046} & 0.685\pmstd{.047} & 0.576\pmstd{.085} & 0.847\pmstd{.042}\\
 & InfMAE       & 0.731\pmstd{.047} & 0.728\pmstd{.049} & 0.657\pmstd{.094} & 0.839\pmstd{.086}\\
 & Concat-3     & 0.660\pmstd{.034} & 0.655\pmstd{.034} & 0.529\pmstd{.033} & 0.851\pmstd{.095}\\
\multirow{3}{*}{Proposed MIL (single enc.)}
 & AnyThermal   & 0.666\pmstd{.052} & 0.661\pmstd{.052} & 0.543\pmstd{.106} & 0.780\pmstd{.061}\\
 & ImageNet-ViT & 0.678\pmstd{.054} & 0.673\pmstd{.053} & 0.553\pmstd{.060} & 0.861\pmstd{.094}\\
 & InfMAE       & 0.737\pmstd{.077} & 0.735\pmstd{.077} & 0.704\pmstd{.087} & 0.834\pmstd{.097}\\
\midrule
\textbf{\method} & Three-encoder fusion & \textbf{0.823}\pmstd{.042} & \textbf{0.820}\pmstd{.044} & \textbf{0.782}\pmstd{.070} & \textbf{0.962}\pmstd{.017}\\
\bottomrule
\end{tabular}
\end{table*}

\paragraph{E.1 What the StressNet variance means.}
Standard deviations on this fold reach $\pm0.220$ (MIMMO F1) and $\pm0.243$
(late max F1). With three test participants, per-seed variation and participant
identity are entangled, so the fold ranking should be read as a controlled
comparison under fixed conditions rather than as a stable ordering. This is the
same caveat the body states in Section~6.4: because the fold was fixed after a
single random draw rather than averaged across draws, method differences within
it should not be read as a dataset-wide ranking.

\paragraph{E.2 Why the original StressNet number is not the right comparison
target.} StressNet's own published stress-detection result is measured under a
single 80\%/10\%/10\% split of individual (subject, session) trial files,
shuffled without any subject grouping; their released training code performs a
flat random shuffle over dataset indices and slices the shuffled list directly
into train/validation/test, with no held-out-subject list at any stage. Because
each dataset item is one (subject, session) recording, a participant's
stress and non-stress sessions can land on opposite sides of that split, so
the published number is not obtained under a person-disjoint protocol. Our
StressNet-dataset evaluation above instead uses a fixed subject-disjoint fold
(Appendix~C.3: 26/5/3 participants, no participant appearing on both sides).
Any result scored under our protocol is therefore expected to sit below the
original paper's headline number for this reason alone, independent of
reimplementation differences: our fold removes a same-subject leakage pathway
that the original evaluation never controlled for.

\section{Craving (RQ3) Full Results}
\setcounter{table}{0}
Supports body Section~6.5. Positive prevalence is $55.5\%$; accuracy, balanced
accuracy, and F1 use a common fixed $0.5$ threshold, and AUROC is threshold-free.
LoRA and MIL-ablation heads are omitted for this endpoint.

\begin{table*}[t]
\centering\small
\setlength{\tabcolsep}{6pt}
\caption{Subject-independent craving vs.\ non-craving detection in the OUD
cohort, $20$\,s windows, mean$\pm$std over three seeds. Bold: best mean;
underline: second best. Read with the supervision caveat: $4{,}863$ windows
carry only $172$ independent block-level labels. $^{\dagger}$ marks high-AUROC entries
whose fixed-$0.5$ balanced accuracy is $\le0.54$; these values remain part of the
AUROC ranking and expose a ranking/operating-point divergence discussed in F.1.}
\label{tab:craving_full}
\begin{tabular}{@{}lcccc@{}}
\toprule
Method & Accuracy & Balanced accuracy & F1 & AUROC\\
\midrule
\multicolumn{5}{@{}l}{\emph{Frozen linear probe (logistic regression on global token)}}\\
AnyThermal   & 0.488\pmstd{.000} & 0.525\pmstd{.000} & 0.644\pmstd{.000} & \textbf{0.970}\pmstd{.000}$^{\dagger}$\\
ImageNet-ViT & 0.226\pmstd{.000} & 0.222\pmstd{.000} & 0.165\pmstd{.000} & 0.162\pmstd{.000}\\
InfMAE       & 0.519\pmstd{.000} & 0.528\pmstd{.000} & 0.555\pmstd{.000} & 0.537\pmstd{.000}\\
Concat-3     & 0.490\pmstd{.000} & 0.527\pmstd{.000} & 0.644\pmstd{.000} & \underline{0.967}\pmstd{.000}$^{\dagger}$\\
\multicolumn{5}{@{}l}{\emph{Weight-space / randomization ensembles}}\\
Deep Ensemble      & 0.549\pmstd{.050} & 0.578\pmstd{.046} & 0.663\pmstd{.025} & 0.529\pmstd{.048}\\
Snapshot Ensemble  & 0.487\pmstd{.036} & 0.500\pmstd{.000} & 0.421\pmstd{.298} & 0.500\pmstd{.000}\\
Fast Geometric Ens.& 0.513\pmstd{.036} & 0.500\pmstd{.000} & 0.211\pmstd{.298} & 0.500\pmstd{.000}\\
SWA-Gaussian       & 0.534\pmstd{.009} & 0.544\pmstd{.032} & 0.443\pmstd{.313} & 0.733\pmstd{.169}\\
BatchEnsemble      & 0.439\pmstd{.094} & 0.465\pmstd{.106} & 0.551\pmstd{.135} & 0.398\pmstd{.149}\\
MIMMO              & \underline{0.576}\pmstd{.071} & \underline{0.603}\pmstd{.063} & \textbf{0.680}\pmstd{.028} & 0.565\pmstd{.046}\\
\multicolumn{5}{@{}l}{\emph{Learned multimodal fusion}}\\
Mixture-of-Experts & 0.491\pmstd{.087} & 0.521\pmstd{.088} & 0.622\pmstd{.069} & 0.536\pmstd{.286}\\
FuseMoE            & 0.538\pmstd{.104} & 0.543\pmstd{.122} & 0.479\pmstd{.277} & 0.575\pmstd{.052}\\
DynMM              & 0.427\pmstd{.170} & 0.453\pmstd{.178} & 0.549\pmstd{.171} & 0.424\pmstd{.264}\\
QMF                & 0.480\pmstd{.038} & 0.512\pmstd{.040} & 0.621\pmstd{.037} & 0.643\pmstd{.026}\\
MMTM               & 0.515\pmstd{.012} & 0.541\pmstd{.017} & 0.624\pmstd{.029} & 0.630\pmstd{.084}\\
Trusted Multi-View & 0.560\pmstd{.038} & 0.590\pmstd{.034} & \underline{0.673}\pmstd{.016} & 0.512\pmstd{.015}\\
\multicolumn{5}{@{}l}{\emph{Fixed posterior combination rules}}\\
Sum rule      & 0.456\pmstd{.038} & 0.487\pmstd{.046} & 0.597\pmstd{.061} & 0.500\pmstd{.020}\\
Product rule  & 0.501\pmstd{.027} & 0.536\pmstd{.025} & 0.647\pmstd{.010} & 0.655\pmstd{.034}\\
Max rule      & 0.506\pmstd{.032} & 0.540\pmstd{.030} & 0.649\pmstd{.014} & 0.506\pmstd{.058}\\
Min rule      & 0.506\pmstd{.032} & 0.540\pmstd{.030} & 0.649\pmstd{.014} & 0.796\pmstd{.070}$^{\dagger}$\\
Median rule   & 0.458\pmstd{.035} & 0.488\pmstd{.043} & 0.596\pmstd{.059} & 0.594\pmstd{.113}\\
Majority vote & 0.458\pmstd{.035} & 0.488\pmstd{.043} & 0.596\pmstd{.059} & 0.451\pmstd{.024}\\
\multicolumn{5}{@{}l}{\emph{Non-deep tabular ensembles (pooled frozen features)}}\\
Random Forest & 0.203\pmstd{.005} & 0.220\pmstd{.006} & 0.338\pmstd{.007} & 0.231\pmstd{.016}\\
ExtraTrees    & 0.173\pmstd{.013} & 0.187\pmstd{.013} & 0.295\pmstd{.019} & 0.247\pmstd{.010}\\
XGBoost       & 0.264\pmstd{.002} & 0.267\pmstd{.001} & 0.276\pmstd{.012} & 0.155\pmstd{.003}\\
LightGBM      & 0.232\pmstd{.008} & 0.239\pmstd{.008} & 0.281\pmstd{.001} & 0.210\pmstd{.025}\\
\midrule
\textbf{\method (proposed)} & \textbf{0.667}\pmstd{.054} & \textbf{0.660}\pmstd{.067} & 0.582\pmstd{.165} & 0.752\pmstd{.023}\\
\bottomrule
\end{tabular}
\end{table*}

\paragraph{F.1 Scope of the feasibility claim.}
To our knowledge, Table~\ref{tab:craving_full} is the first demonstration that
self-reported craving can be decoded from thermal video, and it establishes that
result at a usable operating point: \method attains the best accuracy ($0.667$)
and balanced accuracy ($0.660$) in the table, with MIMMO the closest baseline at
$0.603$ balanced accuracy. Four features of the table delimit what it supports.
\emph{(a)} Rankings are metric-dependent. AnyThermal ($0.970$), Concat-3
($0.967$), and the Min rule ($0.796$) exceed \method's $0.752$ AUROC. AUROC
measures ranking independently of threshold; balanced accuracy measures
decisions at the reported $0.5$ threshold. High AUROC without a transportable
operating point is of limited use for deployment, which is the regime \method
targets. The body result is an operating-point result.
\emph{(b)} The tabular ensembles fall below chance ($0.155$--$0.247$ AUROC),
indicating systematic inversion on these held-out participants rather than a
stable craving classifier. \emph{(c)} Several ensembles return exactly $0.500$
AUROC with $\pm0.000$, i.e.\ degenerate constant predictors. \emph{(d)} Four
participants are in the test split, $16/29$ sessions are single-class, and all
windows in a participant--task block inherit one report; the effective sample
size is closer to 172 block labels than 4,863 windows. 
The archived experiment therefore establishes out-of-person ranking and
operating-point associations with block-level craving reports---the first such
evidence for thermal video in this setting. 

\section{Full Aggregator Comparison}
\setcounter{table}{0}
Table~\ref{tab:mil} (referenced from body Section~\ref{sec:arch}) summarizes the headline aggregator comparison across Test-Both and Test-OUD;
Table~\ref{tab:abl_full} gives the full per-encoder form across Test-Both and Control-to-OUD. 

\begin{table*}[t]
\centering
\small
\setlength{\tabcolsep}{2.6pt}
\caption{\textbf{What each aggregator assumes, and what it scores.} Rows use the same frozen-encoder family, mixed-cohort split, and participant-centering transform (B.8). The separate-MIL row gives each encoder a dedicated temporal/spatial/MIL pipeline and learns attention over branch logits, so it changes fusion depth as well as parameter sharing. AUROC is reported for Test-Both and Test-OUD. Table~\ref{tab:abl_full} instead reports Test-Both and Control-to-OUD, so its right column is not the expanded form of the T-OUD column here.}
\label{tab:mil}
\begin{tabular*}{\textwidth}{@{\extracolsep{\fill}}llccc@{}}
\toprule
Aggregator & Instance & Views & T-Both & T-OUD \\
\midrule
ABMIL       & static vector & 1 & $0.712$ & $0.670$ \\
TransMIL    & static vector & 1 & $0.736$ & $0.653$ \\
DSMIL       & static vector & 1 & $0.698$ & $0.606$ \\
\midrule
\method branch & \textbf{reg.\ trajectory} & 1 & $0.747$ & $0.706$ \\
Separate MIL + logit attn. & reg.\ trajectory & 3 sep.\ &
  \underline{$0.744\pmstd{0.012}$} & \underline{$0.674\pmstd{0.053}$} \\
\textbf{\method} & \textbf{reg.\ trajectory} & \textbf{3 shd.} &
  $\mathbf{0.938\pmstd{.004}}$ & $\mathbf{0.810\pmstd{.002}}$ \\
\bottomrule
\end{tabular*}
\end{table*}

\begin{table*}[t]
\centering\small
\setlength{\tabcolsep}{6pt}
\caption{Aggregator ablation, full form. Published MIL heads and our single-encoder branch are each instantiated per encoder and on early concatenation (Concat-3); the last row is the full three-encoder late-fusion model. Underline: best single-encoder/Concat-3 entry per AUROC column; bold:
best overall. Mean$\pm$std over $3$ seeds.}
\label{tab:abl_full}
\begin{tabular}{@{}llcccccc@{}}
\toprule
& & \multicolumn{3}{c}{Test-Both} & \multicolumn{3}{c}{Control-to-OUD}\\
\cmidrule(lr){3-5}\cmidrule(l){6-8}
Aggregator & Encoder & Acc & F1 & AUROC & Acc & F1 & AUROC\\
\midrule
\multirow{4}{*}{ABMIL}
 & AnyThermal   & 0.681 & 0.778 & 0.676\pmstd{.021} & 0.567 & 0.669 & 0.624\pmstd{.023}\\
 & ImageNet-ViT & 0.609 & 0.707 & 0.712\pmstd{.018} & 0.617 & 0.701 & 0.672\pmstd{.012}\\
 & InfMAE       & 0.625 & 0.735 & 0.604\pmstd{.028} & 0.540 & 0.648 & 0.568\pmstd{.033}\\
 & Concat-3     & 0.717 & 0.808 & 0.711\pmstd{.031} & 0.639 & 0.737 & \underline{0.679}\pmstd{.031}\\
\multirow{4}{*}{TransMIL}
 & AnyThermal   & 0.638 & 0.734 & 0.710\pmstd{.066} & 0.747 & 0.842 & 0.665\pmstd{.018}\\
 & ImageNet-ViT & 0.600 & 0.688 & 0.714\pmstd{.015} & 0.700 & 0.804 & 0.625\pmstd{.071}\\
 & InfMAE       & 0.616 & 0.721 & 0.637\pmstd{.019} & 0.512 & 0.581 & 0.630\pmstd{.030}\\
 & Concat-3     & 0.658 & 0.753 & 0.736\pmstd{.021} & 0.611 & 0.722 & 0.605\pmstd{.016}\\
\multirow{4}{*}{DSMIL}
 & AnyThermal   & 0.659 & 0.765 & 0.632\pmstd{.040} & 0.456 & 0.527 & 0.608\pmstd{.032}\\
 & ImageNet-ViT & 0.545 & 0.633 & 0.661\pmstd{.017} & 0.574 & 0.676 & 0.578\pmstd{.009}\\
 & InfMAE       & 0.639 & 0.742 & 0.650\pmstd{.025} & 0.481 & 0.561 & 0.593\pmstd{.040}\\
 & Concat-3     & 0.608 & 0.705 & 0.698\pmstd{.048} & 0.586 & 0.689 & 0.606\pmstd{.010}\\
\multirow{3}{*}{Proposed MIL (single enc.)}
 & AnyThermal   & 0.691 & 0.782 & 0.727\pmstd{.020} & 0.746 & 0.845 & 0.649\pmstd{.004}\\
 & ImageNet-ViT & 0.538 & 0.611 & \underline{0.747}\pmstd{.013} & 0.680 & 0.787 & 0.635\pmstd{.018}\\
 & InfMAE       & 0.651 & 0.752 & 0.675\pmstd{.023} & 0.538 & 0.632 & 0.569\pmstd{.022}\\
\midrule
\textbf{\method} & 3 enc., learned late fusion & \textbf{0.884} & \textbf{0.926} & \textbf{0.938}\pmstd{.004} & \textbf{0.817} & \textbf{0.894} & \textbf{0.771}\pmstd{.013}\\
\bottomrule
\end{tabular}
\end{table*}

\paragraph{G.1 What these tables establish.}
Tables~\ref{tab:mil} and~\ref{tab:abl_full} give a consistent ordering. Regional trajectories beat static-vector instances under the three published MIL heads we tested. The proposed single-encoder branch is the strongest single-encoder entry on Test-Both ($0.747$). The full three-encoder model is highest on every protocol in either table, reaching $0.938$ on Test-Both and $0.810$ on Test-OUD. \emph{Scope.} Normalization and splits are matched across these rows, so the instance-representation contrast is attributable to the aggregator. The ``separate MIL + logit attention'' row additionally moves fusion from the embedding to the logit level, so its margin reflects fusion depth as well as parameter sharing.

\emph{Takeaway.} Under matched normalization and splits, treating instances as regional trajectories rather than static vectors improves every MIL head tested, and the shared-operator three-encoder model is the strongest configuration on both cohorts.

\section{Implementation and Hyperparameters}
\setcounter{table}{0}
Supports body Section~4 and Section~8.

\paragraph{H.1 System.} PyTorch (torch 2.8, CUDA 12.8,
Python 3.13), single NVIDIA RTX 5090 (32\,GB). Inputs are $25$ sampled frames;
features come from three frozen encoders and only the aggregation network is
updated. Hyperparameter search uses Optuna 4.8 \citep{akiba2019optuna} with the
default TPE sampler and a median pruner (2 startup trials, 15 warmup steps).
The three feature artifacts correspond to (i) AnyThermal, the thermal-distilled
DINOv2 ViT-B/14; (ii) InfMAE, the infrared masked autoencoder; and (iii) the
timm checkpoint \texttt{vit\_base\_patch16\_224(pretrained=True,
num\_classes=0)} with ImageNet input normalization. The third artifact is
therefore called \emph{ImageNet-ViT} throughout.

\paragraph{H.2 Training loop.} Mixed precision via
\texttt{torch.amp.autocast}/\texttt{GradScaler}; gradient clipping (max-norm
$5.0$) after unscaling, to control variance introduced by the class-balanced
sampler and focal loss; cosine schedule with linear warmup and decoupled AdamW
weight decay \citep{loshchilov2019adamw}; stochastic weight
averaging \citep{izmailov2018swa} over the final $30\%$ of epochs, applied only
when SWA weights beat the best single checkpoint on validation; and per-epoch
threshold selection on the validation split, never on test. Early stopping and
checkpoint selection use validation balanced accuracy
\citep{brodersen2010balanced} at the per-epoch-tuned threshold, patience 25.
Evaluation is deterministic under seeds $\{42,100,2023\}$; the first seed's
weights and validation-tuned threshold are persisted.

\paragraph{H.3 Search protocol.} Optuna tunes width $d$, heads, learning rate,
batch size, warmup, dropout, weight decay, and every loss weight on
train$\to$val only, over a ``gentle'' space biased toward slower learning and
stronger regularization; the frozen-feature model overfits an unconstrained
space within one epoch. Each protocol runs 5 trials $\times$ 200 epochs; the best
trial is retrained for 3 seeds.

\begin{table}[t]
\centering\small
\setlength{\tabcolsep}{4pt}
\caption{Optuna-selected configurations (5 trials, 200 epochs each; validation
balanced accuracy $0.938$ and $0.965$). Shared fixed settings: grad-clip $5.0$,
patience 25, final training 200 epochs, seeds $\{42,100,2023\}$. ``---'' marks a
value inactive under the selected configuration.}
\label{tab:hparams}
\begin{tabular}{@{}lcc@{}}
\toprule
Hyperparameter & Within-group & Control-to-OUD\\
\midrule
Width $d$              & 96 & 96\\
Attention heads        & 4 & 2\\
Learning rate          & $1.67\times10^{-4}$ & $2.08\times10^{-5}$\\
Warmup epochs          & 19 & 29\\
Batch size             & 32 & 32\\
Dropout                & 0.568 & 0.447\\
Weight decay           & $5.65\times10^{-4}$ & $1.48\times10^{-4}$\\
Token dropout          & 0.263 & 0.290\\
Feature noise (std)    & 0.112 & 0.078\\
Label smoothing        & 0.132 & 0.055\\
Mixup $\eta$           & 0.466 & 0.270\\
Focal loss             & disabled & enabled\\
\quad focusing $\gamma$& --- & 0.901\\
\quad class weight $\alpha_t$ & --- & 0.383\\
Deep supervision $\lambda_{ds}$   & 0.030 & 0.027\\
Consistency $\lambda_{\text{cons}}$ & 0.419 & 0.556\\
Entropy $\lambda_{\text{ent}}$    & 0.0164 & 0.0110\\
Adversarial $\lambda_{adv}$       & 0 & 0.05\\
Per-subject normalization & on & on\\
\bottomrule
\end{tabular}
\end{table}

\paragraph{H.4 Preliminary four-encoder check (LanguageBind).}
We also extracted LanguageBind features on the same
$49$-token background-masked grid ($768$-d) and reran the within-group protocol
with a four-encoder MIL model. The purpose was to ask whether performance had
already saturated with three encoders and whether LanguageBind added signal.
We used the same fixed person-disjoint split, the same $3$ seeds
$\{42,100,2023\}$, and the same SWA / focal / per-subject-normalization /
per-epoch threshold-tuning machinery. Each seed early-stopped well before the
$200$-epoch cap (epochs $26$, $34$, and $30$), consistent with the $3$-encoder
early-stopping behaviour reported elsewhere in this appendix. Table~\ref{tab:4enc}
reports the outcome: adding LanguageBind did not improve performance, so we
retained the three-encoder configuration (AnyThermal, ImageNet-ViT, InfMAE)
used throughout the rest of the paper.

\begin{table}[t]
\centering\small
\setlength{\tabcolsep}{4pt}
\caption{4-encoder (AnyThermal + ImageNet-ViT + InfMAE + LanguageBind) vs.\ the
3-encoder configuration used throughout this paper (AnyThermal + ImageNet-ViT
+ InfMAE; Table~\ref{tab:ours_full}), within-group protocol only, mean$\pm$std
over $3$ seeds. The 4-encoder row uses uniform gating and un-tuned
(non-Optuna) hyperparameters, so the comparison mixes three factors (encoder
count, gate mode, hyperparameter tuning) at once and cannot isolate
LanguageBind's individual contribution. Control-to-OUD and the matched-cohort
settings are outside the scope of this preliminary check. Adding LanguageBind
did not improve results---accuracy and F1 are lower on all three settings and
AUROC is lower on two of three (Test-Control AUROC is marginally higher,
$0.984$ vs.\ $0.979$)---so we retained the three-encoder configuration for all
other results reported in this paper.}
\label{tab:4enc}
\begin{tabular}{@{}lccc@{}}
\toprule
Setting & Accuracy & AUROC & F1\\
\midrule
\multicolumn{4}{@{}l}{\emph{3-encoder, learned gate, Optuna-tuned (Table~\ref{tab:ours_full})}}\\
Test-Both     & 0.884\pmstd{0.008} & 0.938\pmstd{0.004} & 0.926\pmstd{0.006}\\
Test-Control  & 0.927\pmstd{0.008} & 0.979\pmstd{0.002} & 0.953\pmstd{0.006}\\
Test-OUD      & 0.784\pmstd{0.010} & 0.810\pmstd{0.002} & 0.862\pmstd{0.008}\\
\multicolumn{4}{@{}l}{\emph{4-encoder (+LanguageBind), uniform gate, non-Optuna defaults}}\\
Test-Both     & 0.867\pmstd{0.013} & 0.933\pmstd{0.014} & 0.913\pmstd{0.007}\\
Test-Control  & 0.907\pmstd{0.011} & 0.984\pmstd{0.006} & 0.939\pmstd{0.007}\\
Test-OUD      & 0.774\pmstd{0.017} & 0.781\pmstd{0.035} & 0.855\pmstd{0.006}\\
\bottomrule
\end{tabular}
\end{table}

\FloatBarrier
\section{Attribution Protocol and Maps}
\setcounter{table}{0}
Supports body Section~6.3 (Grouped Kernel-SHAP result).

\begin{figure*}[t]
\centering
\includegraphics[width=\textwidth]{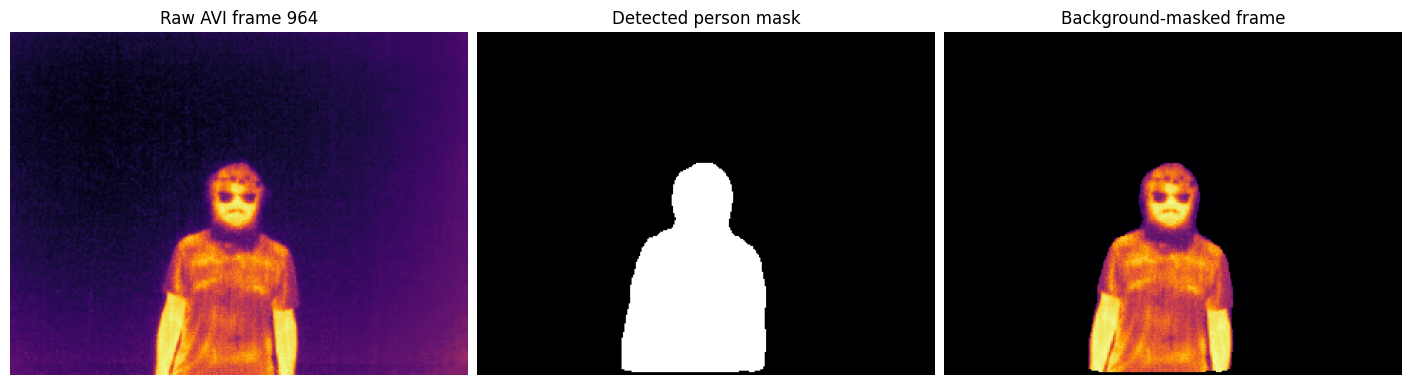}
\caption{From raw thermal frame to model input. \emph{Left}: the rendered
thermal frame used for visualization. \emph{Center}: the detected participant
mask. \emph{Right}: the background-masked thermal frame supplied to the frozen
encoders before $6\times8$ tiling. Pixels outside the person mask are zeroed in
the underlying thermal matrix, while in-mask temperatures are preserved. The
rendered detector image is never substituted for the physiological
measurement.}
\label{fig:bgmask_app}
\end{figure*}

\paragraph{I.1 Visual pipeline and reading conventions.}
The visualization suite follows the same order as the computation. First, the
participant is segmented and the image background is set to zero before any
encoder sees the frame (Figure~\ref{fig:bgmask_app}). The masked frame is then
partitioned by a fixed $6\times8$ image-space grid. Each grid location forms a
regional token, while a separate whole-frame token summarizes the complete
masked image. Repeating this construction for $25$ consecutive frames turns
each grid location into a regional \emph{trajectory}; it is this complete
trajectory, rather than an isolated pixel or a single frame, that the
subsequent attention and SHAP visualizations describe.
The $48$ grid cells and the whole-frame token together form the $N{=}49$ tokens
of Appendix~B. Throughout, a local cell is written R\emph{r}C\emph{c} for the
token at row $r$ (counted from the top of the frame) and column $c$ (counted
from the left), with $r\in\{1,\dots,6\}$ and $c\in\{1,\dots,8\}$.

\begin{figure*}[t]
\centering
\includegraphics[width=\textwidth]{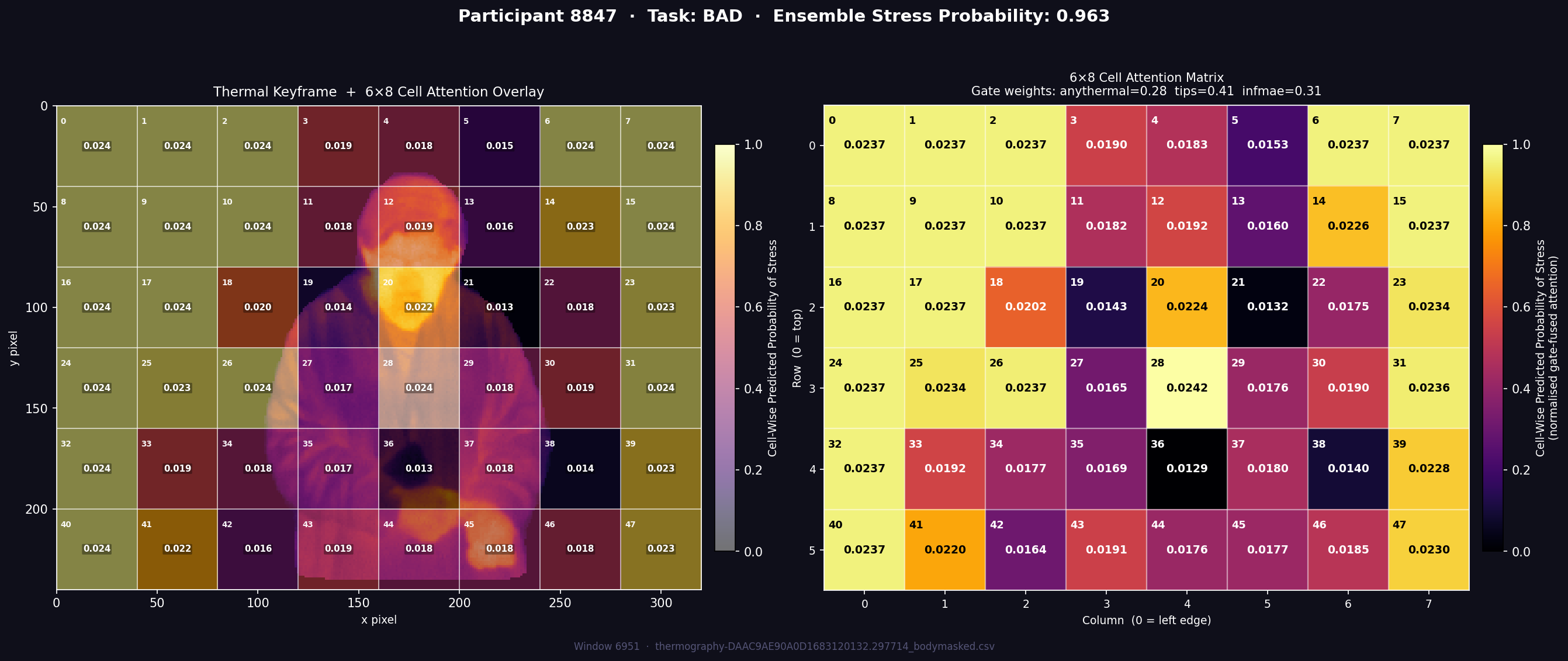}
\caption{Tiling, regional attention, and stress display for participant 8847
during the BAD task. \emph{Left}: branch-averaged MIL attention over the fixed
$6\times8$ grid, superimposed on the masked participant. \emph{Right}: the same
48 weights as a matrix, together with the fused prediction
$p(\mathrm{stress})=0.963$ and the window-specific gate weights for AnyThermal
($0.28$), ImageNet-ViT ($0.41$), and InfMAE ($0.31$). Attention is a normalized
pooling coefficient, not a signed contribution or a causal physiological map;
signed marginal effects are evaluated with grouped SHAP in the figures below.}
\label{fig:stressdisplay_app}
\end{figure*}

Figure~\ref{fig:stressdisplay_app} shows how the regional tokens enter the
prediction for one example stress window. The displayed $6\times8$ values are
the branch-averaged MIL attention weights used to pool the regional tokens;
the accompanying encoder-gate weights show how strongly the final fusion uses
the three branch outputs for that window. These attention weights say where
the pooling mechanism places weight, but they do not say whether a cell raises
or lowers the stress score. The signed grouped-SHAP maps below answer that
second question: red cells increase the fused stress logit relative to the
participant-centered reference, blue cells decrease it, and color magnitude
indicates the size of that change.

The grid is fixed in image coordinates and is not an anatomical atlas. A cell
that appears outside the silhouette in one displayed frame may overlap the
participant in another of the $25$ frames as the body moves. If a cell remains
outside the participant throughout the window, its input is zero after
background masking; a non-negligible attention or SHAP value there therefore
indicates sensitivity to the masked representation, its boundary, or the
pattern of empty grid slots---not evidence that the model measured background
temperature. This convention is essential when reading both the successful
and failed examples later in this section.

\paragraph{I.2 Grouped Kernel-SHAP.} A player is one token's \emph{complete}
trajectory: all $25$ frames and $768$ channels for that token are revealed or
masked jointly across AnyThermal, ImageNet-ViT, and InfMAE. This yields $49$
interpretable players rather than attempting to explain $\approx2.8$M scalar
inputs independently, and it matches the explanatory unit to the model's claim,
which is about regional trajectories rather than isolated channels. We explain
the final fused logit using zero in the normalized feature space as the
missing-feature reference; because inputs are participant-centered, this
corresponds to that participant's reference representation and not a raw
zero-temperature image. Kernel-SHAP \citep{lundberg2017unified} is approximated
with 256 antithetically paired coalitions per window under a constrained
weighted least-squares solve enforcing local accuracy.

\paragraph{I.3 Sampling.} The attribution set is selected independently of model
confidence: from the fixed person-disjoint test split we draw two windows from
every available participant--class stratum ($54$ windows from all $14$ held-out
participants; one OUD participant has no non-stress window). We average within
participant and then across participants, so participants with more windows
cannot dominate the aggregate maps.

\begin{figure*}[t]
\centering
\includegraphics[width=\textwidth]{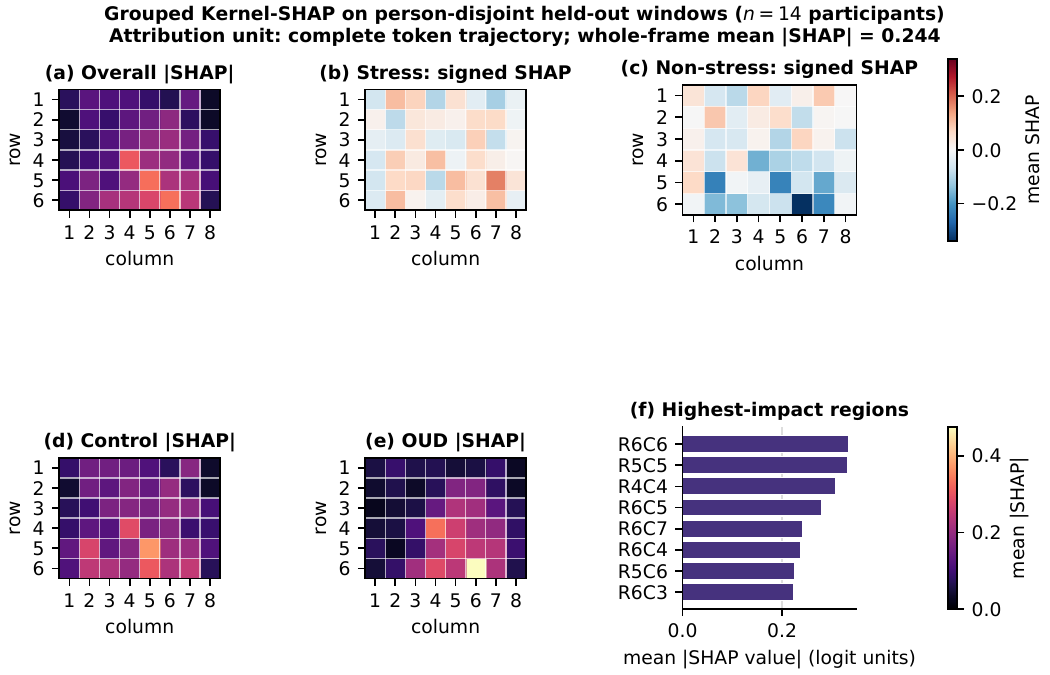}
\caption{Population-level grouped Kernel-SHAP over all $14$ held-out
participants (Appendix~I.2--I.3: 256 antithetic coalitions per window,
participant-centered reference). (a) Overall mean $|$SHAP$|$ over the
$6\times8$ grid; whole-frame token mean $|$SHAP$|$ is $0.244$ logit units.
(b)--(c) Signed mean SHAP restricted to stress-labelled and non-stress-labelled
windows respectively (red supports stress, blue supports non-stress).
(d)--(e) Overall mean $|$SHAP$|$ split by cohort; read only as a hypothesis
about cohort-dependent model behaviour (Appendix~I.4), not an anatomical or
causal finding. (f) The eight regions ranked by mean $|$SHAP$|$; the top three
(R6C6, R5C5, R4C4) are the lower-central trajectories cited in the text.}
\label{fig:shap_aggregate}
\end{figure*}

\paragraph{I.4 Population-level findings and their status.} The aggregate attribution (summarized on all held-out participants) is non-uniform,
with the largest mean absolute contributions in several lower-central grid
trajectories (R6C6, R5C5, R4C4). The whole-frame token remains material (mean
$|$SHAP$|$ $=0.244$ logit units) but is not dominant, indicating that localized
trajectories add information beyond the global descriptor. Control and OUD maps
differ visibly in where magnitude concentrates; we treat this as a hypothesis
about cohort-dependent \emph{model behaviour}, not an anatomical or causal
finding. The linear system satisfies local accuracy to numerical precision
(max absolute residual $8.9\times10^{-16}$), which is an implementation check
and not evidence of explanatory fidelity. Two external checks: masking the five
largest-$|$SHAP$|$ trajectories changes the logit by $0.708$ on average versus
$0.149$ for five random trajectories ($4.76\times$); and an independent
coalition draw preserves the regional mean-absolute-importance ranking
(Spearman $\rho=0.832$), so the pattern is not an artifact of one Monte Carlo
sample. These population-level quantities are visualized in
Figure~\ref{fig:shap_aggregate} and retained as numerical checks; Appendix~I.5
below instead uses five individual case studies to expose
participant-to-participant variation that an average map can conceal.

\paragraph{I.5 Five held-out OUD case studies.} The aggregate quantities in
Appendix~I.4 summarize all held-out participants, which is the right basis for
a population-level claim but can obscure participant-to-participant variation
that the psychophysiology literature already anticipates
(Appendix~A.1(b)): regional thermal responses to stress are not anatomically
universal across individuals. To make that variation visible rather than
merely asserted, Figures~\ref{fig:shap_o01}--\ref{fig:shap_o06} repeat the
identical grouped Kernel-SHAP protocol of Appendix~I.2--I.3 (256 antithetic
coalitions per window, participant-centered reference) separately for five
held-out OUD participants (O01, O02, O03, O05, O06) rather than pooled across
them. For each participant we contrast \method's most confidently scored
stress window against a non-stress window from the \emph{same} person. Panels
(a) and (d) show all $25$ frame outlines with the centroid path, the
elicitation task, and \method's predicted $p(\mathrm{stress})$; panels (b) and
(e) give the signed grouped-SHAP map over the $6\times8$ region grid (red
supports stress, blue supports non-stress); panels (c) and (f) rank the eight
most influential regions by signed contribution. The numbered markers link the
same image-space grid coordinate between the motion panel and the SHAP map.

The five participants do not agree on where the evidence sits. The single
strongest stress-supporting cell is R4C7 for O01, R5C7 for O02, R6C7 for O03,
R3C6 for O05, and R4C4 for O06---five different cells, none of which repeats
as the top contributor for a second participant. O01's and O02's evidence
concentrates in lateral-trunk and forearm cells; O03's strongest cells sit at
the right margin of the grid during a motion-heavy Stroop window; O05's and
O06's strongest cells are upper-body and facial. This is the per-participant
analogue of the aggregate observation in Appendix~I.4 (Control and OUD maps
differ in where magnitude concentrates) and is consistent with the regional
non-universality documented for facial and peripheral thermography
(Appendix~A.1(b)); it is not independent evidence beyond that literature. We
report this as a
qualitative case study, not a population claim: five participants are not a
sample large enough to characterize where evidence localizes for the cohort as
a whole, and each pair uses one selected stress window and one selected
non-stress window rather than an average over that participant's windows.

\begin{figure*}[t]
\centering
\includegraphics[page=1,width=0.85\textwidth]{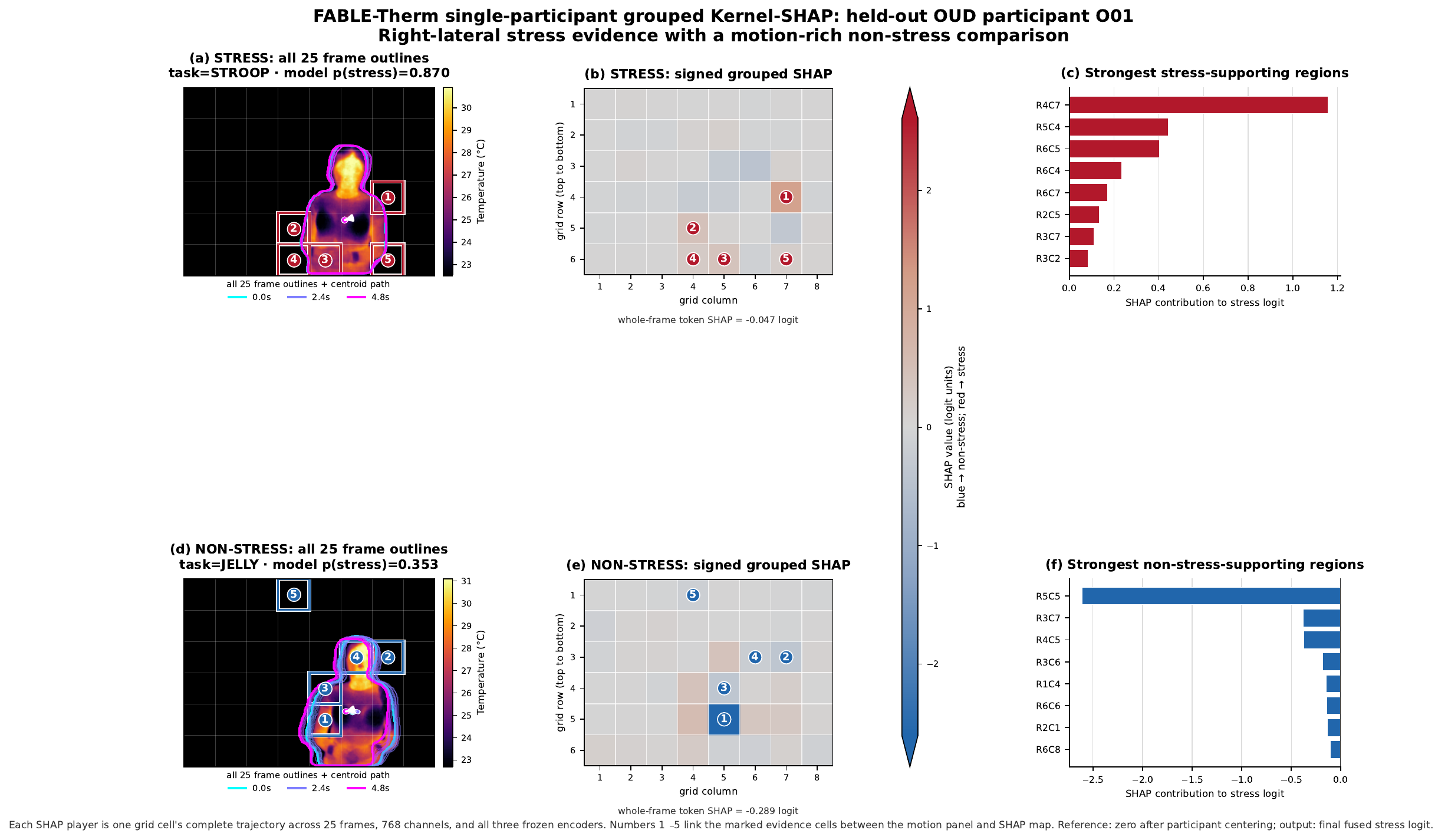}
\caption{Single-participant grouped Kernel-SHAP, held-out OUD participant O01:
right-lateral stress evidence contrasted with a motion-rich non-stress
comparison. Stress window: task Stroop, $p(\mathrm{stress})=0.870$, whole-frame
token SHAP $=-0.047$ logit. Non-stress window: task Jelly (calming video),
$p(\mathrm{stress})=0.353$, whole-frame token SHAP $=-0.289$ logit.
Top row (a--c): the selected stress window's 25-frame motion trace, signed
$6\times8$ grouped-SHAP map, and ranked stress-supporting regions. Bottom row
(d--f): the same three views for a non-stress window from O01. Numbered markers
link identical image-space grid coordinates across the motion and SHAP panels;
red raises and blue lowers the fused stress logit relative to the
participant-centered reference (Appendix~I.5).}
\label{fig:shap_o01}
\end{figure*}

\begin{figure*}[t]
\centering
\includegraphics[page=2,width=0.85\textwidth]{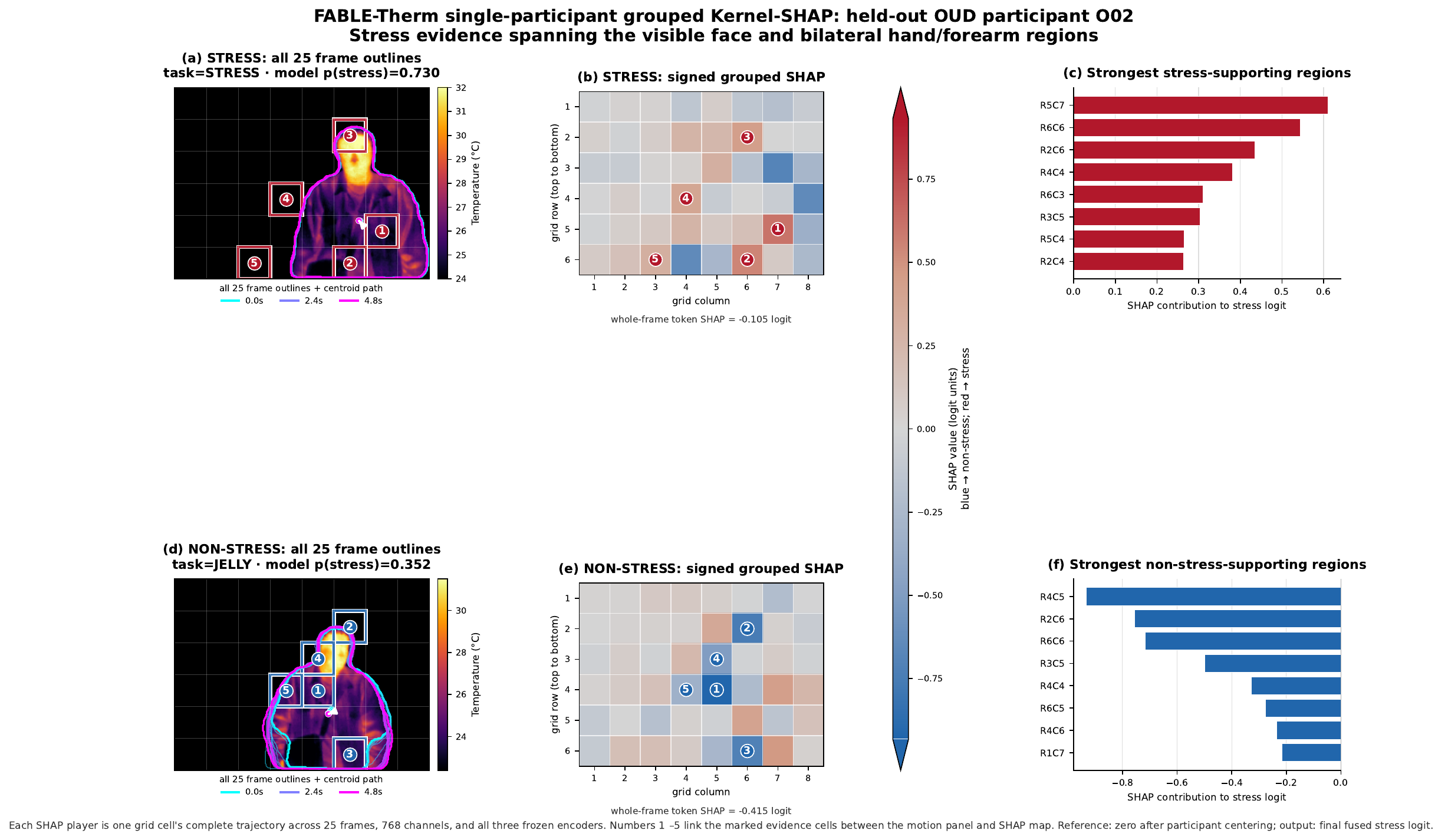}
\caption{Single-participant grouped Kernel-SHAP, held-out OUD participant O02:
stress evidence spanning the visible face and bilateral hand/forearm regions.
Stress window: task Stress-induction block, $p(\mathrm{stress})=0.730$,
whole-frame token SHAP $=-0.105$ logit. Non-stress window: task Jelly,
$p(\mathrm{stress})=0.352$, whole-frame token SHAP $=-0.415$ logit. Panel
layout and attribution protocol follow Figure~\ref{fig:shap_o01}
(Appendix~I.5).}
\label{fig:shap_o02}
\end{figure*}

\begin{figure*}[t]
\centering
\includegraphics[page=3,width=0.85\textwidth]{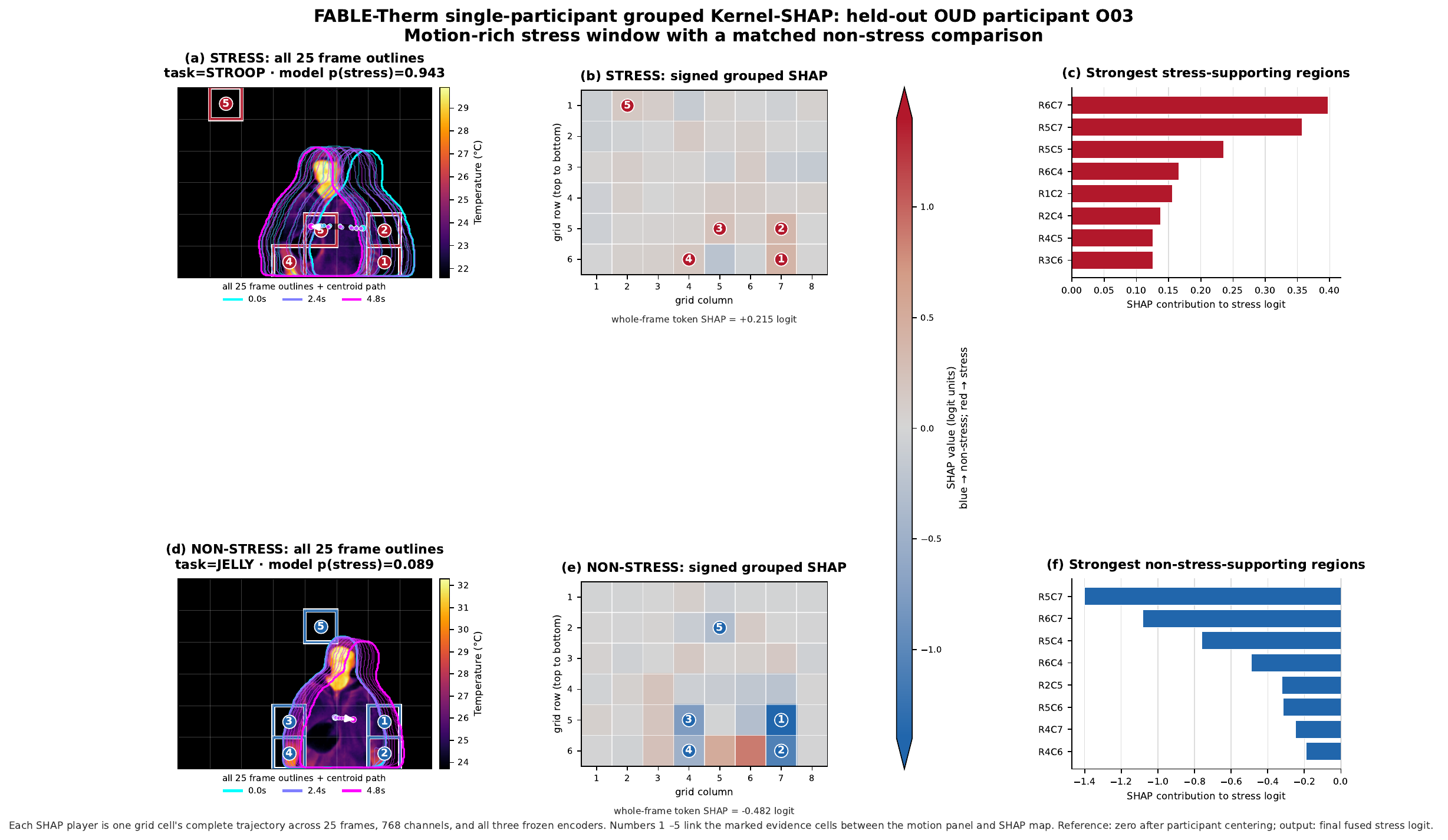}
\caption{Single-participant grouped Kernel-SHAP, held-out OUD participant O03:
a motion-rich stress window against a matched non-stress comparison. Stress
window: task Stroop, $p(\mathrm{stress})=0.943$, whole-frame token SHAP
$=+0.215$ logit. Non-stress window: task Jelly, $p(\mathrm{stress})=0.089$,
whole-frame token SHAP $=-0.482$ logit. Panel layout and attribution protocol
follow Figure~\ref{fig:shap_o01} (Appendix~I.5).}
\label{fig:shap_o03}
\end{figure*}

\begin{figure*}[t]
\centering
\includegraphics[page=4,width=0.85\textwidth]{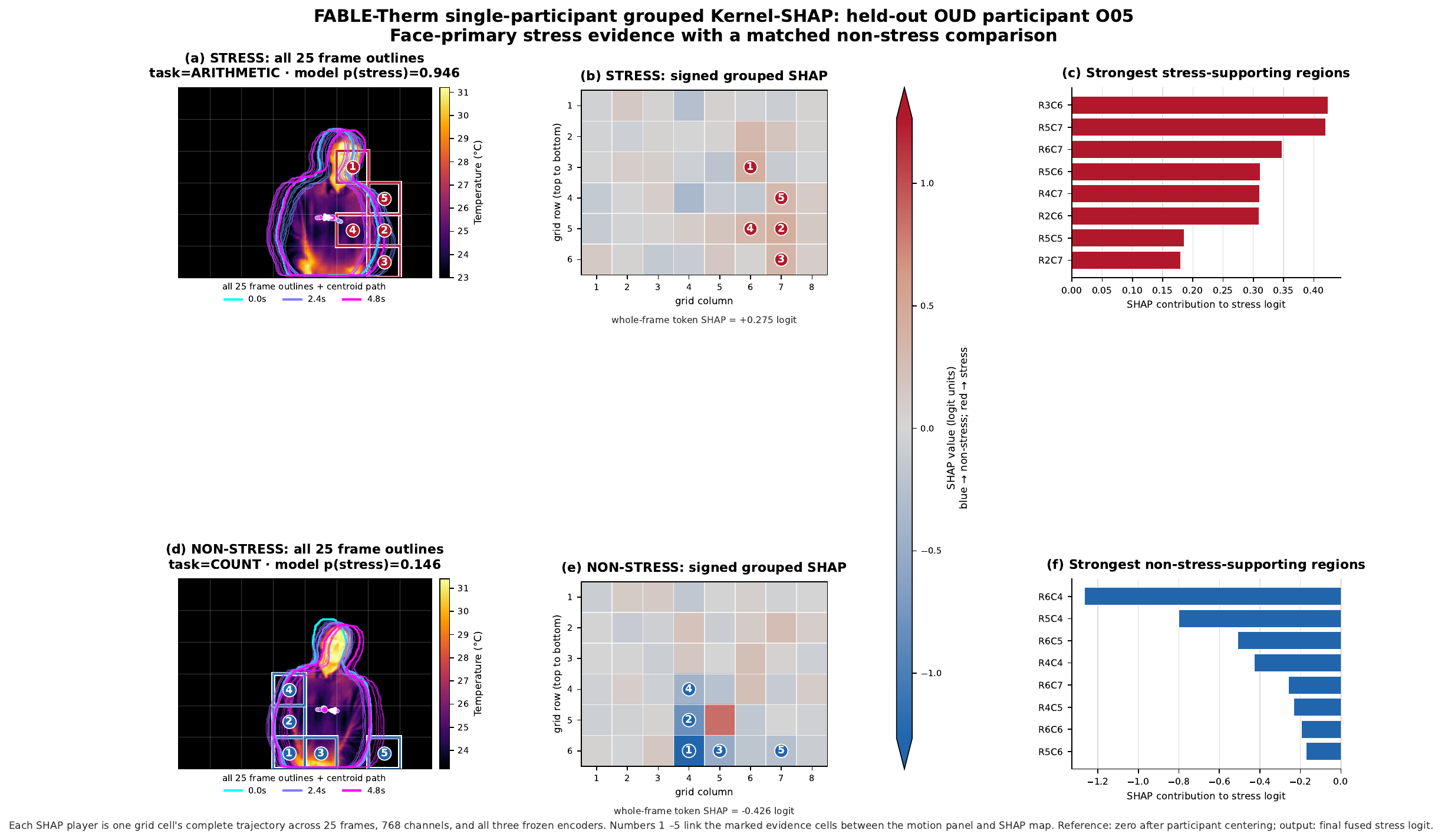}
\caption{Single-participant grouped Kernel-SHAP, held-out OUD participant O05:
face-primary stress evidence against a matched non-stress comparison. Stress
window: task Arithmetic, $p(\mathrm{stress})=0.946$, whole-frame token SHAP
$=+0.275$ logit. Non-stress window: task Count, $p(\mathrm{stress})=0.146$,
whole-frame token SHAP $=-0.426$ logit. Panel layout and attribution protocol
follow Figure~\ref{fig:shap_o01} (Appendix~I.5).}
\label{fig:shap_o05}
\end{figure*}

\begin{figure*}[t]
\centering
\includegraphics[page=5,width=0.85\textwidth]{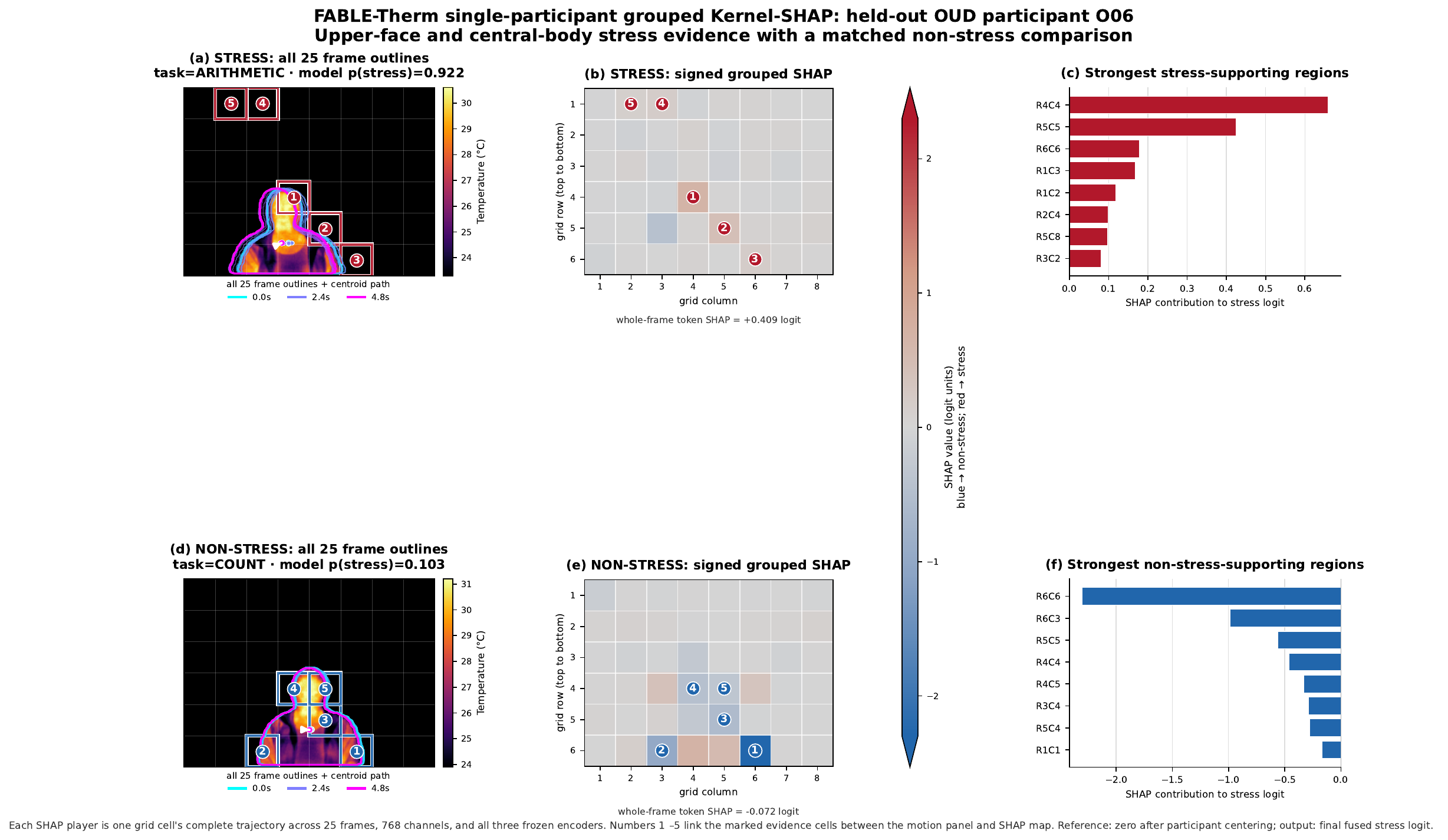}
\caption{Single-participant grouped Kernel-SHAP, held-out OUD participant O06:
upper-face and central-body stress evidence against a matched non-stress
comparison. Stress window: task Arithmetic, $p(\mathrm{stress})=0.922$,
whole-frame token SHAP $=+0.409$ logit. Non-stress window: task Count,
$p(\mathrm{stress})=0.103$, whole-frame token SHAP $=-0.072$ logit. Panel
layout and attribution protocol follow Figure~\ref{fig:shap_o01}
(Appendix~I.5).}
\label{fig:shap_o06}
\end{figure*}

\begin{figure*}[t]
\centering
\includegraphics[width=\textwidth]{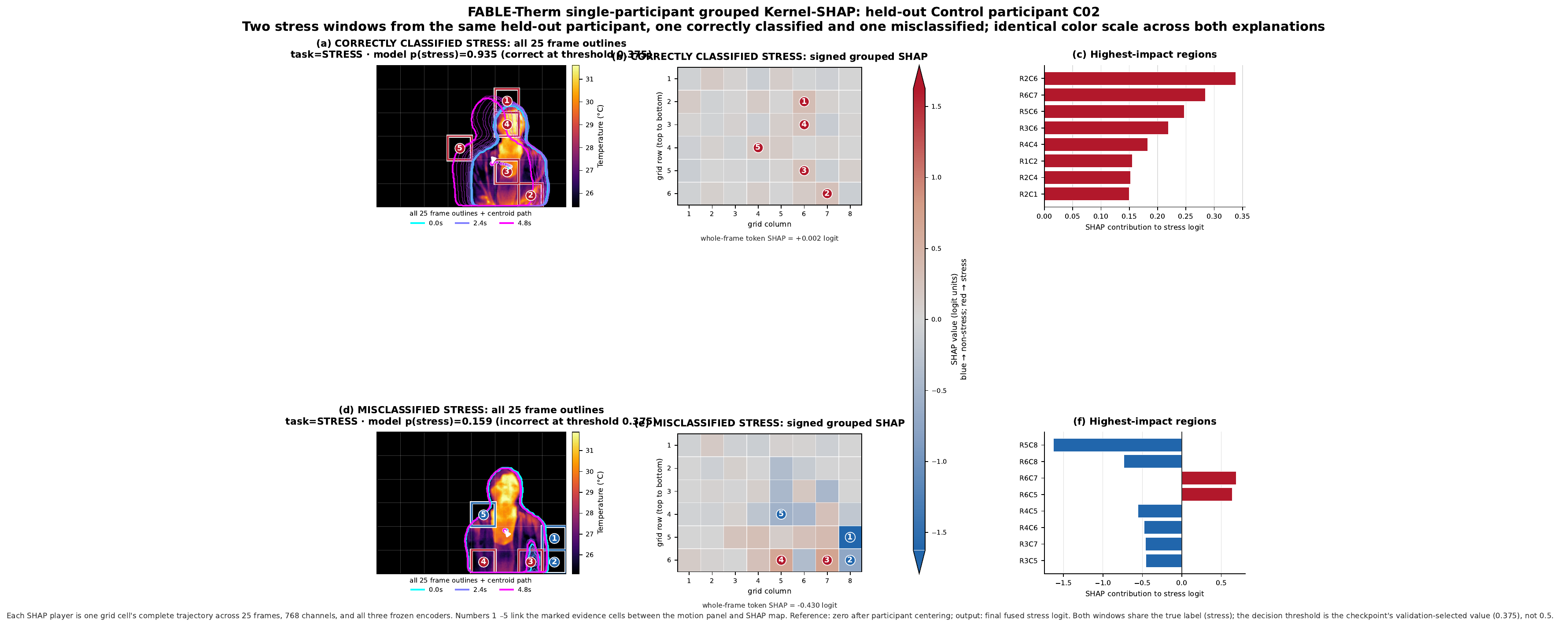}
\caption{Correctly classified vs.\ misclassified stress windows, held-out
Control participant C02, both from the \emph{stress} elicitation block.
Top row: $p(\mathrm{stress})=0.935$ (correct at threshold $0.375$). Bottom row:
$p(\mathrm{stress})=0.159$ (incorrect). Regions are ranked by absolute SHAP
impact (Appendix~I.6); red supports stress and blue supports non-stress. The
failed prediction is dominated by the peripheral R5C8 and R6C8 trajectories at
the right grid boundary. Attribution protocol follows Appendix~I.2--I.3.}
\label{fig:shap_correct_control}
\end{figure*}

\begin{figure*}[t]
\centering
\includegraphics[width=\textwidth]{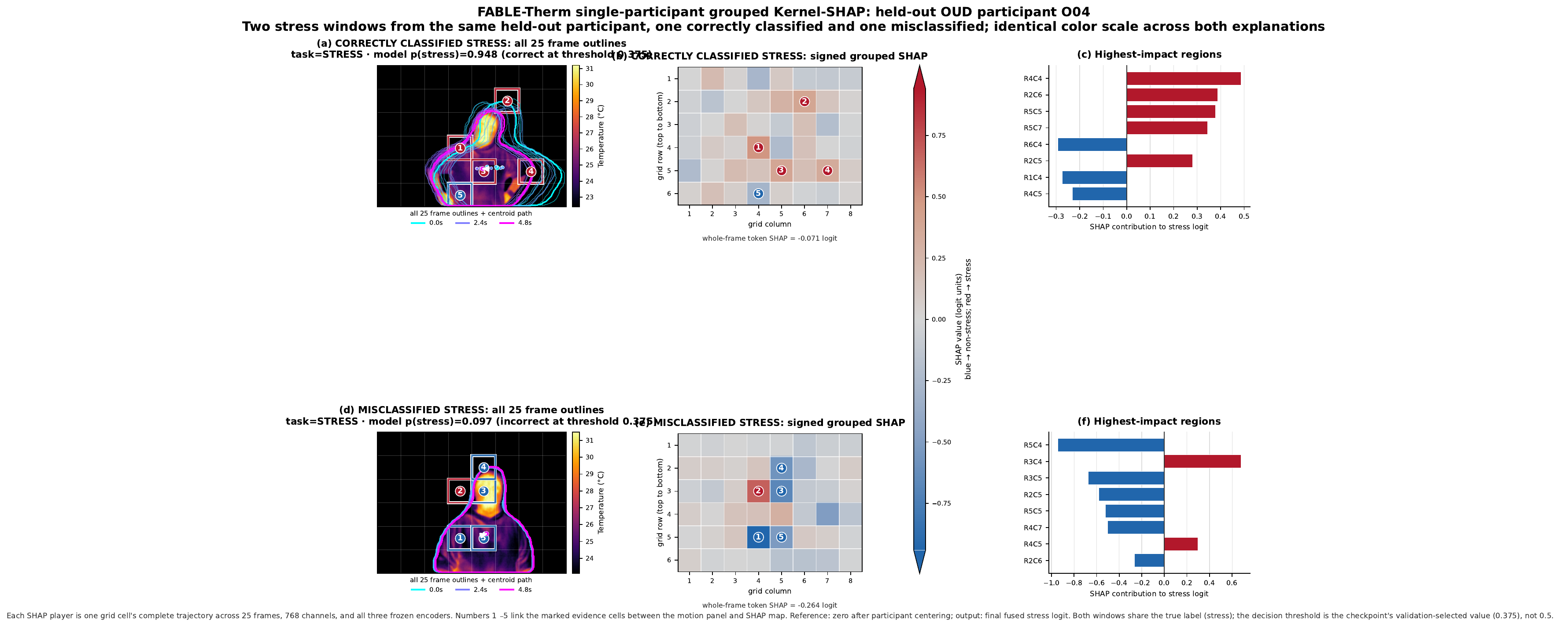}
\caption{Correctly classified vs.\ misclassified stress windows, held-out OUD
participant O04, both from the \emph{stress} elicitation block. Top row:
$p(\mathrm{stress})=0.948$ (correct at threshold $0.375$). Bottom row:
$p(\mathrm{stress})=0.097$ (incorrect). Regions are ranked by absolute SHAP
impact (Appendix~I.6); red supports stress and blue supports non-stress. Here
the strongest error-driving trajectory, R5C4, remains on or near the body,
showing that boundary reliance is not the only observed failure pattern.
Attribution protocol follows Appendix~I.2--I.3.}
\label{fig:shap_correct_oud}
\end{figure*}

\paragraph{I.6 Correctly classified vs.\ misclassified stress windows.}
\label{sec:shap_correct_wrong}
Figures~\ref{fig:shap_o01}--\ref{fig:shap_o06} contrast a stress window against
a non-stress window. A different failure-analysis question is: for the
\emph{same} true label, why does the model get one window right and another
window from the \emph{same person} wrong? Figures~\ref{fig:shap_correct_control}
and~\ref{fig:shap_correct_oud} answer this for one held-out Control participant
(C02) and one held-out OUD participant (O04), using the identical grouped
Kernel-SHAP protocol (Appendix~I.2--I.3: 256 antithetic coalitions,
participant-centered reference, seed-42 checkpoint) applied to two windows that
both carry the true label stress but land on opposite sides of that
checkpoint's validation-selected decision threshold ($0.375$, not $0.5$;
Appendix~H.2). Both selected pairs happen to come from the same elicitation
task (\emph{stress} block) for their participant, so the contrast is not
confounded by task identity: same person, same task, same true label, opposite
model decision. Because both windows share the true label, ranking regions by
\emph{evidence for that label} would rank both rows identically; instead we
rank by absolute impact and let the sign of the marked cells carry the story
(red pushes toward stress, blue toward non-stress).

For C02, the correctly classified window ($p(\mathrm{stress})=0.935$) combines
positive evidence distributed across the head/upper-body area and the moving
right/lower body boundary (including R2C6, R5C6, and R6C7). In the
misclassified window from the same stress block
($p(\mathrm{stress})=0.159$), the two largest effects instead sit at the
rightmost grid boundary and strongly oppose stress (R5C8: $-1.627$; R6C8:
$-0.732$). They outweigh R6C7 ($+0.690$), the strongest cell supporting the
true label. This is the clearest example of the proposed failure signature:
the prediction is controlled by spatially peripheral, background-adjacent
slots rather than by a stable pattern over the participant.

The O04 comparison shows why that interpretation must not be overgeneralized.
Its correctly classified window ($p(\mathrm{stress})=0.948$) is supported by
several cells on or near the silhouette (R4C4, R2C6, and R5C5). Its
misclassified window ($p(\mathrm{stress})=0.097$) is again dominated by a
localized contribution toward non-stress, but the largest such cell (R5C4:
$-0.942$) lies on or near the lower torso in the motion panel rather than at an
obviously empty image boundary; it overwhelms the positive R3C4 contribution
($+0.679$). Thus, out-of-body or boundary sensitivity is directly visible in
the C02 failure but is not a universal explanation for every error. What the
two cohorts share is a spatially concentrated negative contribution that
overrides evidence for the true stress label.

These maps support a diagnostic, not a physiological causal claim. Because a
marker is drawn over all $25$ outlines, an apparently empty cell in any one
outline can still contain the moving participant elsewhere in the window.
Only a cell that remains unoccupied across the trajectory should be described
as out-of-body. Large effects in such cells indicate reliance on the geometry
of the masked input or on empty-slot patterns---a potential shortcut and a
useful target for occupancy-aware regularization or auditing---rather than a
thermal stress response measured outside the body.

\paragraph{I.7 Per-encoder attribution and 25-frame window probabilities.}
\label{sec:shap_branch}
The analyses above all explain the fused logit $\ell^\star=\mathrm{FinalHead}(b^\star)$, the final head's output on the fused embedding $b^\star$ of Eq.~\eqref{eq:productbag}, with $\hat p=\sigma(\ell^\star)$ the reported stress probability. 
A separate question is how the three frozen encoders individually value the regions the fused model relies on. This is well-posed here because of how branch logits are computed (Appendix~B): each encoder's branch logit $\ell^{(e)}$ is a function of that encoder's own projected, temporally collapsed, spatially contextualized, MIL-pooled bag $b^{(e)}$ alone---$\ell^{(e)}=\mathrm{Head}_e(b^{(e)})$
never reads the other two encoders' tensors. A branch is therefore explainable
by the identical grouped Kernel-SHAP construction used throughout this
appendix, substituting $\ell^{(e)}$ for $\ell^\star$ and masking only encoder
$e$'s 49 tokens per coalition (the other two encoders' tensors are left
unmasked but are inert to $\ell^{(e)}$ by construction). We reuse the same 256
coalitions and weights already drawn for the fused explanation of each window,
so the three branch explanations and the fused explanation of a given window
are computed from matched coalition draws.

\begin{figure*}[t]
\centering
\includegraphics[width=\textwidth]{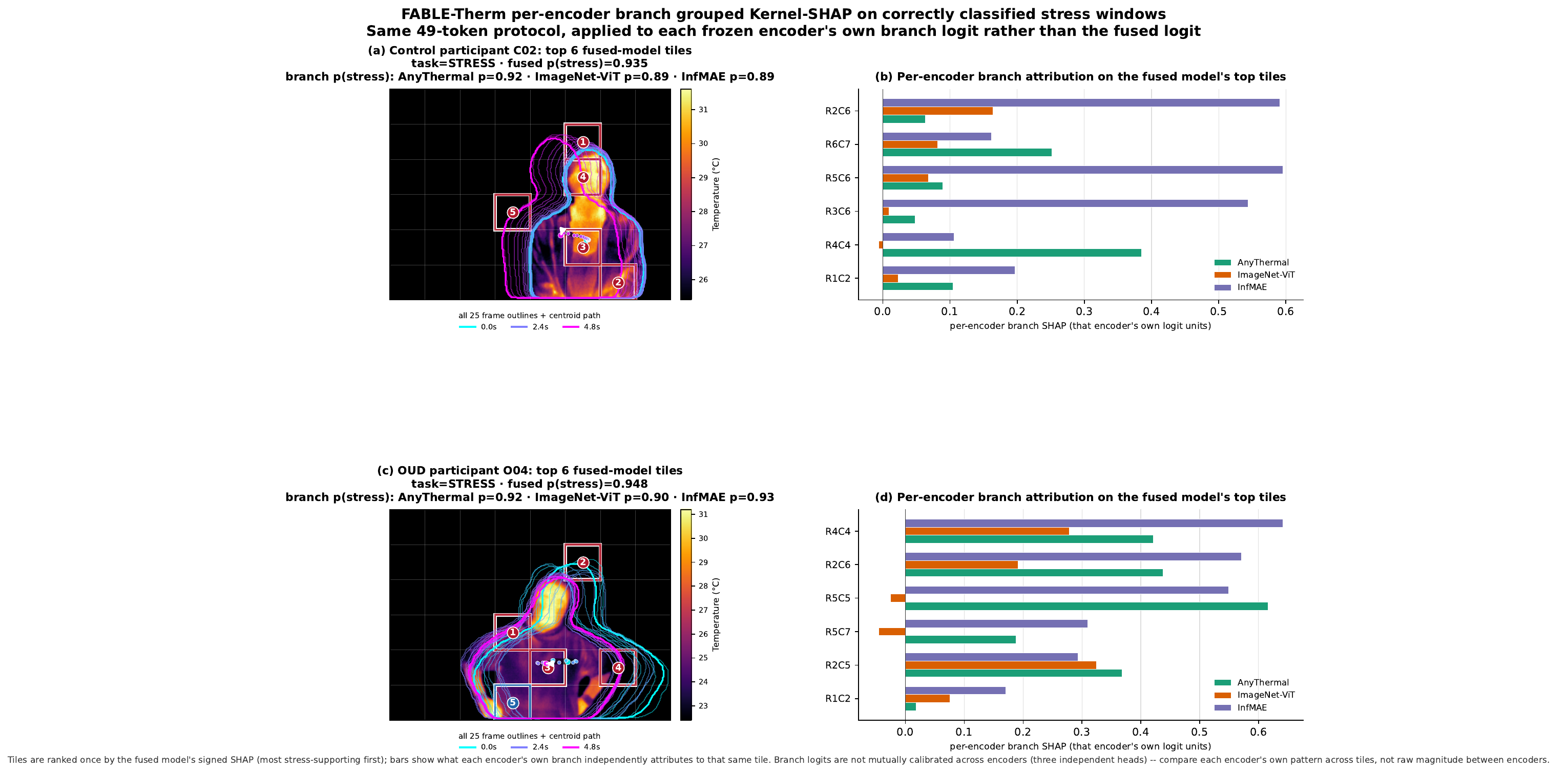}
\caption{Per-encoder branch grouped Kernel-SHAP on one correctly classified
stress window per cohort (top: Control C02; bottom: OUD O04; same windows as
the top rows of Figures~\ref{fig:shap_correct_control}
and~\ref{fig:shap_correct_oud}). Tiles are ranked once by the fused model's
signed SHAP (most stress-supporting first, left panel); bars in the right
panel show each encoder's own branch-SHAP value for the complete 25-frame
trajectory at that tile, in that encoder's own branch-logit units (not mutually
calibrated across encoders; Appendix~I.7). The subtitle reports one
$p(\mathrm{stress})$ per encoder for the entire 25-frame window; it is not a
sequence of frame-wise probabilities.}
\label{fig:shap_branch}
\end{figure*}

For one correctly classified stress window per cohort (the top rows of
Figures~\ref{fig:shap_correct_control} and~\ref{fig:shap_correct_oud}), we rank
the $48$ local trajectories once by the fused model's signed SHAP value and
retain the six strongest stress-supporting trajectories. Figure~\ref{fig:shap_branch}
then asks each encoder the same question: how much does that complete regional
trajectory, spanning all $25$ frames, change this encoder's own branch logit?
The bars are therefore regional SHAP contributions accumulated over the
25-frame input trajectory. They are not 25 separate frame-level probabilities.
Each encoder produces one branch probability for the complete window, shown in
the panel subtitle, and the fusion produces one final window probability.

For C02, the complete-window branch probabilities are $0.92$ for AnyThermal,
$0.89$ for ImageNet-ViT, and $0.89$ for InfMAE; their fusion gives
$p(\mathrm{stress})=0.935$. For O04, the corresponding probabilities are
$0.92$, $0.90$, and $0.93$, and the fused probability is $0.948$. The encoders
therefore agree on the stress decision while telling different spatial
stories. InfMAE has the largest branch-SHAP value on five of the six displayed
trajectories for C02 and four of six for O04, whereas ImageNet-ViT is near zero
or slightly negative on several of the same trajectories (C02 R4C4:
$-0.006$; O04 R5C5: $-0.025$; O04 R5C7: $-0.045$). The pattern is not uniform,
however: AnyThermal is largest at C02 R4C4 ($0.385$, compared with InfMAE's
$0.106$) and O04 R5C5 ($0.617$, compared with InfMAE's $0.549$).

Because the branch heads use independently scaled representations, their logit magnitudes are not mutually calibrated (Appendix~B.6). So bar heights are compared within an encoder's explanation rather than across encoders. Read that way, the branch attributions answer a question the fused map cannot: the three encoders converge on nearly the same whole-window probability ($0.89$--$0.93$ on the two examined windows) while assigning their evidence to different regional trajectories. Complementarity is therefore observable in the attributions themselves rather than inferred from the fusion gain alone, giving the learned window-specific gate a mechanistic rationale that matches its measured benefit in Table~\ref{tab:full_main}.

\section{Reproducibility Artifact}
\label{app:repro}
\setcounter{table}{0}
Supports body Section~8. The artifact is organized so that a third party can
reproduce the paper at three increasing levels of data access.

\paragraph{J.1 Tier 1 --- fully public, no request required.}
The StressNet pipeline (RQ2) end to end: window construction, region tiling,
frozen-feature extraction, the aggregation model, the fixed subject-disjoint
fold definition, training and evaluation scripts, and the table-generating
script. StressNet is public \citep{kumar2021stressnet}, so
Tables~\ref{tab:sn_main} and \ref{tab:sn_mil} are reproducible without contacting
the authors. This is the recommended entry point for anyone applying the method
to a new thermal corpus.

\paragraph{J.2 Tier 2 --- derived features, release conditional on consent.}
Per-encoder frozen feature tensors for the primary corpus
($25\times49\times768$ per window per encoder), the fixed split files
(\texttt{fixed\_split.json}, \texttt{fixed\_split\_control\_oud.json}), per-seed
prediction files, confusion matrices, balanced-accuracy/precision/recall
breakdowns, selected hyperparameters, and the seed-42 attribution checkpoint.
Region tiling and participant centering may reduce direct visual
reconstructibility, but they do not make these records anonymous: the features
remain person-linkable, and craving labels disclose sensitive information about
people with OUD. Reidentification risk and the scope of the original consent are
separate questions. Release of this tier is therefore conditional on explicit
IRB/consent authority, a documented threat assessment, data-use terms, and
removal of unnecessary participant/task metadata; otherwise it belongs under
the governed procedure in J.3.

\paragraph{J.3 Tier 3 --- raw recordings, governed request.}
Primary-corpus raw thermal recordings are human-participant data and cannot be
presumed public. Access follows a documented request procedure consistent with
the approved protocol; code and derived test data for the tiers above are
available at the repository linked at the start of this appendix.

\paragraph{J.4 Applying the method to a new domain.}
The architecture depends only on the token count $N$, so porting requires: a
person/object mask, a tiling choice ($N=1+RC$), any frozen encoder emitting a
per-region embedding, and window labels. We provide the StressNet configuration
($N{=}97$) alongside the primary configuration ($N{=}49$) precisely to
demonstrate that the grid is a parameter and not a hard-coded body model.

\paragraph{J.5 Operating-point provenance.}
Table~\ref{tab:oppoint} uses the saved seed-42 learned-fusion checkpoint and one
validation-selected threshold, $0.374906$, chosen to maximize validation
accuracy (not balanced accuracy). That identical threshold is then applied to
both held-out cohorts. The Control confusion counts are TP 1,694, FP 43, FN 133,
TN 436; the OUD counts are TP 666, FP 110, FN 112, TN 112. Intervals use 2,000
participant-cluster resamples. In addition to the intervals printed in the
table, Control FNR is $[0.009,0.171]$ and specificity $[0.796,0.987]$; OUD FNR is
$[0.052,0.246]$ and sensitivity $[0.754,0.948]$. With eight and six participants,
these intervals quantify substantial uncertainty and do not turn the selected
threshold into a deployment recommendation.


\begin{table}[t]
\centering\small
\setlength{\tabcolsep}{3pt}
\caption{\textbf{Per-cohort deployment operating point} at the model's
validation-selected threshold (seed 42), with participant-bootstrap $95\%$ CIs.
FPR = unwanted-check-in rate; FNR = missed-stress rate. The comparison is within
one model at the same threshold, but population, site, and cohort composition
remain entangled; it is descriptive rather than causal. Supports body Section~8.}
\label{tab:oppoint}
\begin{tabular}{@{}lcccc@{}}
\toprule
Cohort & FPR & FNR & Sensitivity & Specificity \\
\midrule
Control ($n{=}8$) & $0.090$ & $0.073$ & $0.927$ & $0.910$ \\
& \multicolumn{2}{c}{\scriptsize FPR 95\% CI: $[0.013,0.204]$} & \multicolumn{2}{c}{\scriptsize Sens. 95\% CI: $[0.829,0.991]$} \\[2pt]
OUD ($n{=}6$)     & $\mathbf{0.495}$ & $0.144$ & $0.856$ & $0.505$ \\
& \multicolumn{2}{c}{\scriptsize FPR 95\% CI: $[0.300,0.715]$} & \multicolumn{2}{c}{\scriptsize Spec. 95\% CI: $[0.285,0.700]$} \\
\bottomrule
\end{tabular}
\end{table}

\FloatBarrier

\section{Ethics, Governance, and Non-Uses}
\setcounter{table}{0}
Supports body Section~8.

\paragraph{K.1 Intended benefit pathway.} An opt-in, non-contact signal could
support low-burden check-ins when wearing, charging, or maintaining a sensor is
impractical. The appropriate role is to prompt a person-centred conversation or
offer support---never to diagnose stress, infer substance use, or replace
self-report and clinical judgment.

\paragraph{K.2 Prohibited uses, as licence terms.} Law enforcement; probation,
parole, or drug-court supervision; employment screening or workplace monitoring;
insurance underwriting; determination of eligibility for benefits, housing, or
treatment; and any covert or non-consensual monitoring. We state these as
licence conditions rather than recommendations because a contactless sensor is
usable without the subject's cooperation, which is exactly the property that
makes advisory language insufficient. A licence can govern authorized
recipients, but it cannot by itself prevent copying, independent reimplementation,
or all downstream misuse; technical access controls, contracts, audit, and
enforcement are also required.

\paragraph{K.3 Conditions for a responsible pilot.} Explicit and revocable
consent; a clear and stated benefit to the participant; short retention or
on-device processing; access and deletion controls; human review of every
model-prompted action; prospective multi-site validation with common calibration
procedures; pre-specified subgroup audits; and governance co-designed with
people with OUD, clinicians, and community advocates. \method is a research
prototype and does not currently satisfy these requirements.

\paragraph{K.4 Failure modes and who bears them.} False positives can trigger
unwanted intervention or stigma; false negatives can create false reassurance.
At the seed-42 operating point in Table~\ref{tab:oppoint}, both error types are
higher for OUD participants than for Control participants, which means the
population targeted for benefit carries the higher observed error rate in that
analysis. The intervals are wide and the study is not a deployment validation.
Any deployment that cannot support a per-person
calibration period should not be deployed for this population.

\paragraph{K.5 On ``privacy-preserving''.} Thermal imaging avoids conventional
colour appearance but records a body-derived signal, is person-linkable, and is
not anonymous. Our subject-adversarial term (Appendix~B.7) is a training-time
pressure against identity encoding, not a privacy guarantee, and we make no
formal privacy claim; body-derived affective signals of this kind increasingly
fall under special-category and cognitive-biometric data protections
\citep{ienca2024cognitive}.

\end{document}